\pdfoutput=1
\documentclass[10pt,journal,compsoc]{IEEEtran}
\usepackage{booktabs} 
\usepackage{graphicx}
\usepackage{subfigure}
\usepackage{amsthm}
\usepackage{amssymb}
\usepackage{amsmath}
\usepackage{multirow}
\usepackage{enumerate}
\usepackage{amsfonts}
\usepackage{color}
\usepackage{url}
\usepackage{tabularx}
\usepackage{hyperref}
\usepackage{ragged2e}
\usepackage[justification=centering]{caption}

\ifCLASSOPTIONcompsoc
  \usepackage[nocompress]{cite}
\else
  \usepackage{cite}
\fi
\ifCLASSINFOpdf
\else
\fi
\begin{document}
%
\title{Reinforced Imitative Graph Learning for Mobile User Profiling}
%
%
%
%

\author{
Dongjie Wang, Pengyang Wang, Yanjie Fu, Kunpeng Liu, Hui Xiong~\IEEEmembership{Fellow,~IEEE} and Charles E. Hughes

\IEEEcompsocitemizethanks{
\IEEEcompsocthanksitem Dongjie Wang and Pengyang Wang are with equal contribution.
\IEEEcompsocthanksitem Dongjie Wang, Yanjie Fu, Kunpeng Liu ,and Charles E. Hughes are with the University of Central Florida (UCF).
Email: {wangdongjie@knights.ucf.edu,
yanjie.fu@ucf.edu, kunpengliu@knights.ucf.edu, 
charles.hughes@ucf.edu}.
\IEEEcompsocthanksitem Pengyang Wang is with the University of Macau Email: {pywang@um.edu.mo}.
\IEEEcompsocthanksitem Hui Xiong is with the Rutgers University Email: {hxiong@rutgers.edu}.}
}

%
%

\markboth{Journal of \LaTeX\ Class Files,~Vol.~14, No.~8, August~2015}%
{Shell \MakeLowercase{\textit{et al.}}: Bare Advanced Demo of IEEEtran.cls for IEEE Computer Society Journals}

\IEEEtitleabstractindextext{%
\justifying
\begin{abstract}


Mobile user profiling refers to the efforts of extracting users' characteristics from mobile activities.
In order to capture the dynamic varying of user characteristics for generating effective user profiling, we propose an imitation-based mobile user profiling framework. 
Considering the objective of teaching an autonomous agent to imitate user mobility based on the user's profile, the user profile is the most accurate when the agent can perfectly mimic the user behavior patterns.
The profiling framework is formulated  into a  reinforcement learning task, where an agent is a next-visit planner, an action is a POI that a user will visit next, and the state of the environment is a fused representation of a user and spatial entities.
An event in which a user visits a POI will construct a new state, which helps the agent predict users' mobility more accurately.  
In the framework, we introduce a spatial Knowledge Graph (KG) to characterize the semantics of user visits over connected spatial entities.
Additionally, we develop a mutual-updating strategy to quantify the state that evolves over time.
Along these lines, we develop a reinforcement imitative graph learning framework for mobile user profiling.
Finally, we conduct extensive experiments to demonstrate the superiority of our approach.

\end{abstract}

\begin{IEEEkeywords}
Mobile User Profiling; Incremental Learning; Reinforcement Learning; Spatial Knowledge Graph.
\end{IEEEkeywords}}

\maketitle

\IEEEdisplaynontitleabstractindextext

%
\IEEEpeerreviewmaketitle


\section{Introduction}
\label{sec:introduction}

\IEEEPARstart{M}{obile} user profiling refers to the efforts of extracting user interests and behavioral patterns from mobile activities.
Consider the existence of many mobile users in a city, 
each  user is equipped with mobile sensing equipment moving from one location to another location and generates a mobility event stream in real time. 
Classical mobile user profiling collects large-scale spatio-temporal event data, and then, learns profile representations to characterize user patterns and preferences using the collected data. 

Prior literature in mobile user profiling includes: (1) explicit profile extraction~\cite{godoy2005user}, (2) factorization-based approaches~\cite{griesner2015poi,lian2014geomf}, and (3) deep learning-based approaches~\cite{yang2017bridging,yin2017spatial}.
Other studies exploit adversarial deep learning to emphasize substructure patterns in mobile user profiling~\cite{wang2019adversarial}.
All of these approaches can be regarded as exploiting user activities' prediction bias or reconstruction loss of the users' profile structure as the learning criteria to model users' profiles.
The key limitation of these approaches is the modeling procedure is solely based on the individual user, which lacks a global perception of the dynamic varying environment.
However, the behavioral data generated by the mobile users are usually a mixed-user, spatially and temporally discrete, event stream. 
Users inject their impacts in the environment mixed and chronologically, leading to the dynamic changes in the environment that inversely affect a users' decision.

After exploring many profiling methods, we found that minimizing user activities' prediction bias or reconstruction loss may not be the best criteria to evaluate profiling accuracy. 
Unlike traditional loss minimization, we identify a better criteria, which we call the \emph{imitation based criteria}: considering the objective of teaching an autonomous agent to imitate a mobile user to plan where a user will visit next, based on the profile of the user. The user profile is the most accurate when the agent can perfectly copy the activity patterns of the user.

The emerging reinforcement learning can train an agent to plan for its next actions in order to function in its environment. Such an ability provides great potential to implement the imitation based criteria in order to achieve more accurate user profiling.

As a result, we propose to formulate the problem into a reinforcement learning framework. 
In this framework, an agent is a next-visit planner that tries to perfectly imitate a set of mobile users.
The state of environment is a fused representation of a given user and spatial entities ({\it e.g.}, POIs, activity types, functional zones).
An action is a POI that a given mobile user will visit, which is estimated based on the state of the environment by the agent. 
An event where the user takes the action to visit the POI, will change the environment, resulting into a new state of the user and the spatial entities, which helps the agent to better estimate the next visit.
The reward of an action is the reduction of the gap between the agent's activity patterns and the user's activity patterns. 
After the reformulation, our new objective is to exploit the reinforcement learning framework to extract dynamic profile representations of various users in the state of the environment by incremental learning from an user-visit event stream.


To further improve the profiling accuracy of the framework, we analyze how a mobility event connect mobile users with spatial entities ({\it e.g.}, POIs, functional zones), and identify \emph{two important structured information}.

Firstly, there is \emph{semantic connectivity among spatial entities}. 
Specifically, the profile of a user can be reflected by  a sequential composition of mobility events. The semantics of an event are about which building the user visits (POI), what type of activity the user conducts during the visit (POI category), and in which region the POI is located (urban functional zone). 
Therefore, improving the representations of POIs, activity types, and functional zones of events can, in return, improves user profiling. 
The semantic connectivity among these spatial entities refers to the observation that every time a mobile user visits a POI, a new connection is established or reinforced among a POI, an activity type, and a functional zone, which indeed is a heterogeneous graph with geographic knowledge. 
Therefore, we propose to use a spatial knowledge graph ({\it KG}) to describe such semantic connectivity, and a translation based embedding method is employed to learn the embedding of the spatial {\it KG}. 

Secondly, there is \emph{mutual influence between users and spatial entities}. 
An event that a user visits and interacts with the POI, will change and reinforce the edges (semantic connectivity) of the spatial knowledge graph, resulting into new state representations of POIs, activity types, and functional zones. 
If the states of spatial entities are updated, once a user visits these newly-updated spatial entities, then the state of the user profile will be updated as well. 
In other words, the state updates between spatial entities and users are sequentially nested together per mobility event.
We propose a sequentially nested updating strategy to update user states by jointly considering the spatial  KG and temporal contexts, and update spatial  KG states by jointly considering user states and temporal contexts.

In summary, we propose a reinforced imitative graph learning framework for mobile user profiling by integrating a reinforcement learning framework with spatial KG to solve. Our contributions are: 
(1) We propose a new imitation based criteria for evaluating the accuracy of user profiling: the better an agent imitates a mobile user, the more accurate the user profile is.   
(2) Motivated by the imitation based criteria, we reformulate the mobile user profiling problem into a reinforcement learning framework, where an agent is a next-visit planner, the state of environment is the fused representation of users and spatial KG, an action is a POI visited by a user, and the reward of an action is how well the agent can imitate mobile users. 
(3) We develop a new reward function to evaluate the quality of actions.
We extend the reward function in~\cite{wang2020incremental} to be negative when the predicted POI visits are far away from the real, which would positively encourage the agent on good actions and punish the agent on bad actions.
(4) We extend the policy network in~\cite{wang2020incremental} based on Double DQN with hierarchical graph pooling, which avoids the overestimation of specific POI visits.
(5) We identify and describe the semantic connectivity of spatial entities by a spatial KG, which is integrated into the reinforcement learning framework. 
(6) We develop a new sequentially nested state update strategy by modeling the long-short term effects of interactions between  users and spatial KG, by paying more attention to recent ones.
(7) We present extensive experimental results with real-world mobile check-in data to demonstrate improved performances. 
(8) Our framework can be adapted and generalized to the tasks of incremental learning with mixture event streams to support mobile user profiling and other applications.

\begin{table}[!t]
  \caption{Summary of Notations.} 
  \begin{center}
    \begin{tabularx}{\linewidth}{c|X}
     \hline
      \textbf{Symbol} & \textbf{Definition} \\ \hline
    $u, \mathbf{u}^{(\cdot)}$ & User, user state (representations). \\ \hline
    $c_{(\cdot)}$ & POI category. \\ \hline
    $a_{(\cdot)}$ & Action. \\ \hline
    $P_{(\cdot)}, \mathbf{h}_{P_{(\cdot)}}$ & POI, POI representations. \\ \hline
    $\mathbf{rel}_{(\cdot)}$ & Relation representations. \\ \hline
    $\mathbf{t}_{(\cdot)}$ & Tail (categories, locations) representations.  \\ \hline
    $\mathbf{g}^{(\cdot)} = <\mathbf{h}_{P_{(\cdot)}}, \mathbf{r}_{(\cdot)}, \mathbf{t}_{(\cdot)}>$ & Spatial {\it KG} state (representations). \\ \hline
    $s^{(\cdot)}=(\mathbf{u}^{(\cdot)}, \mathbf{g}^{(\cdot)})$ & State. \\ \hline
    $r_{(\cdot)}$ & Reward. \\ \hline
    $Q$ & Policy. \\ \hline
    $x_{(\cdot)}$ & Priority score. \\ \hline
    $\mathbf{T}^{(\cdot)}$, $\tilde{\mathbf{T}}^{(\cdot)}$ & Temporal context, transformed. \\ \hline
    $\mathbf{W}_{(\cdot)}, \mathbf{b}_{(\cdot)}$ & Weights, biases of model. \\ \hline
    $\lambda_{(\cdot)}$ & Weights of rewards.
    \\ \hline
    \end{tabularx}
  \end{center}
  \label{tab:notation-def}
\end{table}
\section{Preliminaries}
\subsection{Definitions and Problem Statement}
We first introduce the key definitions and the problem statement. Then, we present an overview of the proposed framework, followed by the discussion of how our approach differs from those presented in the current literature. All the notations are summarized in Table~\ref{tab:notation-def}.

\begin{figure*}[!t]
	\centering
	\includegraphics[width=0.8\linewidth]{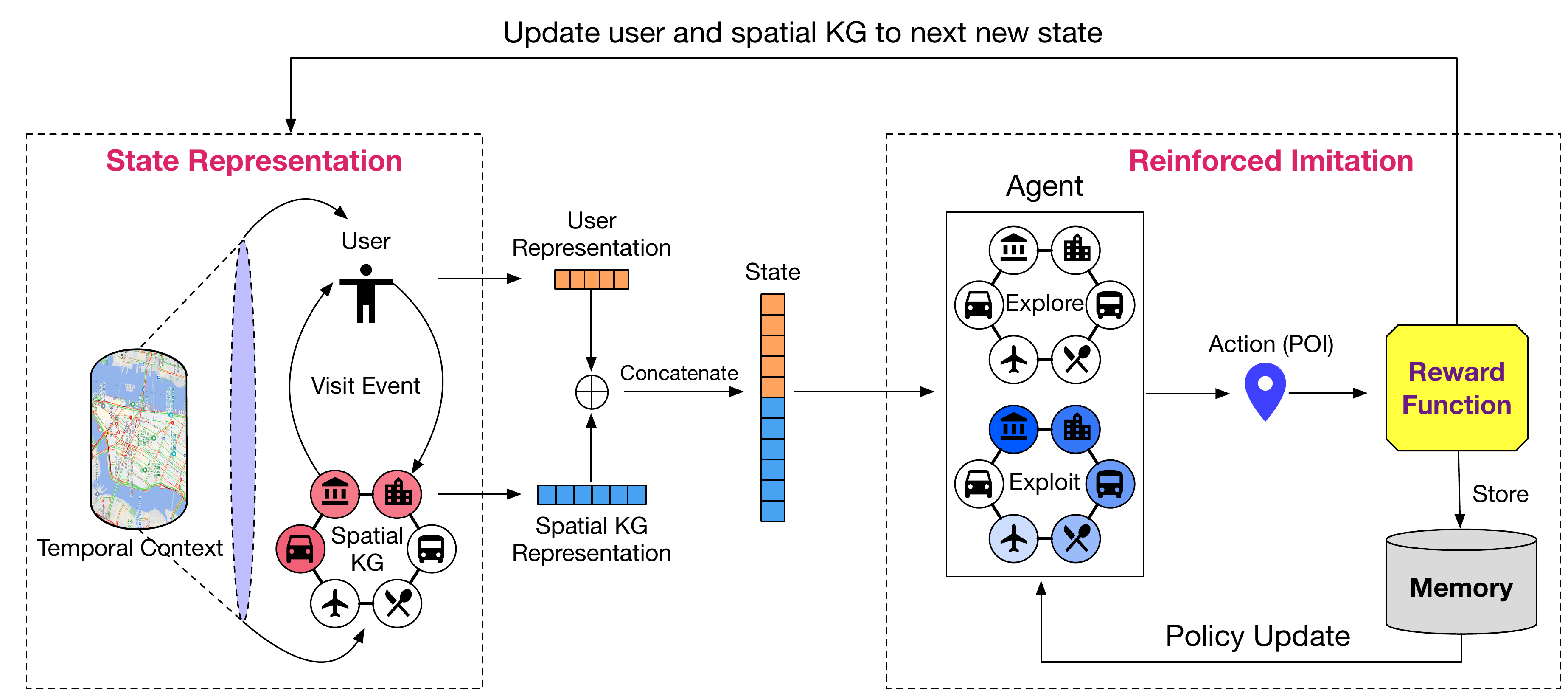}
	\captionsetup{justification=centering}
	\caption{Framework Overview.}
	\label{fig:framework overview}
\end{figure*}

\subsubsection{{\bf Spatial Knowledge Graph ({\it KG})}.}

We construct spatial {\it KG} to demonstrate semantic connectivities between spatial entities.
Specifically, in the spatial {\it KG}, there are three types of spatial entities: POIs, POI categories and locations ({\it i.e.} functional zones); and two types of relations: ``belong to'' which is to demonstrate the affiliation relations between POIs and POI categories, and ``locate at'' which is to demonstrate the geospatial relation between POIs and the functional zones.
The spatial {\it KG} is defined as the following groups of triplet facts: (1) $<$POI, ``belong to'', POI category$>$, and (2)$<$POI, ``locate at'', functional zones$>$.

\subsubsection{{\bf  Temporal Context.}}

In this paper, temporal context quantifies the temporal factors when users make the decision to visit a POI.
Following the approach in previous studies~\cite{liu2018modeling}, we utilize the snapshot of transportation traffic in a small time window for quantifying temporal context.
Specifically, we first segment the entire area into $m$ grids.
For a given grid, we calculate the inner traffic, in-flow traffic and out-flow traffic.
Then, we obtain a traffic matrix $\mathbf{T} \in \mathbb{R}^{m\times 3}$, where each row denotes a grid, each column denotes the inner traffic, in-flow traffic and out-flow traffic of the given grid respectively.
We use the traffic matrix $\mathbf{T}$ to represent the temporal context.

\subsubsection{{\bf Key Components of Reinforcement Learning.}}

In our problem setting, we define the key components of our reinforcement learning framework as follows:
\begin{enumerate}
\item {\bf Agent.} We consider the next-visit
planner as the agent. The agent provides the personalized POI prediction based on the current environment status. 
\item {\bf Actions.} Actions are defined as the visit event such that at each step, the user will visit which POI. Formally, let $a_j$ denote the action that visits the POI $P_j$. The action space is the number of POIs. Suppose the user visits the POI $P_j$ at the step $l$, then the policy would take the action $a^{l}=a_j$.

\item {\bf Environment.}  The environment is defined as the combination of all users and the spatial {\it KG}.
Within the environment, users would interact with spatial {\it KG} by visiting any POIs in the spatial {\it KG}.
Thus, on one hand, the visit behavior of users would affect the representation of the spatial {\it KG}; on the other hand, the spatial  {\it KG} inversely would affects each user's representation as well.

\item {\bf State.} The state $s$ is to describe the environment composed by users and the spatial {\it KG}. At the step $l$, the state $s^{l}$ is defined as a pair of  $(\mathbf{u}^l, \mathbf{g}^l)$. 
Specifically, $\mathbf{u}^l = \{\mathbf{u}_i^l | u_i \in \mathcal{U}\}$, where $\mathbf{u}_i^l$ denotes the representation of user $u_i$ at step $l$; 
$\mathbf{g}^l = <\mathbf{h}^l, \mathbf{rel}, \mathbf{t}^l>$, where $\mathbf{h}^l$ denotes the heads ({\it i.e.}, POIs) and $\mathbf{t}^{l}$ denotes the tails ({\it i.e.}, categories and functional zones).
\item {\bf Reward.} 
In the conference version~\cite{wang2020incremental}, we propose the reward function as the weighted summation of reciprocal of the distance $r_d$, the category similarity $r_c$, and binary accuracy between the real and predicted the POI visit $r_p$.
The range of such reward function is above zero, which means the agent would be always encouraged positively no matter the generated action is good or bad.
However, intuitively, the agent should be encouraged positively for the good actions, but be punished for the bad actions.
Based on such intuition, we propose a new formulation for the reward function, which would assign
positive values for good actions as the encouragement, and negative values for bad actions as the penalty.
Formally, the new reward function can be denoted as 
    
    \begin{equation}
         r = \lambda_d \times (r_d - b_d) + \lambda_c \times (r_c-b_c) + \lambda_p \times (r_p-b_p),
    \end{equation}
    where $b_d$, $b_c$ and $b_p$ are the reward baselines of $r_d$, $r_c$ and $r_p$ respectively.
    We set the reward baselines as follows: 
    (1) we first run the experiment for 100 rounds, and obtain the empirical range of the reward values;
    (2) we then set  the value of $b_d$, $b_c$ and $b_p$ as the first quartile of the corresponding empirical reward range.
    So if the agent behaves better than baselines, the reward is positive, otherwise, it is negative. 
    This reward function urges the agent to find the next POI that is close to the real visited POI from POI category and location.

Specifically, we calculate the euclidean distance between the real and predicted POI according to their longitude and latitude coordinates, and regard the reciprocal value of the distance as $r_d$.
We exploit {\it GloVe}\footnote{\url{https://nlp.stanford.edu/projects/glove/}} pre-trained word vectors of the category names to calculate the cosine similarity between the predicted and real POI categories as the similarity $r_c$.
\end{enumerate}

\subsubsection{{\bf Problem Statement.}}

In this paper, we study the problem of learning to profile users from user mobile activity data.
Due to the large-scale, nested, sequential and semantic nature of user mobile activities, we reformulate the mobile user profiling problem as an incremental user modeling with the integration of reinforcement learning and spatial {\it KG}.

Formally, given a mobile activity sequence of mixed-user and spatial {\it KG}, we aim to find a mapping function $f: (\mathbf{u}^{l}, \mathbf{g}^{l}) \rightarrow (\mathbf{u}^{l+1}, \mathbf{g}^{l+1}$), that takes as input the state of the environment (representations of users and spatial {\it KG}) at the step $l$, and outputs the state at the next step $l+1$, while simultaneously following the imitation-based criteria to provide accurate personalized prediction, based on the incrementally updated user representations from the mutual interactions with spatial {\it KG}.

\subsection{Framework Overview}
Figure~\ref{fig:framework overview} shows an overview of our  framework that includes two key components:
(1) state representation, and 
(2) reinforced imitation.
For the state representation, we consider the mutual interactions between users and spatial {\it KG} given the temporal context. Specifically, at each step, the user representations are updated based on the influence from spatial {\it KG} given the temporal context; inversely, the spatial {\it KG} representations are updated based on the influence from the user given the temporal context.
For the reinforced imitation, we employ a reinforced agent to imitate the user mobility patterns via exploration and exploitation.
The prediction of the agent is evaluated by a reward function.
Following that, the user visit trajectories are stored in memory to offline train the agent for updating learned policies.
In addition, the user and spatial KG are changed to the next state for starting a new visit event.
When the reinforced imitation mimics user mobility perfectly, our framework outputs the most effective user profiles.

\noindent{\bf Comparison with literature.} 
Despite the promising results from prior studies on mobile user profiling, one major concern arises that most of these methods are trained on offline data without the ability of self-updating, which is essential for quantifying user' dynamic mobility behaviors changing over time.
Therefore, in our work, we embrace reinforcement learning for mobile user modeling. 
When exploiting reinforcement learning for user modeling, online recommender systems have achieved great performance by regarding the online behavior as a sequential decision-making process.
Different from the online user modeling, mobile user behavior ({\it e.g.}, POI visiting) is offline behavior which is constrained by many offline factors ({\it e.g.}, time, traffic, location).
Therefore, we incorporate temporal contexts and simultaneously update user and spatial {\it KG} representations based on the mutual interactions.
In the meantime, the integrated spatial {\it KG} provides rich semantics to better understand user patterns and preferences.

\section{Policy Design}
In this section, we will introduce how to teach the next-visit planner (agent) to mimic users' patterns and preferences, given state (representations of user and spatial {\it KG}) and temporal contexts.

\begin{figure}[!t]
	\centering
	\includegraphics[width=0.9\linewidth]{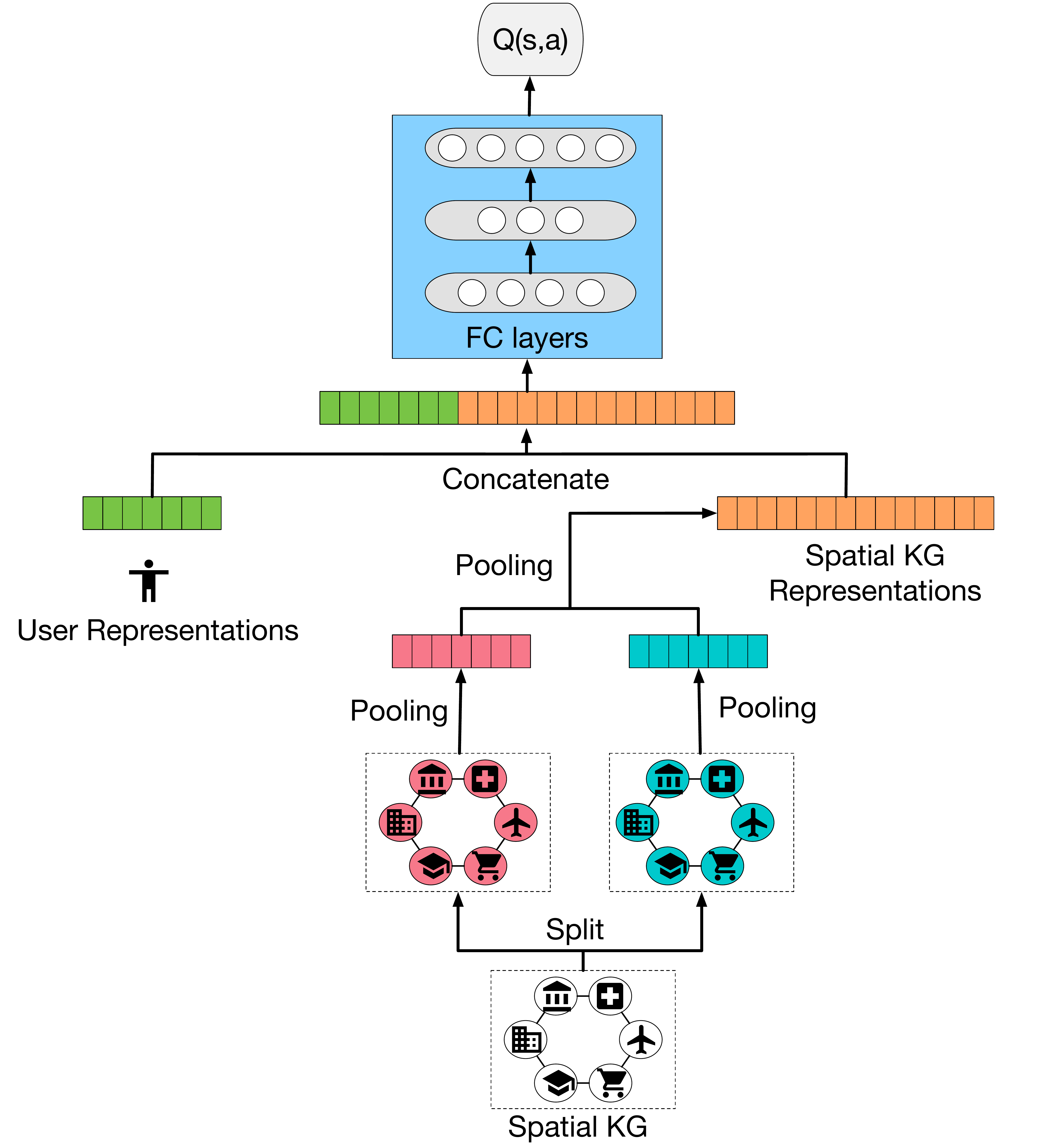}
	\captionsetup{justification=centering}
	\caption{DQN Network structure.}
	\label{fig:q-netwrok}
\end{figure}

\subsection{Network Structure}


While DQN shows exceptional performance on learning strategies~\cite{wang2020incremental}, the maximization step of the action value estimation process leads the  DQN to overestimate Q-value, resulting in learning local optimal policies.  
In order to alleviate this issue, we replace the DQN to the  Double Deep Q-Network (DDQN) ~\cite{van2016deep}.

\begin{figure}[!t]
	\centering
	\includegraphics[width=0.9\linewidth]{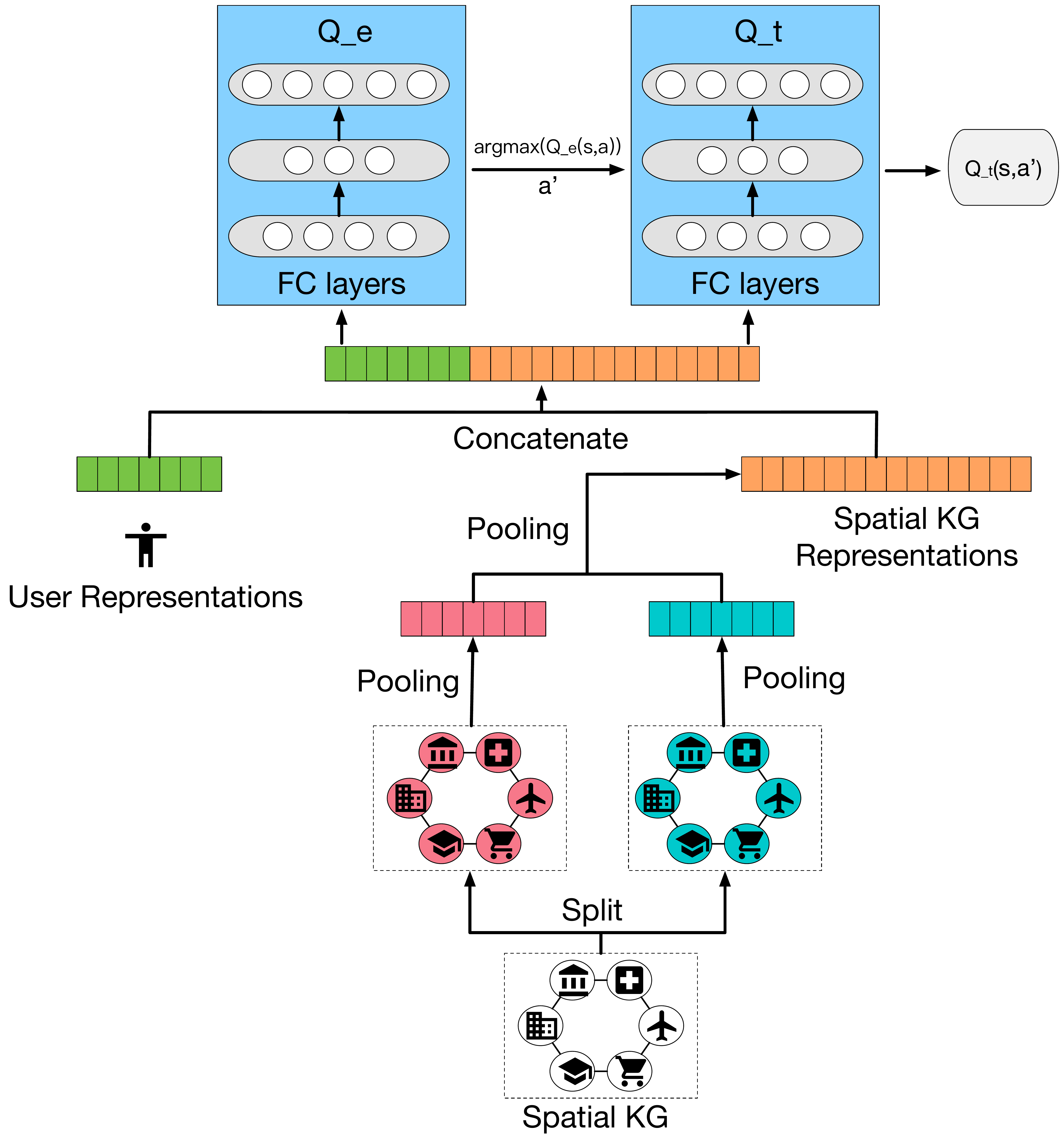}
	\captionsetup{justification=centering}
	\caption{Double-DQN Network structure.}
	\vspace{-0.5cm}
	\label{fig:ddqn-netwrok}
\end{figure}

Figure \ref{fig:ddqn-netwrok} shows the structure of DDQN.
Compared to the model structure (Figure \ref{fig:q-netwrok}) of the conference version ~\cite{wang2020incremental}, we implement two neural networks ($Q\_e$ and $Q\_t$) to decouple the maximization operation in DQN into action selection and action evaluation.
Specifically, we first concatenate the user representations and spatial KG representations together as the state vector.
Then, we input the state into the $Q\_e$ network to select the action (POI) with the highest Q-value.
Moreover, we input the state and the previously selected action into the $Q\_t$ network to estimate the real value of the action
in order to update learned policies.
During this process, $Q\_e$ and $Q\_t$ own the same model structure and the same initialization.
But we first update the parameters of $Q\_e$ for several steps and freeze the parameters of $Q\_t$.
After that, we store the parameters of $Q\_e$ into $Q\_t$ and reiterate the previous learning step.
Thus, $Q\_e$ and $Q\_t$ have different estimation abilities at the same time.
The max Q-value estimations in the two networks associated with different actions. 
So, the Q-value used to update policies is more rational than the Q-value in DQN.
Such a learning strategy has been proved effective in avoiding the over-optimistic problem of DQN and learning more effective policies ~\cite{van2016deep}.

Since DDQN only takes vectors/matrices as input, graph pooling is desired for graph-structure state $\mathbf{G}$ transformed into vectors.
But current graph pooling operations do not fit for the heterogeneity of spatial {\it KG}.
Therefore, we design the graph pooling operation in a hierarchical fashion.
We first split the spatial {\it KG} into two graphs based on the two types of relations as (1) ``location'' graph and (2) ``belonging'' graph.
Then, we leverage graph average pooling for both the ``location'' and ``belonging'' graph to generate graph-level vectorized representations respectively.
Finally, we employ average pooling over the two graph-level vectors to obtain the single unified vectorized representation $\mathbf{g}^{l}$ for the spatial {\it KG} at step $l$.

After the hierarchical pooling module, we concatenate the user state $\mathbf{u}_i^{l}$ with the spatial {\it KG} state $\mathbf{g}^{l}$ as the input of the fully connected ({\it FC}) layers.
Then, the {\it FC} layers would map the given state $\mathbf{s}^l$ into a group of $Q(s^l, a)$ for each POI in the spatial {\it KG}.
The policy chooses the POI with the highest $Q(s^l, a)$ as the prediction result.

\subsection{Improved Sampling Strategy in Experience Replay}
Since the space of POIs visiting events is very large, the proposed reinforcement learning framework is computationally intensive.
Therefore, we propose a training strategy to accelerate exploration procedure based on experience replay~\cite{lin1993reinforcement}.
The training strategy involves two stages: (1) priority assignment, which assigns the priority score for each data sample $(s^{l}, a^{l}, r^{l}, s^{l+1})$; and (2) sampling strategy, which selects data samples from memory for training.

\subsubsection{Priority}
We design two types of priority scores, including reward-based and temporal difference-based.

\begin{enumerate}
\item Reward-based. Intuitively, the higher the reward of the POI visit action $a^{l}$ is, the better the next-visit planner (agent) mimic the users, then the more significant the data sample can contribute to the policy training.
Therefore, formally, for each data sample $(s^{l}, a^{l}, r^{l}, s^{l+1})$, the reward-based priority score $x_r$ is defined as the reward $r^{l}$:
\begin{equation}
    x_r(s^{l}, a^{l}, r^{l}, s^{l+1})=r^{l}.
\end{equation}

\item Temporal difference ({\it TD})-based. The {\it TD} error is originally set for updating the {\it DQN}. The larger the {\it TD} error is, the more valuable and informative is the data sample for the next-visit planner (agent) to learn. Therefore, we define the {\it TD} error as the {\it TD}-based priority score that
\begin{footnotesize}
\begin{equation}
\vspace{-0.1cm}
    x_{TD}(s^{l}, a^{l}, r^{l}, s^{l+1})=r^{l}+\gamma \max_{a^{l+1}}Q(s^{l+1}, a^{l+1}) - Q(s^{l}, a^{l}),
\end{equation}
\end{footnotesize}
where $\gamma$ is the discount factor.
\end{enumerate}

We will evaluate and discuss the performance of these two priority scores in the experiment.
\subsubsection{Sampling Strategy}
After we obtain the priority score $x_{\ast}$, we need to construct a distribution from the priority score $x_{\ast}$ for sampling data.
Therefore, we employ {\it softmax} to convert the priority score into a distribution 
\begin{equation}
    P(k) = \frac{e^{x^{(k)}}}{\sum \limits_{k^{'}=1}^{K} e^{x^{(k^{'})}}}, 
\end{equation}
where $P(k)$ is the sampling probability of the $k$-th data sample given the corresponding priority score $x^{(k)}$.

Then, we sample a batch size of data from the memory based on the assigned probability.


\section{State Representation Learning}
In this section, we introduce details about how to initialize and update states based on the mutual interactions between mobile users and spatial {\it KG}.
\subsection{State Initialization}
Here, we discuss the initialization of user state (representations), followed by how to initialize spatial {\it KG} state (representation).
\subsubsection{User}
We leverage {\it StructRL}~\cite{wang2019adversarial} to initialize the user state (representations).
{\it StructRL} is a mobile user profiling framework that models the global and substructure patterns of user behaviors by minimizing the structural loss between the input and reconstructed human mobility graph through adversarial learning.
For each user, we extract the very first portion ($10\%$) of user mobility trajectories to construct human mobility graph, where the nodes are POI categories and edges are visit transition frequencies or duration among POI categories.
Then, we employ {\it StructRL} to learn user representations.
The learned user representations are regarded as the initialized user state fed to the reinforcement learning framework.

\begin{figure}[!t]
	\centering
	\includegraphics[width=0.9\linewidth]{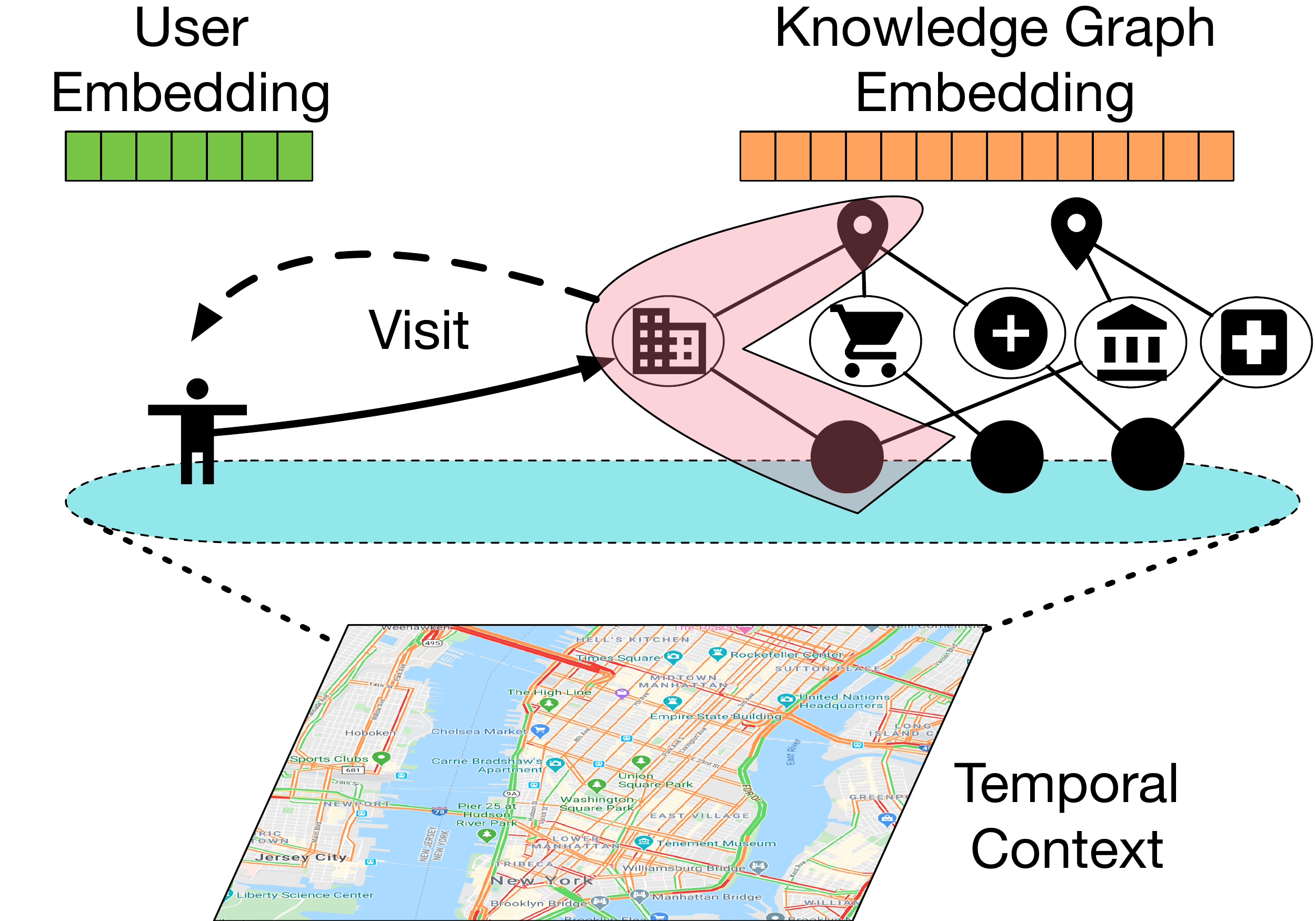}
	\captionsetup{justification=centering}
	\caption{An example of mutual interactions between users and spatial {\it KG}.}
	\vspace{-0.4cm}
	\label{fig:interaction}
\end{figure}

\vspace{-0.1cm}
\subsubsection{Spatial {\it KG}}
We exploit {\it TransD}~\cite{ji2015knowledge} to initialize spatial {\it KG} state (representations).
{\it TransD} is a translation-based model that utilizes two vectors to learn entity/relation representations and construct mapping matrix dynamically with considering both the diversity of entities and relations, which outperforms {\it TransE}~\cite{bordes2013translating}, {\it TransH}~\cite{wang2014knowledge}, and {\it TransR}~\cite{lin2015learning} in learning {\it KG} representations.
Specifically, to be more compatible with reinforcement learning framework, we adopt the projected entity embedding as the initialization of spatial {\it KG}.
In this way, the representations of entity and relations will be in the same feature space, which will make the state updating more feasible.

\subsection{State Update}
In this section, we will introduce how to update states based on the mutual interactions between user and spatial {\it KG}. 
Figure~\ref{fig:interaction} shows an interaction example in which a user visits a POI in spatial {\it KG}
During the interaction, the influence from the user visit is directly injected into the visited POI.
Then, the influence is propagated to other spatial entities through the semantics and topology of spatial {\it KG}.
The influence from spatial {\it KG} then inversely affects the user state (representations). 
Without the loss of the generality, we consider the scenario of a POI visit event where a user $u_i$ visits the POI $P_j$ at step $l$. The states will update for the step $l+1$.

In the conference version~\cite{wang2020incremental}, we propose a state update strategy that updates the states incrementally by equally treating the impacts of each visit.
However, intuitively, the impact of each visit should be different, so the most recent visits should be given more attention than previous ones.
To solve the problem, we follow the idea from the Long Short Term Memory (LSTM) networks, in which the model should forget partial old information and combine new changes simultaneously. 
Based on this rule, we propose the new state update strategy as follows.

\begin{enumerate}[(1)]
    \item {\bf User}
    
    We update the user state $\mathbf{u}_i^{l+1}$ based on old user state $\mathbf{u}_i^{l}$ and  the interaction between $\mathbf{u}_i^{l}$  and $\mathbf{h}_{P_j}^{l}$ in a temporal context.
    The update process can be represented as:
    \begin{equation}
        \mathbf{u}_i^{l+1} = \alpha_u \times \mathbf{u}_{i}^l + (1-\alpha_u) \times (\mathbf{W}_u \cdot (\mathbf{h}_{P_j}^{l})^{\intercal} \cdot  \tilde{\mathbf{T}}^l)),
    \end{equation}
    where $\alpha_u$ is a scalar that indicates the ratio of information of $\mathbf{u}_i^l$ in $\mathbf{u}_i^{l+1}$ .
    At this point, $\alpha_u$ is calculated by $\mathbf{u}_i^l$ and the calculation process can be represented as:
    \begin{equation}
        \alpha_u = \sigma(\mathbf{W}_{\alpha_u}\cdot \mathbf{u}_i^l + \mathbf{b}_{\alpha_u}),
    \end{equation}
    where $\mathbf{W}_{\alpha_u} \in \mathbb{R}^{1\times N}$ is weight, $N$ is the dimension of the weight matrix, and $\mathbf{b}_{\alpha_u} \in \mathbb{R}^{1\times1}$ is bias. 
    \item {\bf Spatial \it KG}
    \begin{enumerate}
        \item Updating visited POI $\mathbf{h}_{P_j}^{l+1}$
        
        Similar to the update process of $\mathbf{u}_i^{l+1}$, the new POI state $\mathbf{h}_{P_j}^{l+1}$ is updated by the old POI state $\mathbf{h}_{P_j}^{l}$ and the interaction between $\mathbf{u}_i^{l}$ and $\mathbf{h}_{P_j}^{l}$ in a temporal context.
        The update process can be represented as:
        \begin{equation}
            \mathbf{h}_{P_j}^{l+1} = \alpha_{p} \times \mathbf{h}_{P_j}^{l} + (1-\alpha_{p}) \times ( \mathbf{W}_p \cdot (\mathbf{u}_{i}^{l})^{\intercal} \cdot  \tilde{\mathbf{T}}^l) ,
        \end{equation}
        where $\alpha_p$ is a scalar that represents the proportion of information of  $\mathbf{h}_{P_j}^{l}$ in  $\mathbf{h}_{P_j}^{l+1}$.
        Here, $\alpha_p$ is calculated by $\mathbf{h}_{P_j}^{l}$ and the process can be represented as:
        \begin{equation}
        \alpha_p = \sigma(\mathbf{W}_{\alpha_p}\cdot \mathbf{h}_{P_j}^{l} + \mathbf{b}_{\alpha_p}),
        \end{equation}
        where $\mathbf{W}_{\alpha_p} \in \mathbb{R}^{1\times N}$ is  weight , $N$ is the dimension of the weight matrix, $\mathbf{b}_{\alpha_p} \in \mathbb{R}^{1 \times 1}$ is  bias.
        \item Updating category and functional zones (tail) $\mathbf{t}_{\ast}^{l+1}$ 
        
        We update tail $\mathbf{t}_{\ast}^{l+1}$ based on the $\mathbf{t}_{\ast}^{l}$ and the combination between $\mathbf{h}_{P_j}^{l+1}$ and $\mathbf{rel}_{(P_j,\cdot)}$.
        The update process can be represented as:
        \begin{equation}
     \mathbf{t}_{(P_j,\cdot)}^{l+1} = \alpha_t \times \mathbf{t}_{(P_j,\cdot)}^{l} + (1-\alpha_t) \times (\mathbf{h}_{P_j}^{l+1} + \mathbf{rel}_{(P_j,\cdot)}),
     \label{equ:kg}
        \end{equation}
        where $\alpha_t$ is a scalar that denotes the ratio of information of $\mathbf{t}_{(P_j,\cdot)}^{l}$ in $\mathbf{t}_{(P_j,\cdot)}^{l+1}$.
      Here, $\alpha_t$ is calculated by $\mathbf{t}_{(P_j,\cdot)}^{l}$, the process can be represented as:
        \begin{equation}
            \alpha_t = \sigma(\mathbf{W}_{\alpha_t}\cdot \mathbf{t}_{(P_j,\cdot)}^{l} + \mathbf{b}_{\alpha_t}),
        \end{equation}
        where $\mathbf{W}_{\alpha_t} \in \mathbb{R}^{1\times N}$ is  weight,  $N$ is the dimension of the weight matrix, $\mathbf{b}_{\alpha_t} \in \mathbb{R}^{1\times 1}$ is bias.
        \item Updating same category and location POIs $\mathbf{h}_{P_{j^{-}}}^{l+1}$
        
        We update the $\mathbf{h}_{P_{j-}}^{l+1}$ based on $\mathbf{h}_{P_{j-}}^{l}$ and the current state of $\mathbf{t}_{(P_{j^-},\cdot)}^{l}$ and $\mathbf{rel}_{(P_{j^-},\cdot)}$.
        The update process can be represented as:
        \begin{equation}
        \left\{
        \begin{array}{lr}
        \mathbf{h}_{P_{j^{-}}}^{l^n}
        = \mathbf{t}_{(P_{j^{-}},\cdot)}^{l+1} -  \mathbf{rel}_{(P_{j^{-}},\cdot)},
        \\
         \mathbf{h}_{P_{j^{-}}}^{l+1} = \alpha_{p^-} \times \mathbf{h}_{P_{j^{-}}}^{l} + (1-\alpha_{p^-}) \times \mathbf{h}_{P_{j^{-}}}^{l^n},
         \end{array}
         \right.
         \label{equ:other poi}
        \end{equation}
        where $\mathbf{h}_{P_{j^{-}}}^{l^n}$ represents the new state of $P_{j^{-}}$ that is calculated by $\mathbf{t}_{(P_{j^-},\cdot)}^{l}$ and $\mathbf{rel}_{(P_{j^-},\cdot)}$; 
        $\alpha_{p^-}$ indicates the ratio of information of $\mathbf{h}_{P_{j^{-}}}^{l}$
        in $\mathbf{h}_{P_{j^{-}}}^{l+1}$.
        Here, $\alpha_{p^-}$ is calculated by $\mathbf{h}_{P_{j^{-}}}^{l}$, the process can be represented as:
        \begin{equation}
            \alpha_{p^-} = \sigma(\mathbf{W}_{\alpha_{p^-}}\cdot \mathbf{h}_{P_{j^{-}}}^{l} + \mathbf{b}_{\alpha_{p^-}}),
        \end{equation}
        where $\mathbf{W}_{\alpha_{p^-}} \in \mathbb{R}^{1 \times N}$ is weight, $N$ is the dimension of the weight matrix, $\mathbf{b}_{\alpha_{p^-}}\in \mathbb{R}^{1\times 1}$ is  bias.
        The same process is used to update strategy1, here, we only update the local subgraph, which makes the update process efficiently when the spatial {\it KG} is large.
    \end{enumerate}
\end{enumerate}

\begin{figure*}[!thb]
	\centering
	\subfigure[Precision on Category]{\label{fig:substructure_precision_newyork}\includegraphics[width=4.35cm]{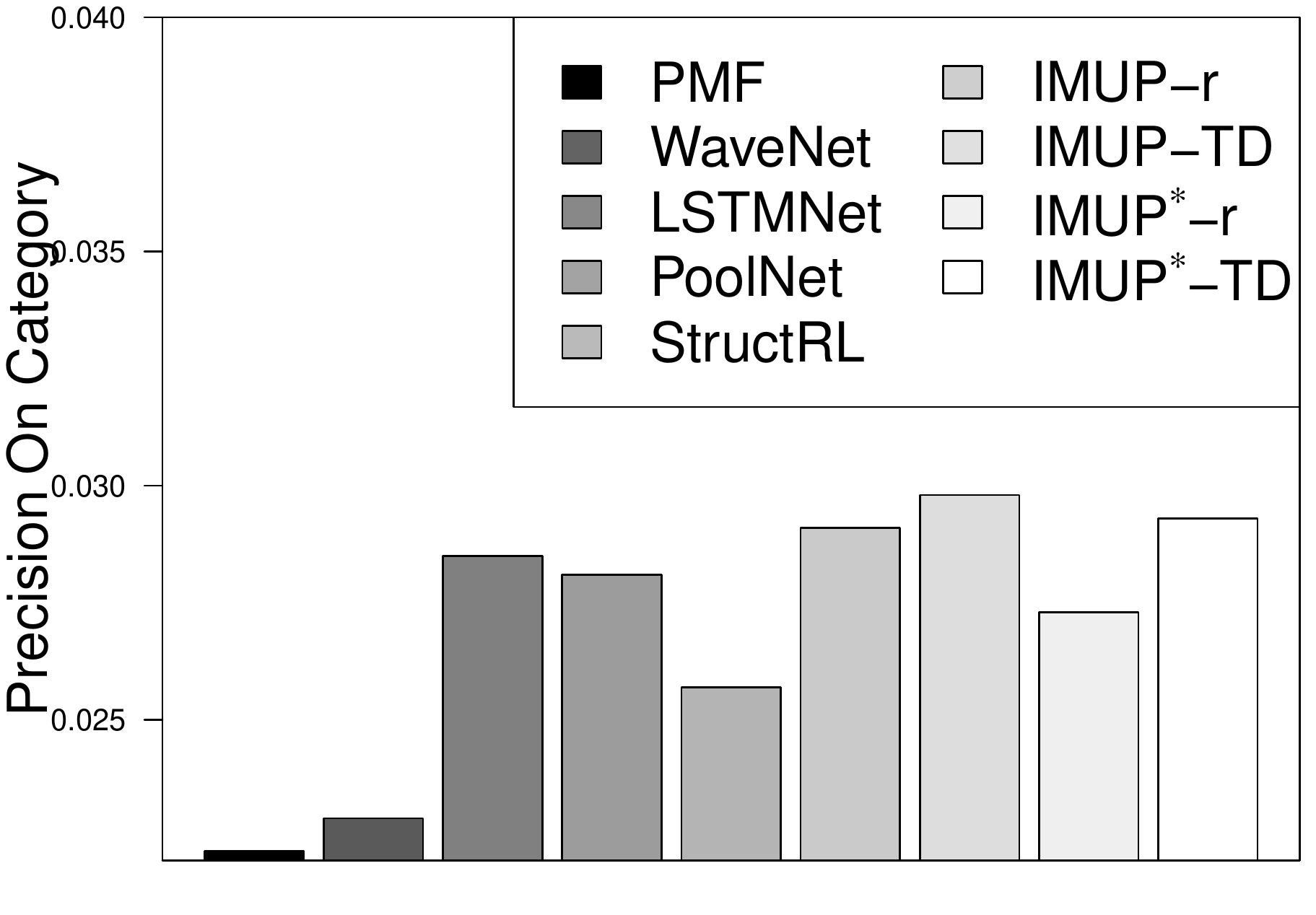}}
	\subfigure[Recall on Category]{\label{fig:substructure_newprecision_newyork}\includegraphics[width=4.35cm]{{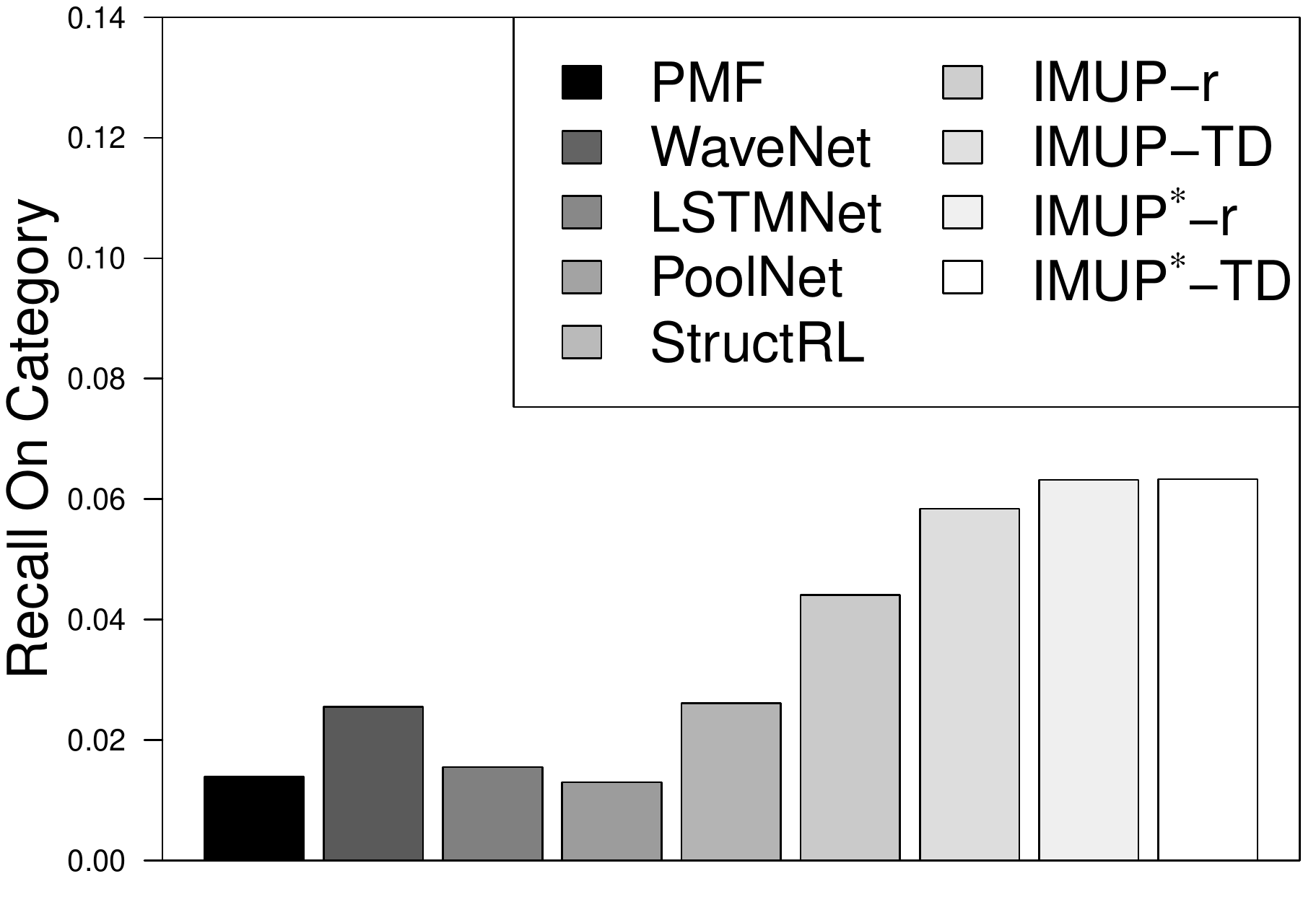}}}
	\subfigure[Average Similarity]{\label{fig:substructure_precision_tokyo}\includegraphics[width=4.35cm]{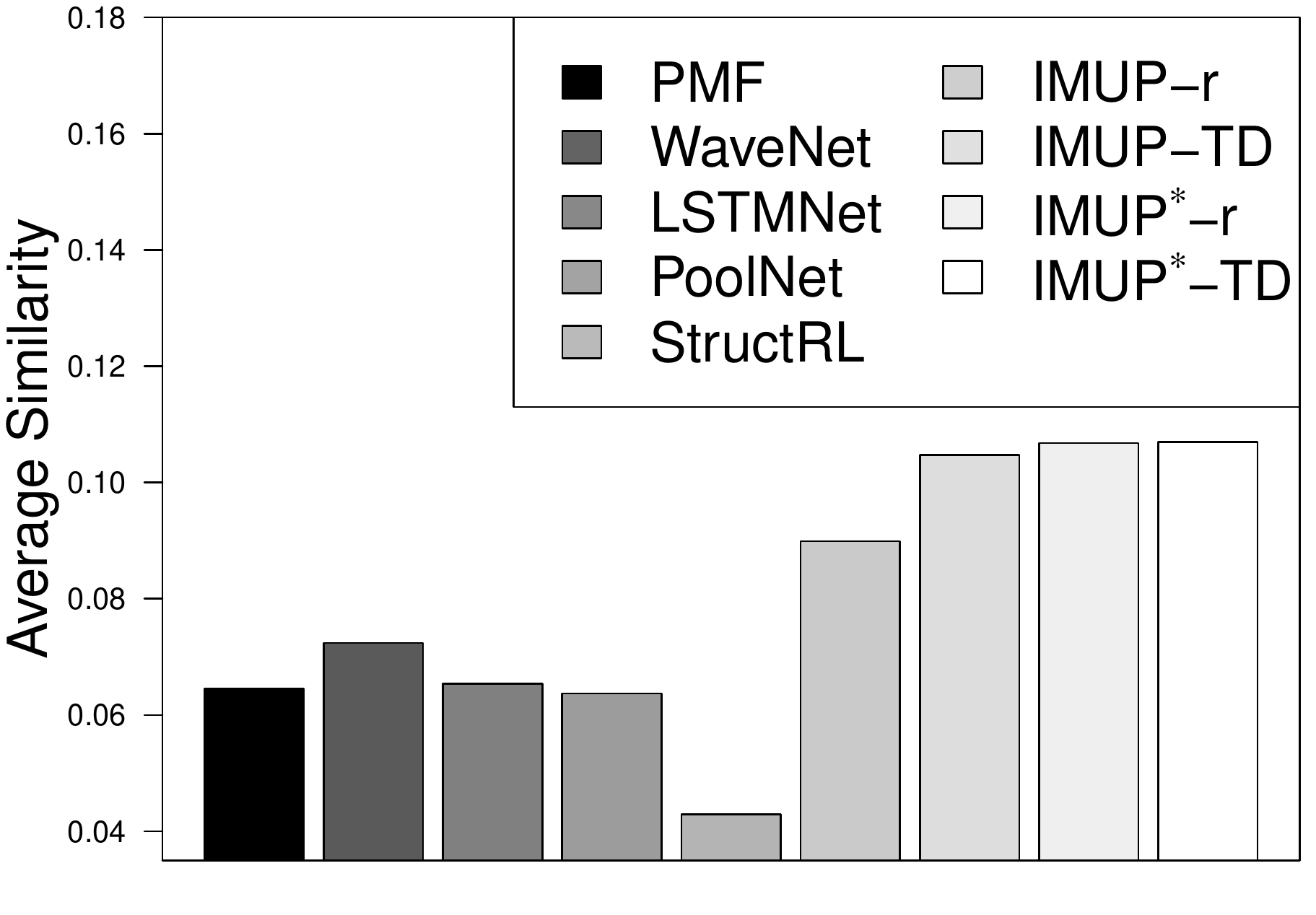}}
	\subfigure[Average Distance]{\label{fig:substructure_newprecision_tokyo}\includegraphics[width=4.35cm]{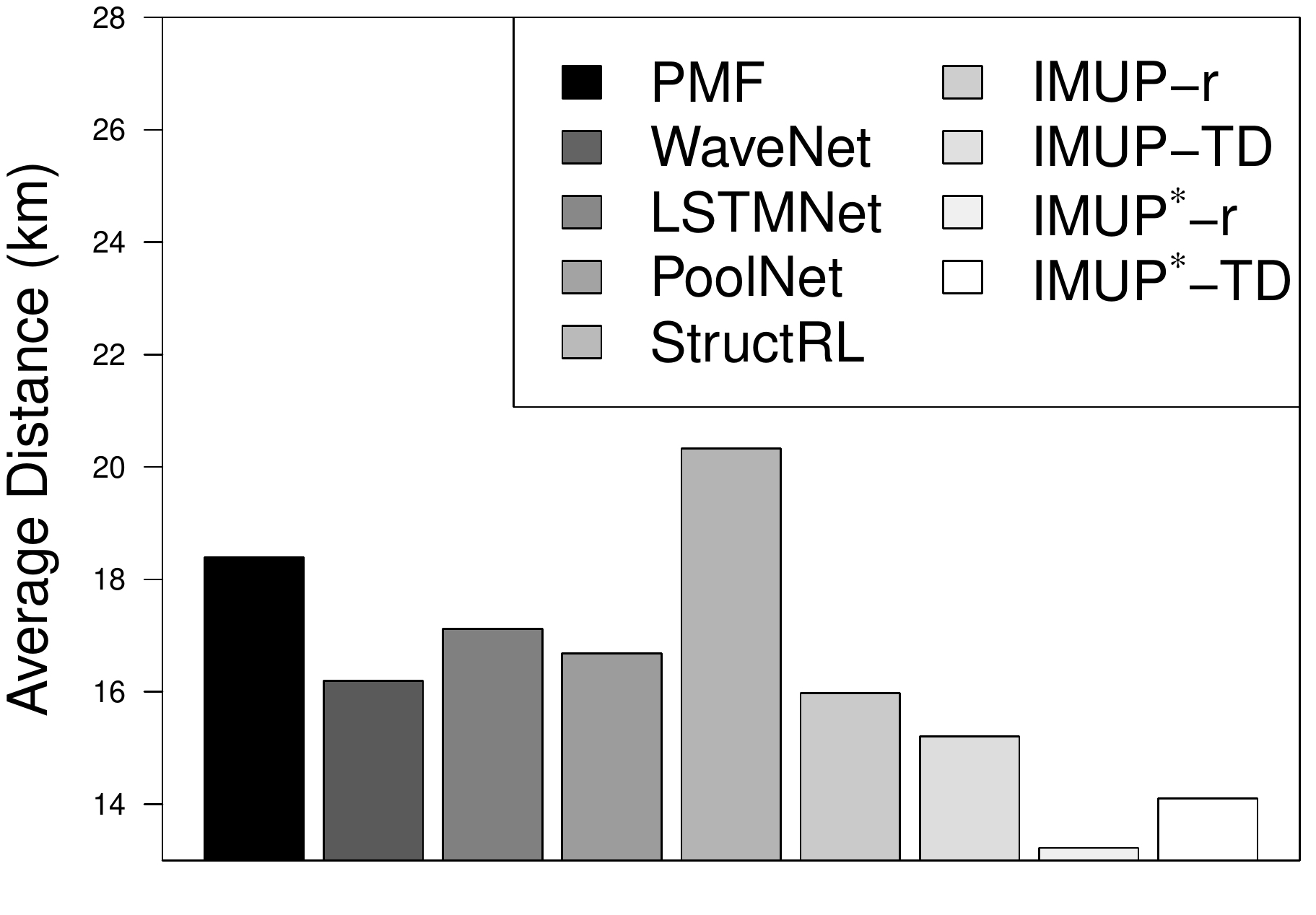}}
		\vspace{-0.4cm}
	\captionsetup{justification=centering}
	\caption{Overall comparison {\it w.r.t.} New York dataset.}
		\vspace{-0.5cm}
	\label{fig:nyc_overall}
\end{figure*}

\begin{figure*}[!tb]
	\centering
	\subfigure[Precision on Category]{\label{fig:substructure_precision_newyork}\includegraphics[width=4.35cm]{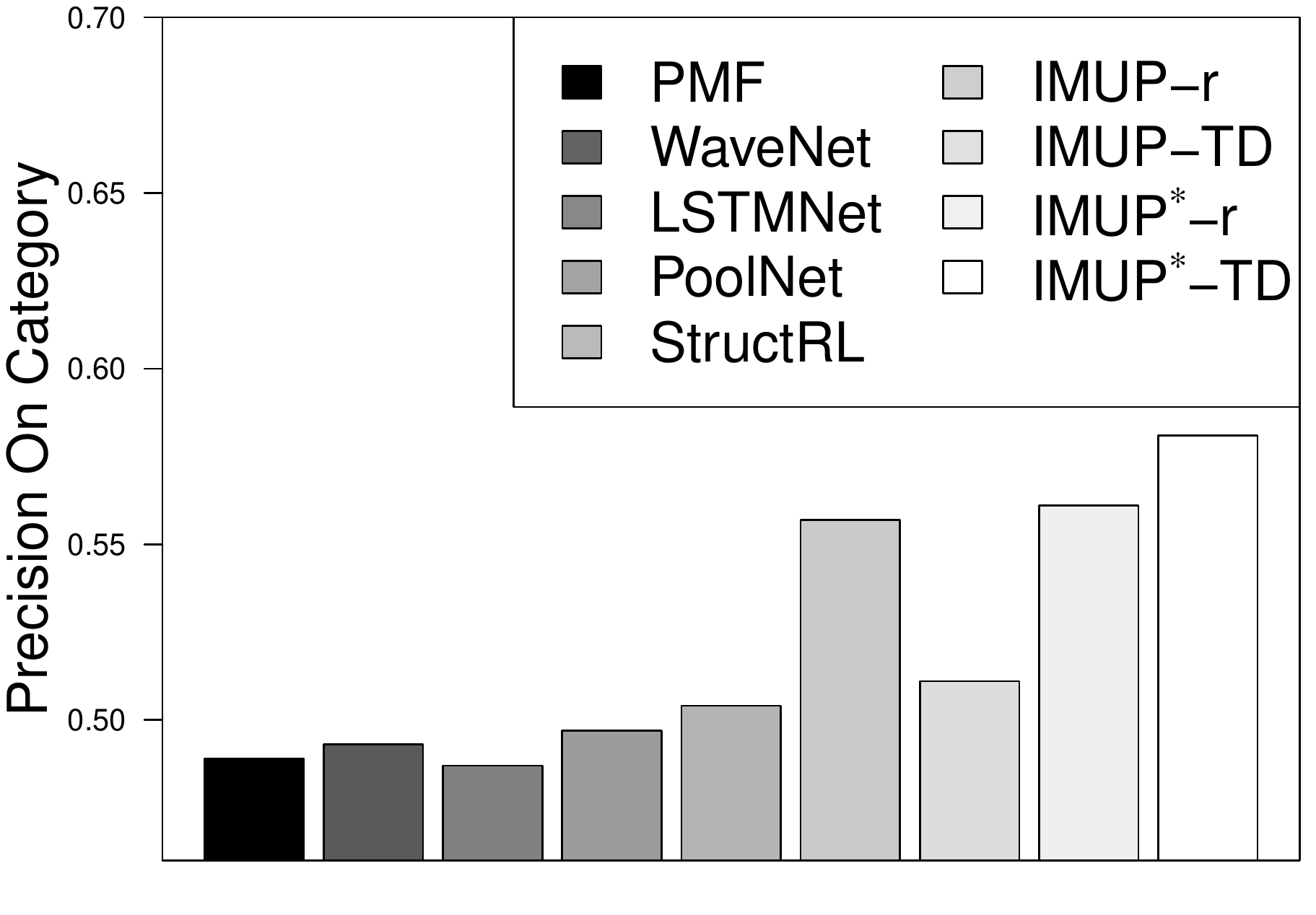}}
	\subfigure[Recall on Category]{\label{fig:substructure_newprecision_newyork}\includegraphics[width=4.35cm]{{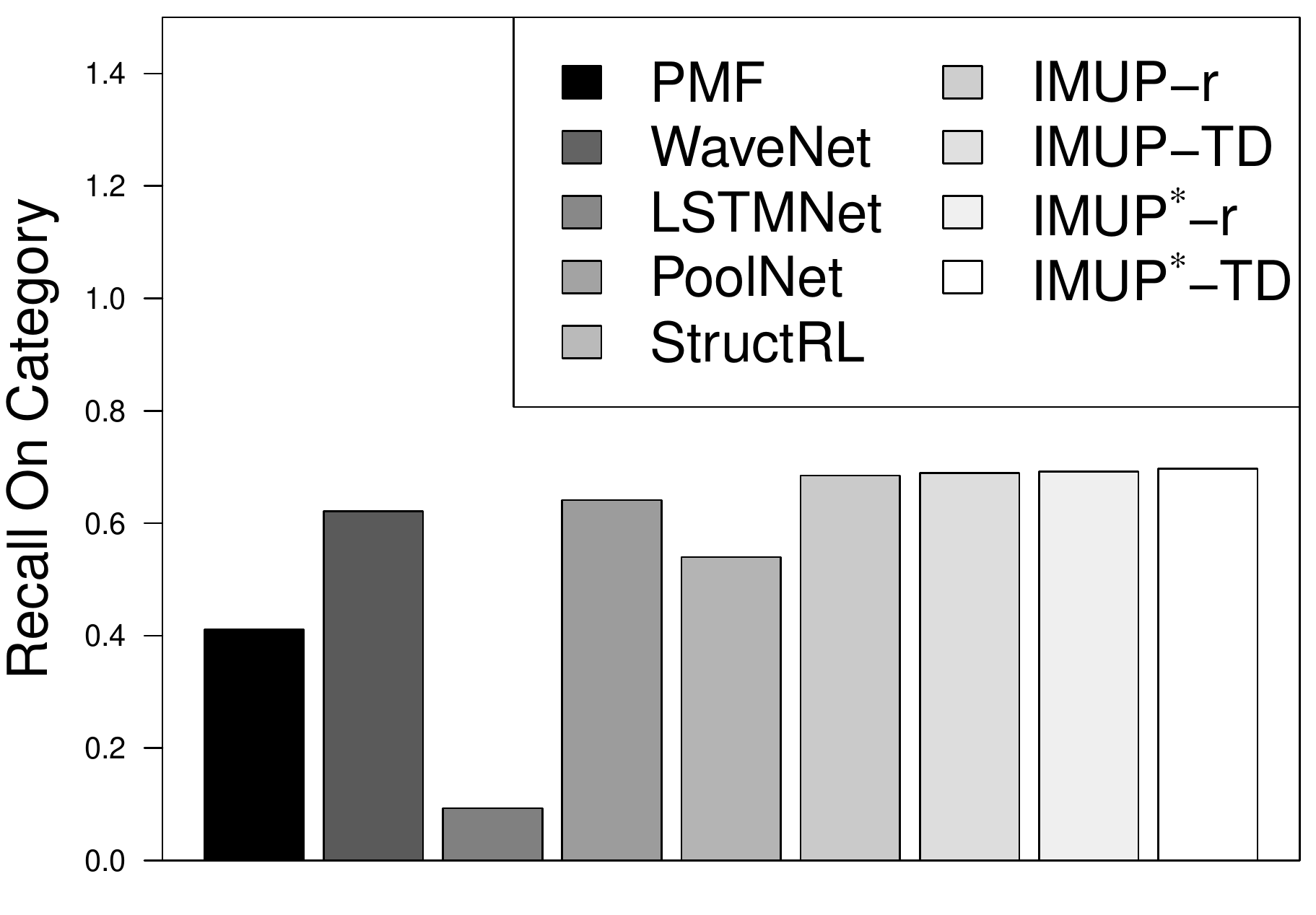}}}
	\subfigure[Average Similarity]{\label{fig:substructure_precision_tokyo}\includegraphics[width=4.35cm]{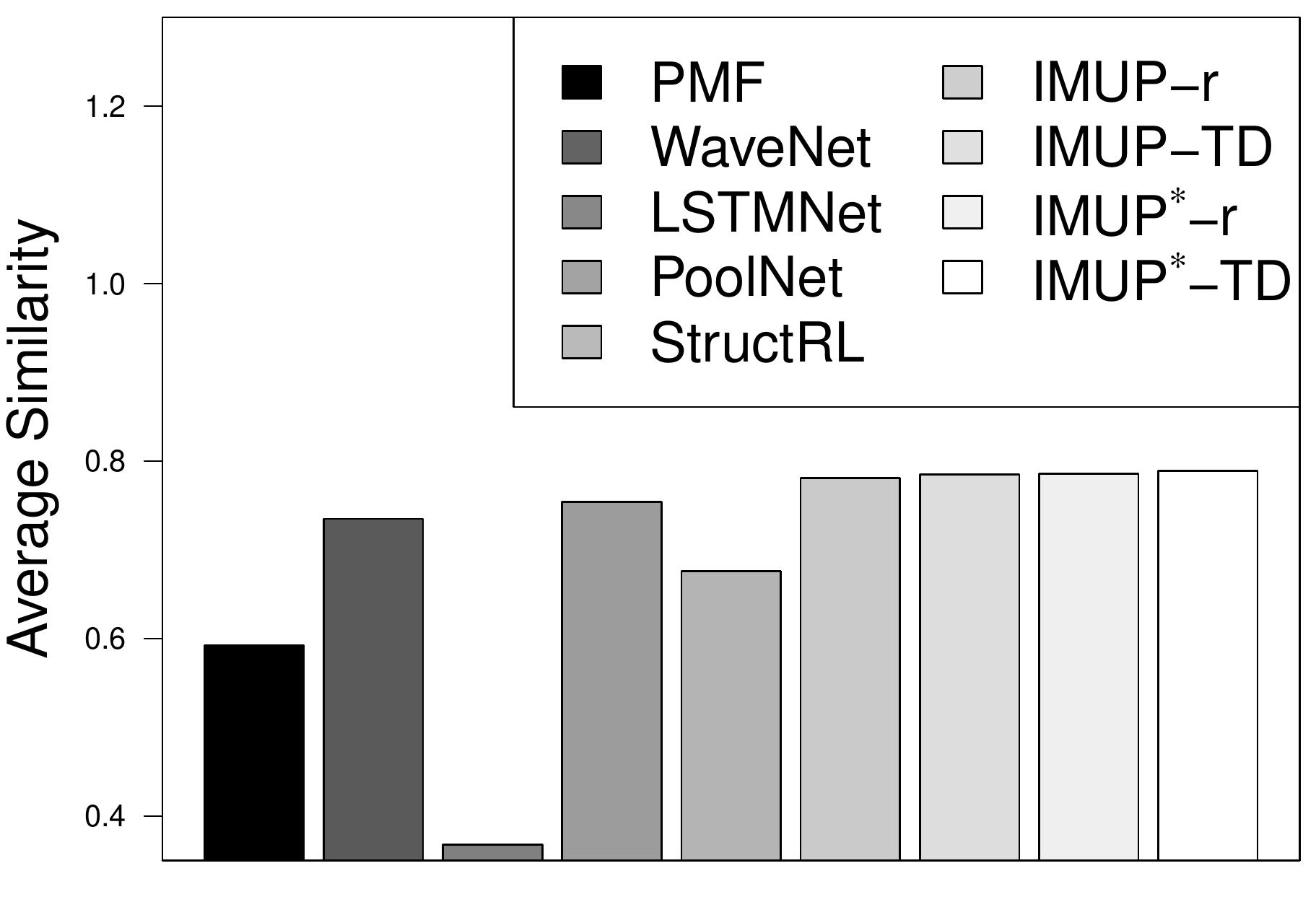}}
	\subfigure[Average Distance]{\label{fig:substructure_newprecision_tokyo}\includegraphics[width=4.35cm]{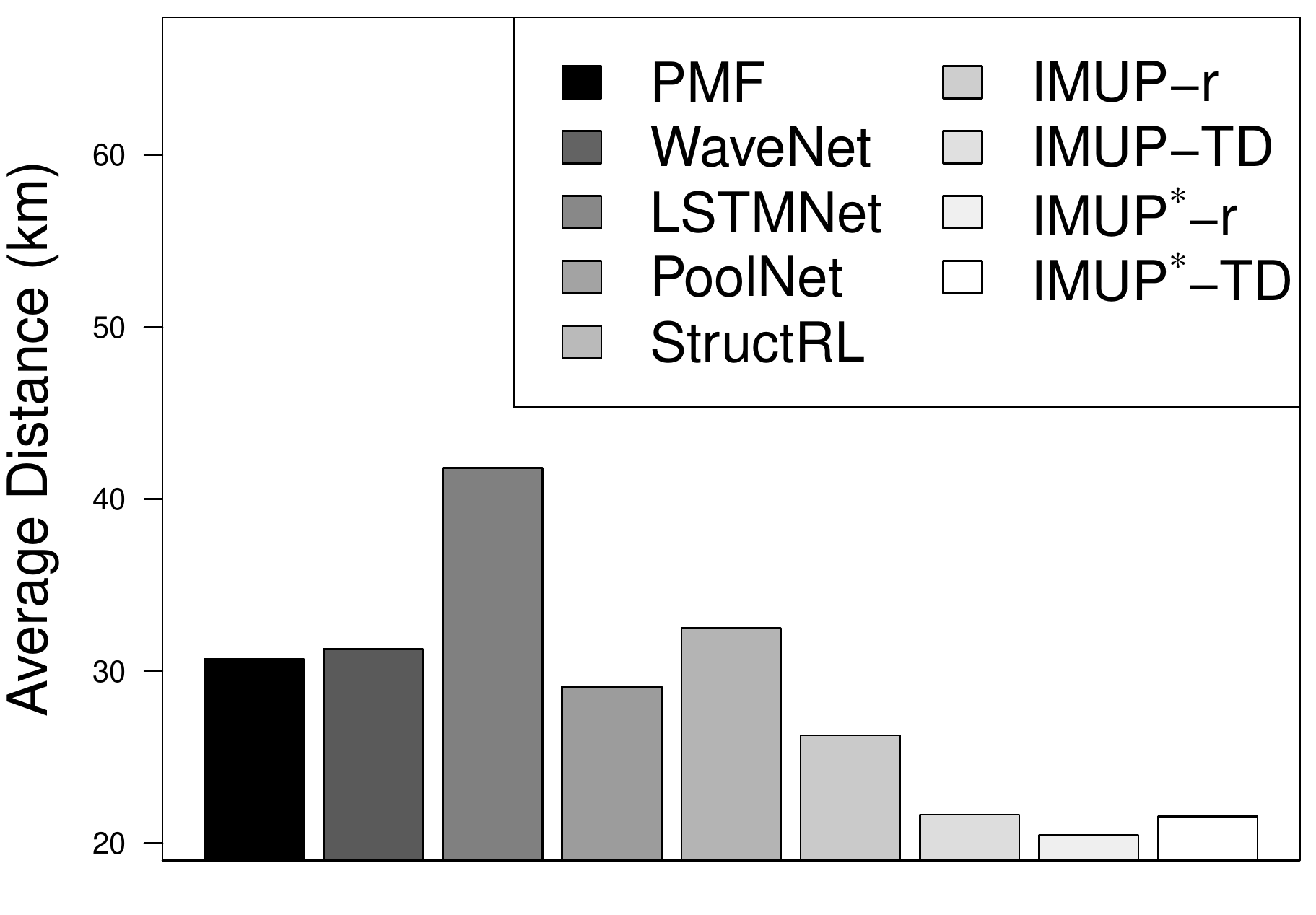}}
		\vspace{-0.4cm}
	\captionsetup{justification=centering}
	\caption{Overall comparison {\it w.r.t.} Beijing dataset.}
		\vspace{-0.45cm}
	\label{fig:bj_overall}
\end{figure*}

\section{Experiment}
This section details our empirical evaluation of the proposed method on real-world data.
\subsection{Data Description}

Table~\ref{table:data_stat_nyc} shows the statistics of our two check-in datasets from two cities: New York~\cite{yang2014modeling} and Beijing~\cite{wang2018ensemble}. Each dataset includes User ID, Venue ID, Venue Category ID, Venue Category Name, Latitude, Longitude, and Time.
Additionally, we also collected taxi data to represent traffic of the New York and Beijing in Table~\ref{table:data_stat_bj}. The taxis traffic data includes ID, Pick-up Latitude, Pick-up Longitude, Drop-off Latitude , Drop-off Longitude,
Pick-up time and Drop-off time. We split the taxi data into small segments with one-hour time window. For each time window, we calculate the temporal context $\mathbf{T}$, compatible with time users' visit events.

\begin{table}[t!hbp]
	\centering
	\scriptsize
	\tabcolsep 0.04in
	\caption {Statistics of the checkin data.}
	\begin{tabular}[t]{c|c|c|c|c}
		\hline
		\textbf{City}         & \textbf{\# Check-ins} & \textbf{\# POIs} &\textbf{\# POI Categories} & \textbf{Time Period}  \\ \hline
		New York & $227,428$ & $38,334$ &251 & 4/2012-2/2013 \\ \hline
		Beijing & $6,465$ & $3,434$ &9 & 3/2011-5/2011 \\
		\hline
	\end{tabular}
	\label{table:data_stat_nyc}
\end{table}

\begin{table}[t!hbp]
	\centering
	\scriptsize
	\tabcolsep 0.04in
	\caption {Statistics of the taxi data.}
	\begin{tabular}[t]{c|c|c }
		\hline
		\textbf{City}         & \textbf{\# Transactions} & \textbf{Time Period}  \\ \hline
		New York & $161,211,550$ & 4/2012-2/2013 \\ \hline
		Beijing & $12,000,000$ & 3/2011-5/2011 \\
		\hline
	\end{tabular}
	\label{table:data_stat_bj}
\end{table}

\subsection{Evaluation Metrics}
We evaluate the model performances over the prediction on the activity types ({\it i.e.}, POI categories) and locations in terms of the following four metrics: 

\noindent {\bf (1) $\text{Precision on Category (Prec\_Cat)}$:} We regard the prediction on activity type as the multi-class classification task. 
We evaluate the classification prediction via the weighted precision.
Let $c_k$ denote the $k$-th category, $|c_k|$ denote the number of activity types, $I_{TP}^k$ denote the number of true positive predictions, and $I_{FP}^k$ denote the number of false positive predictions, then the weighted precision on categories can be represented as
\begin{equation}
    \text{Prec\_Cat} = \frac{|c_k| \cdot I_{TP}^k}{\sum \limits_k |c_k| (I_{TP}^k+I_{FP}^k)}
\end{equation}

\noindent {\bf (2) $\text{Recall on Category (Rec\_Cat)}$:} Continuing with the definition of $\text{Prec\_Cat}$, we use the weighted recall to evaluate the recall on category prediction.
Let $I_{FN}^{k}$ denote the number of false negative prediction for the category $c_k$, the weighted recall can be represented as
\begin{equation}
    \text{Rec\_Cat} = \frac{|c_k| \cdot I_{TP}^k}{\sum \limits_k |c_k| (I_{TP}^k+I_{FN}^k)}
\end{equation}

\noindent {\bf (3) $\text{Average Similarity (Avg\_Sim)}$:} In addition to evaluating the precision and recall of prediction, we also evaluate the average similarity between the real and predicted POI categories.
We adopt the pre-trained {\it GloVe} word vectors to calculate the cosine similarity between the category word vector ``$\text{word}^l$'' of the real visited POI and the category word vectors ``$\hat{\text{word}}^{l}$'' of the predicted POI. Let $L$ denote the total visit number, then the average similarity on categories is
\begin{equation}
    \text{Avg\_Sim} = \frac{\sum \limits_{l} cosine(\text{word}^l, \hat{\text{word}}^{l})}{L}.
\end{equation}
The higher the value of $Avg\_Sim$, the better the prediction.

\begin{figure*}[!tb]
	\centering
	\subfigure[Precision on Category]{\label{fig:nyc_diff_reward_precision}\includegraphics[width=4.35cm]{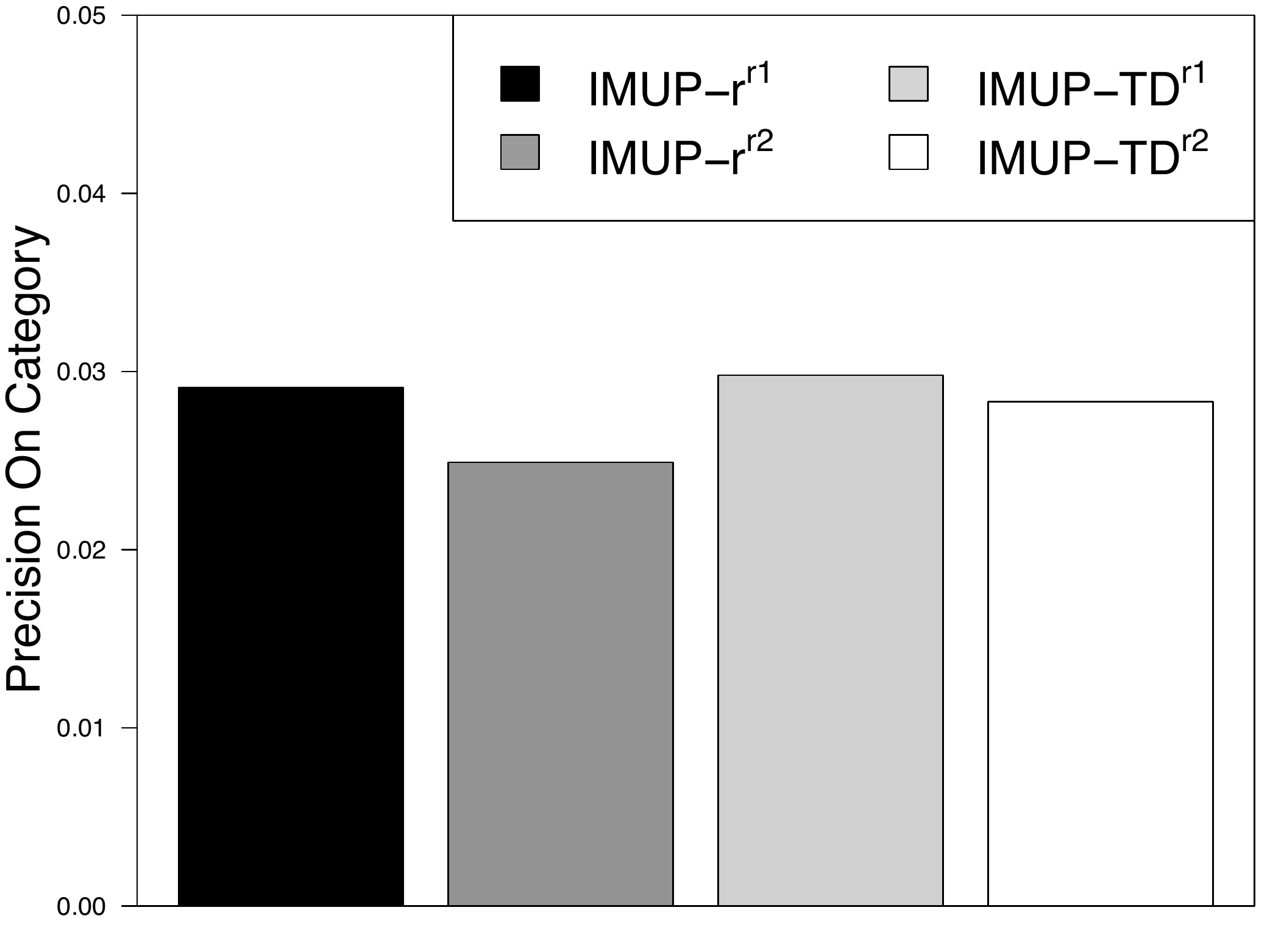}}
	\subfigure[Recall on Category]{\label{fig:nyc_diff_reward_recall}\includegraphics[width=4.35cm]{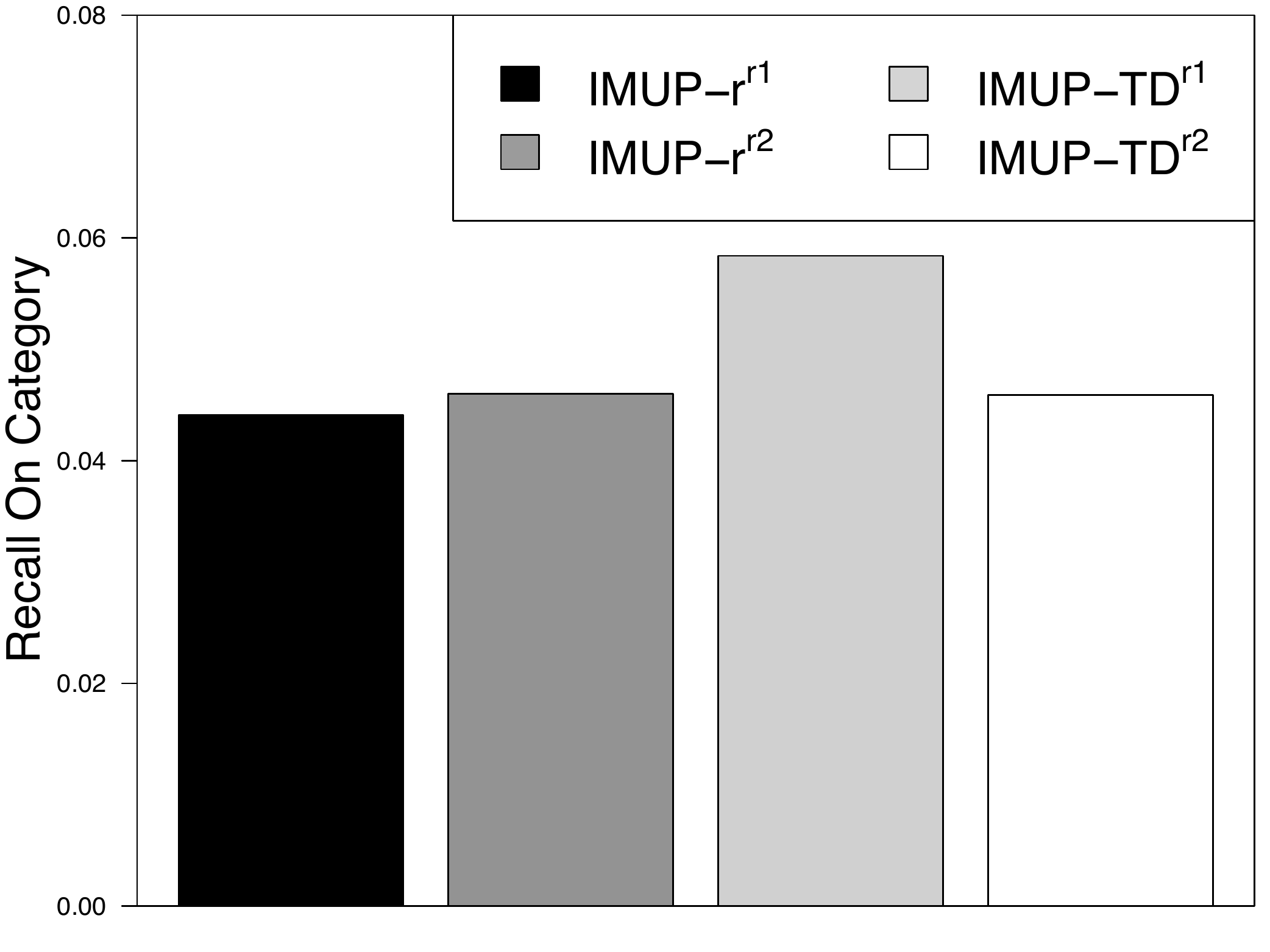}}
	\subfigure[Average Similarity]{\label{fig:nyc_diff_reward_sim}\includegraphics[width=4.35cm]{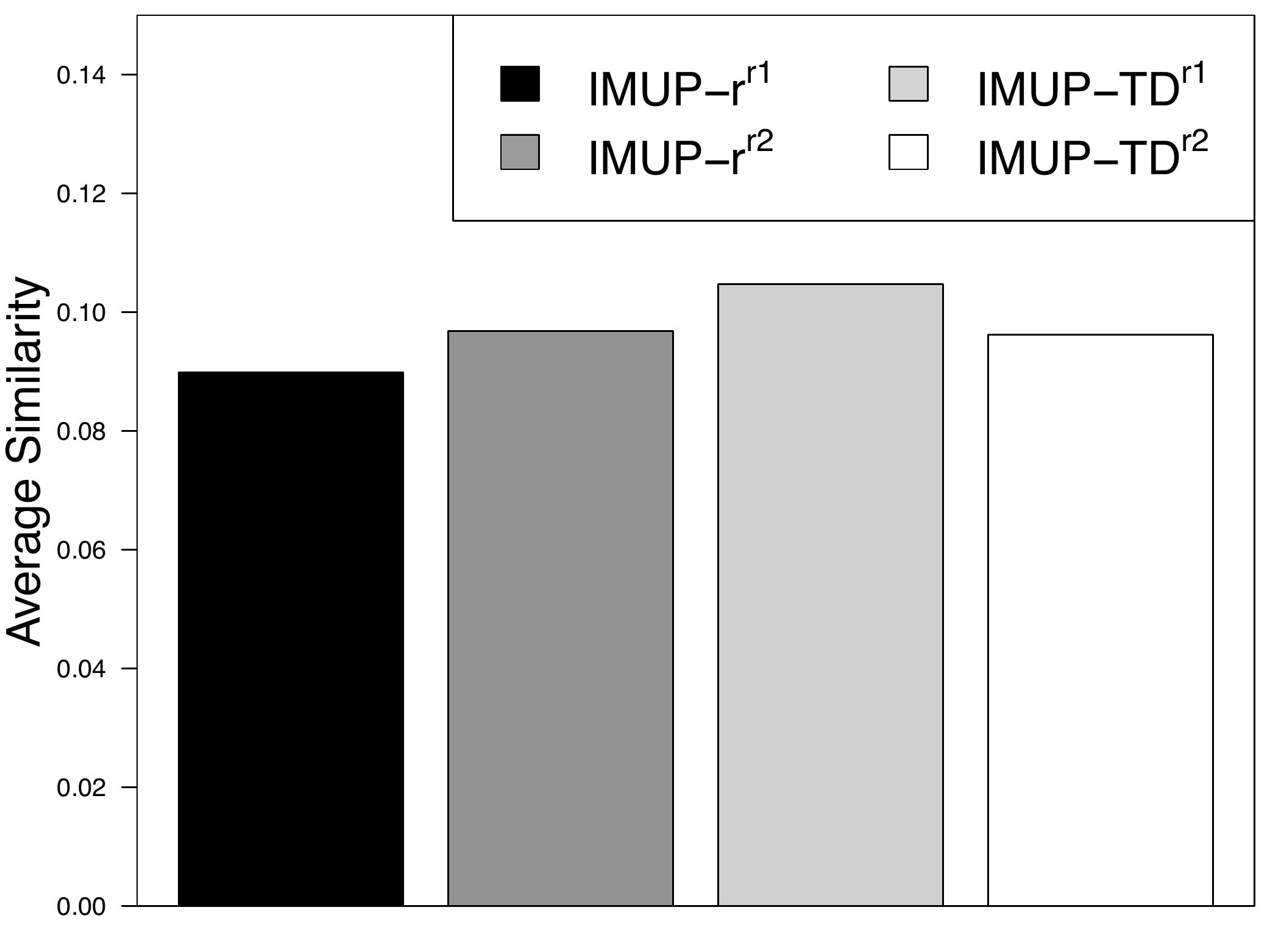}}
	\subfigure[Average Distance]{\label{fig:nyc_diff_reward_dis}\includegraphics[width=4.35cm]{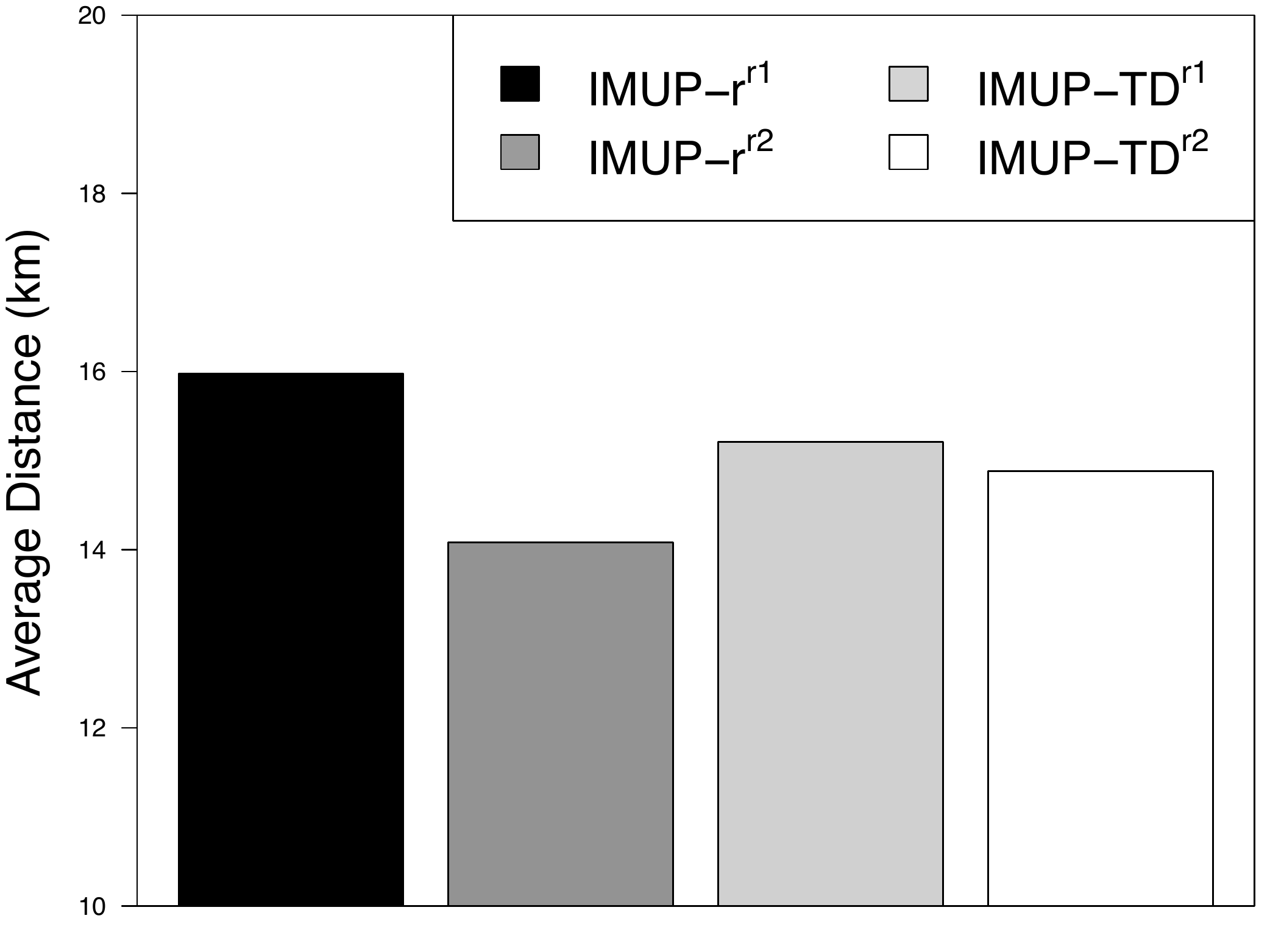}}
 	\vspace{-0.4cm}
	\captionsetup{justification=centering}
	\caption{The comparison between the original reward ($r1$) and the enhanced reward ($r2$) {\it w.r.t.} New York dataset. }
 	\vspace{-0.4cm}
	\label{fig:diff_reward_nyc}
\end{figure*}

\begin{figure*}[!tb]
	\centering
	\subfigure[Precision on Category]{\label{fig:bj_diff_reward_precision}\includegraphics[width=4.35cm]{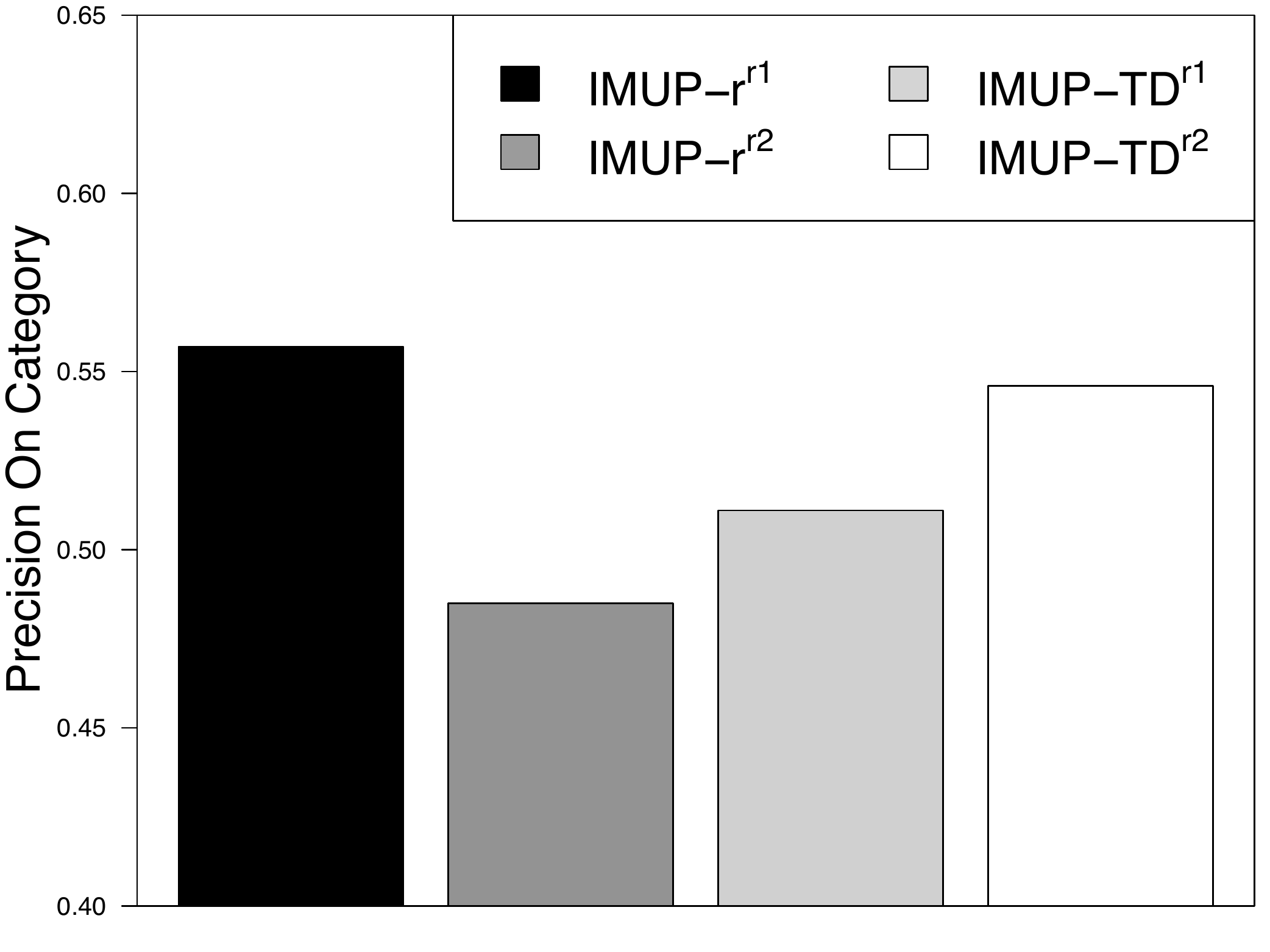}}
	\subfigure[Recall on Category]{\label{fig:bj_diff_reward_recall}\includegraphics[width=4.35cm]{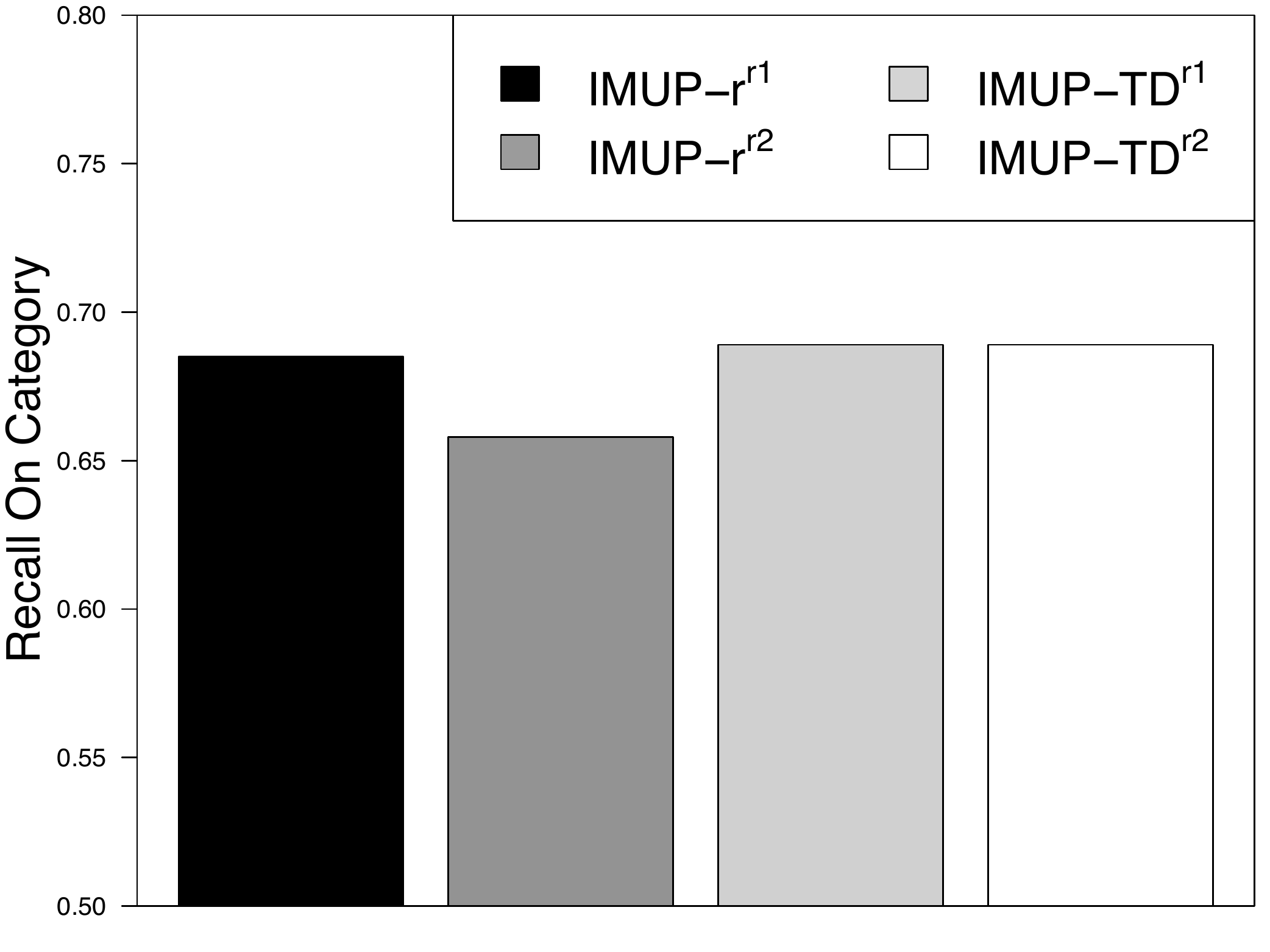}}
	\subfigure[Average Similarity]{\label{fig:bj_diff_reward_sim}\includegraphics[width=4.35cm]{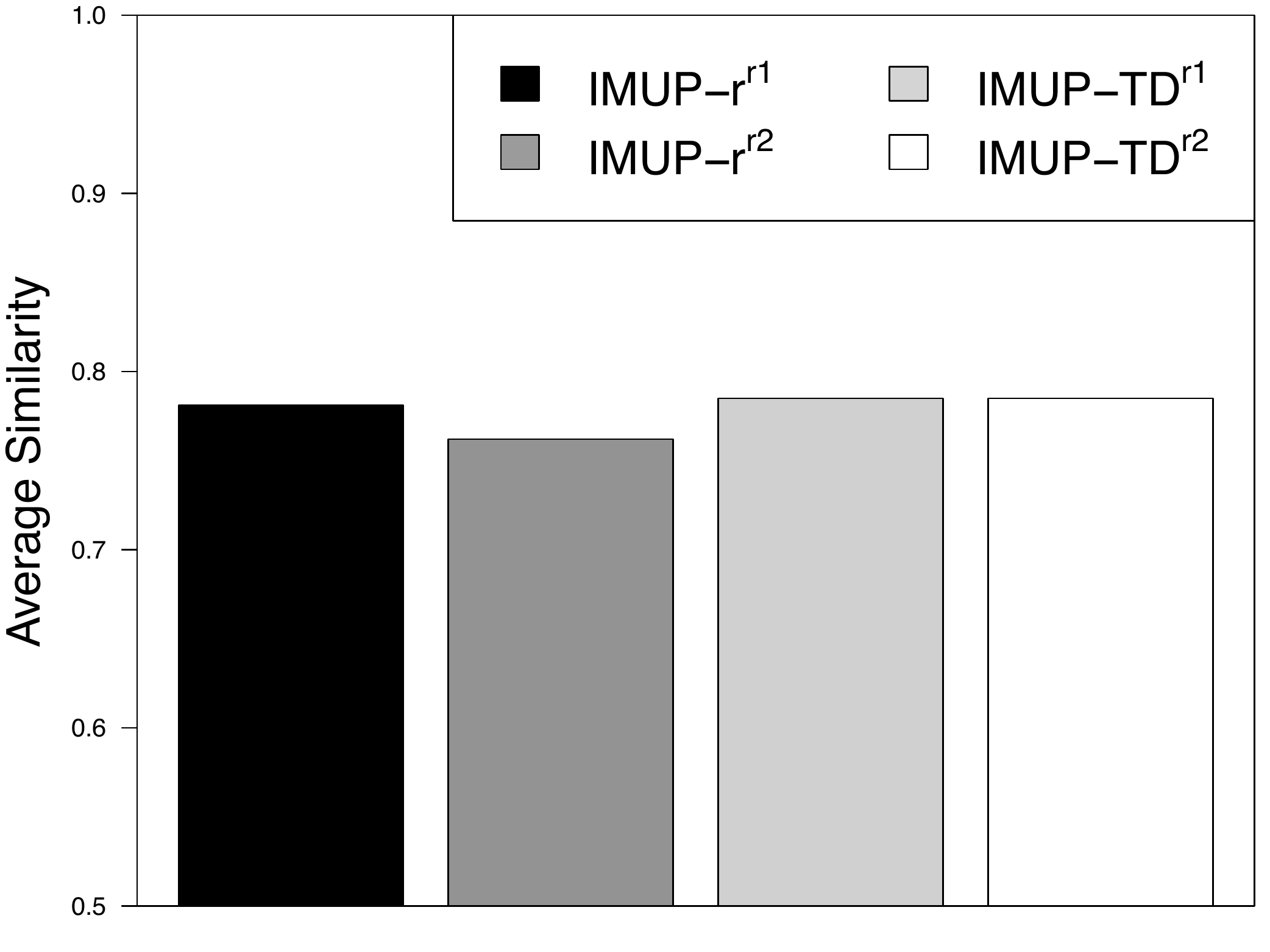}}
	\subfigure[Average Distance]{\label{fig:bj_diff_reward_dis}\includegraphics[width=4.35cm]{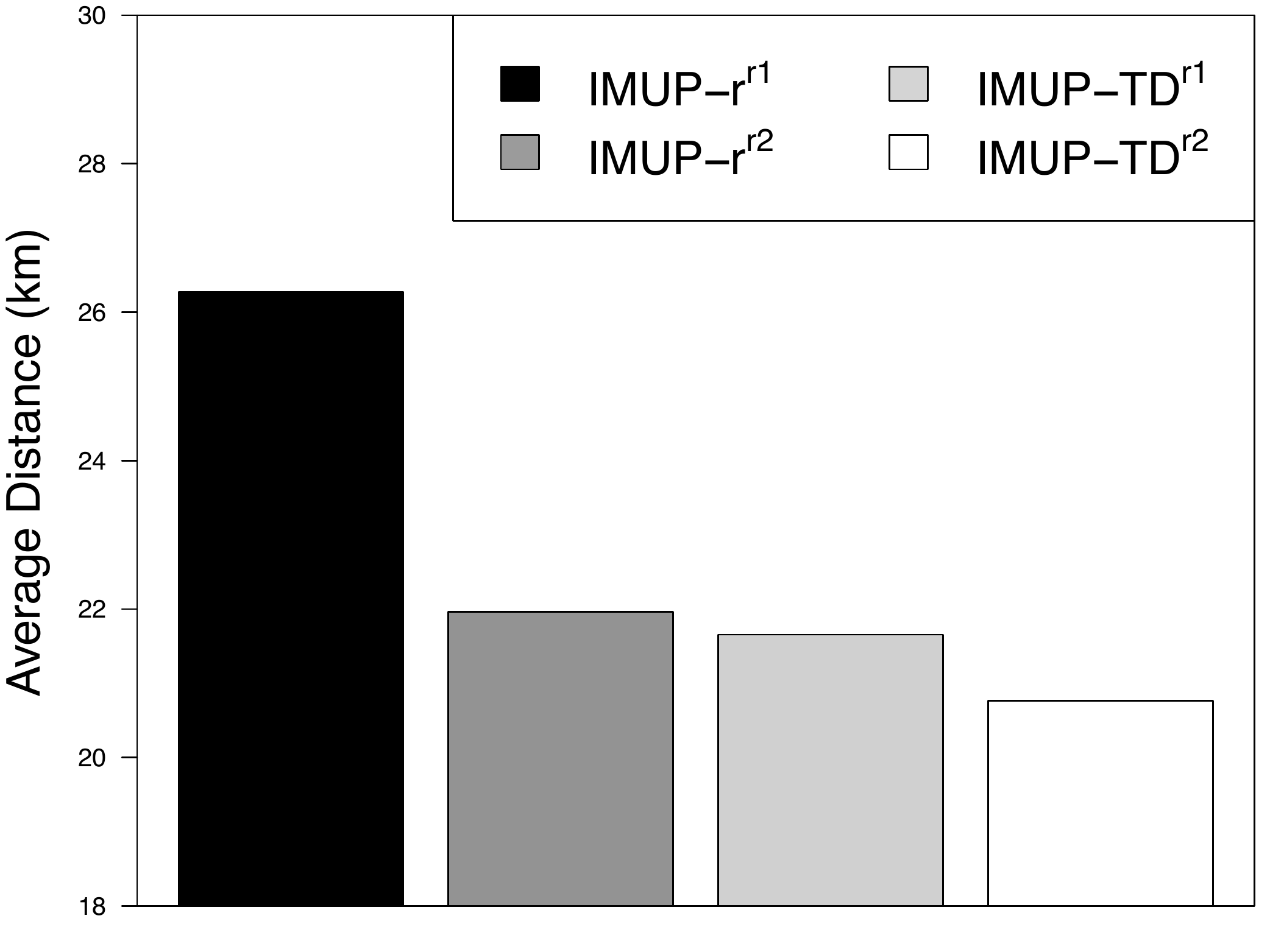}}
 	\vspace{-0.4cm}
	\captionsetup{justification=centering}
	\caption{The comparison between the original reward ($r1$) and the enhanced reward ($r2$) {\it w.r.t.} Beijing dataset.}
	\label{fig:diff_reward_bj}
\end{figure*}

\noindent {\bf (4) $\text{Average Distance (Avg\_Dist)}$:} We evaluate the prediction on location with the average distance.
Let $Dist(P^{l}, \hat{P}^{l})$ denote the distance between the real location $P^{l}$ and the predicted location $\hat{P}^l$ for the $l$-th visit, then the average distance is 
\begin{equation}
    \text{Avg\_Dist} = \frac{\sum \limits_{l} Dist(P^{l}, \hat{P}^{l})}{L}.
\end{equation}
The lower the value of $Avg\_Dist$, the better the prediction.

\subsection{Baseline Algorithms}
We compare the performances of our method (namely ``IMUP'') against the following baseline algorithms.

\noindent{\bf (1) PMF}  is a classic framework for modeling users through the probabilistic matrix factorization over the user-item interaction matrix ~\cite{mnih2008probabilistic}.

\noindent{\bf (2) PoolNet} is proposed to learn user representations by averaging the representations of items with which they have interacted through a deep neural network model ~\cite{covington2016deep}. 

\noindent{\bf (3) WaveNet} is originally designed for generating raw audio waveforms in a generative way. It can also be applied to learning representations for sequential decision-making through stacked causal atrous convolutions~\cite{oord2016wavenet}.

\noindent{\bf (4) LSTMNet} represents users as the hidden state at each timestep with a recurrent neural network by feeding the visiting sequence in to the model~\cite{hidasi2015session}.

\noindent{\bf (5) StructRL} models substructures of user mobility graph for quantifying specific activity patterns through the adversarial learning paradigm~\cite{wang2019adversarial}.


By contrast, in the conference version \cite{wang2020incremental}, there are two variants for incremental user profiling: 
{\bf (1) IMUP-r}, where the model utilizes the sampling strategy with the priority $x_r$;
{\bf (2) IMUP-TD}, where the model utilizes the sampling strategy with the priority $x_{TD}$. 
In this work,  we propose a new reward function, a new policy network, and a new state update strategy enhance the conference version work. 
To validate the effectiveness of each part, we develop

Here, we develop two variants for this paper:
{\bf (1) IMUP$^\mathbf{*}$-r}, {\bf (2) IMUP$^\mathbf{*}$-TD}.
Compared with {\bf IMUP-r} and {\bf IMUP-TD},
the difference is the two variants utilize all new modifications of this paper during the process of incremental user profiling.

In the experiment,  we split the datasets into two non-overlapping sets: for each user, the earliest $90\%$ of check-ins are the training set and the remaining $10\%$ check-ins are testing set. 
We set the dimension of user representations as 200 for all the baselines and our proposed methods.
We adopt the implementation~\footnote{\tiny\url{https://github.com/fuhailin/Probabilistic-Matrix-Factorization}} for evaluating PMF.
And we adopt ``spotlight''~\cite{kula2017spotlight} for evaluating PoolNet, WaveNet, and LSTMNet, where the learning rate is set as 0.01.
We follow the parameter setting of StructRL in~\cite{wang2019adversarial}.
Our code and data are released by Dropbox~\footnote{\tiny\url{https://www.dropbox.com/sh/l6syw989teu9b96/AADXfq-6Es6LjMEFfrf4t3Kna?dl=0}}.

All evaluations were conducted using Ubuntu 18.04.3 LTS, on an Intel(R) Core(TM) i9-9920X CPU@ 3.50GHz, with Titan RTX 32 GB and RAM memory size 128G.

\subsection{Overall comparison}
Figure~\ref{fig:nyc_overall} and Figure~\ref{fig:bj_overall} show the comparison results of user profile quality in terms of ``Precision on Category'', ``Recall on Category'', ``Average Distance'' and ``Average Similarity''.
We can find that  ``IMUP$^*$-r'' and ``IMUP$^*$-TD'' 
outperform other baseline models on both New York and Beijing datasets, and significantly enhance  ``Average Distance'' and ``Average Similarity''.
A potential reason for the observation is that the representation of the spatial {\it KG} involves the geospatial correlations into state vectors. 
So, the reinforced agent can capture the user mobility semantics and interests based on such states to mimic user mobility patterns more accurately.
Moreover, the new state update strategy is able to automatically discard old user visit interests and incorporate new user visit preferences. 
Thus, the next visit prediction of our method becomes as close as possible to the real one from the semantic and geographical perspectives.

\begin{figure*}[!tb]
	\centering
	\subfigure[Precision on Category]{\label{fig:nyc_diff_rein_precision}\includegraphics[width=4.35cm]{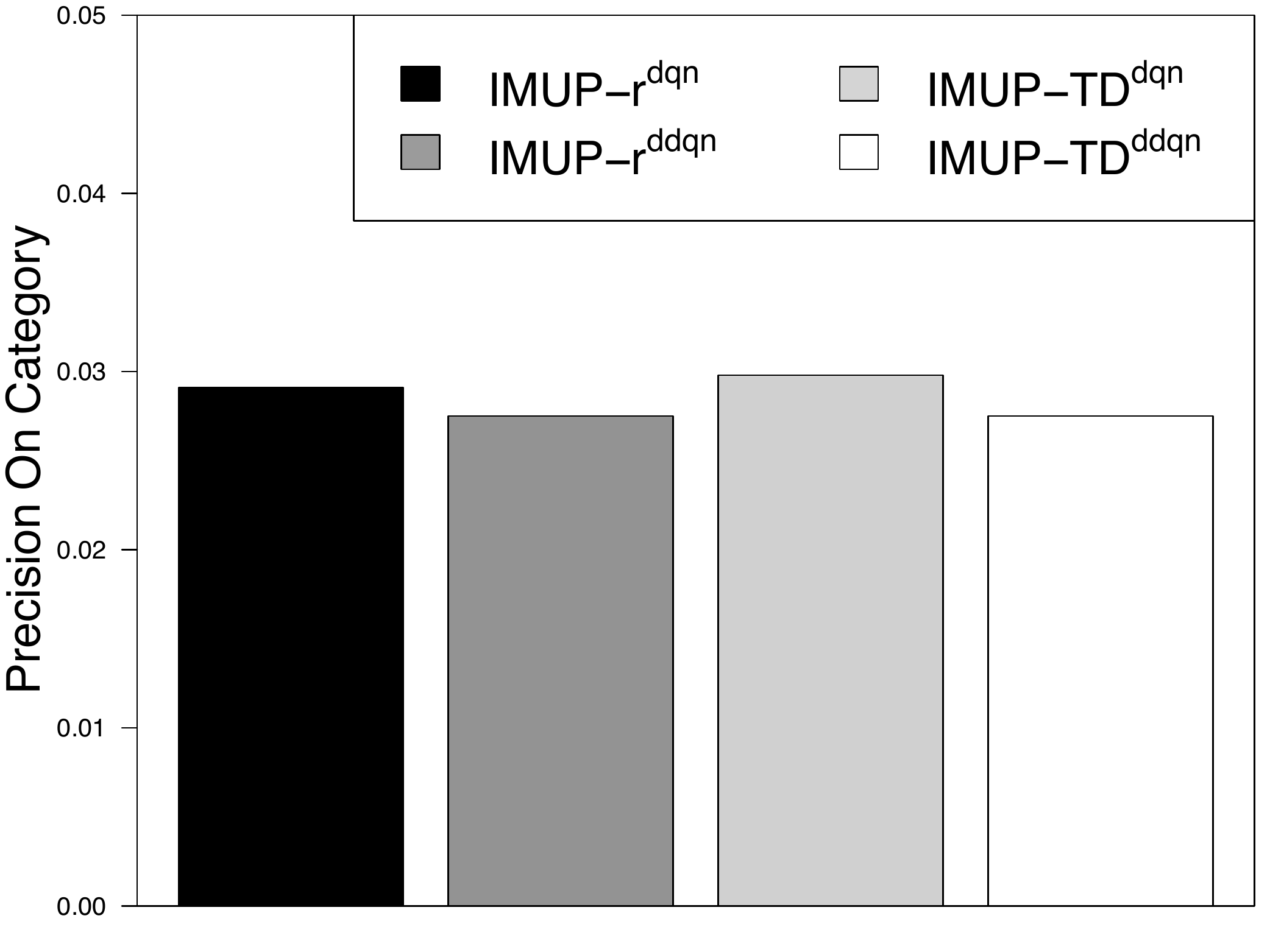}}
	\subfigure[Recall on Category]{\label{fig:nyc_diff_rein_recall}\includegraphics[width=4.35cm]{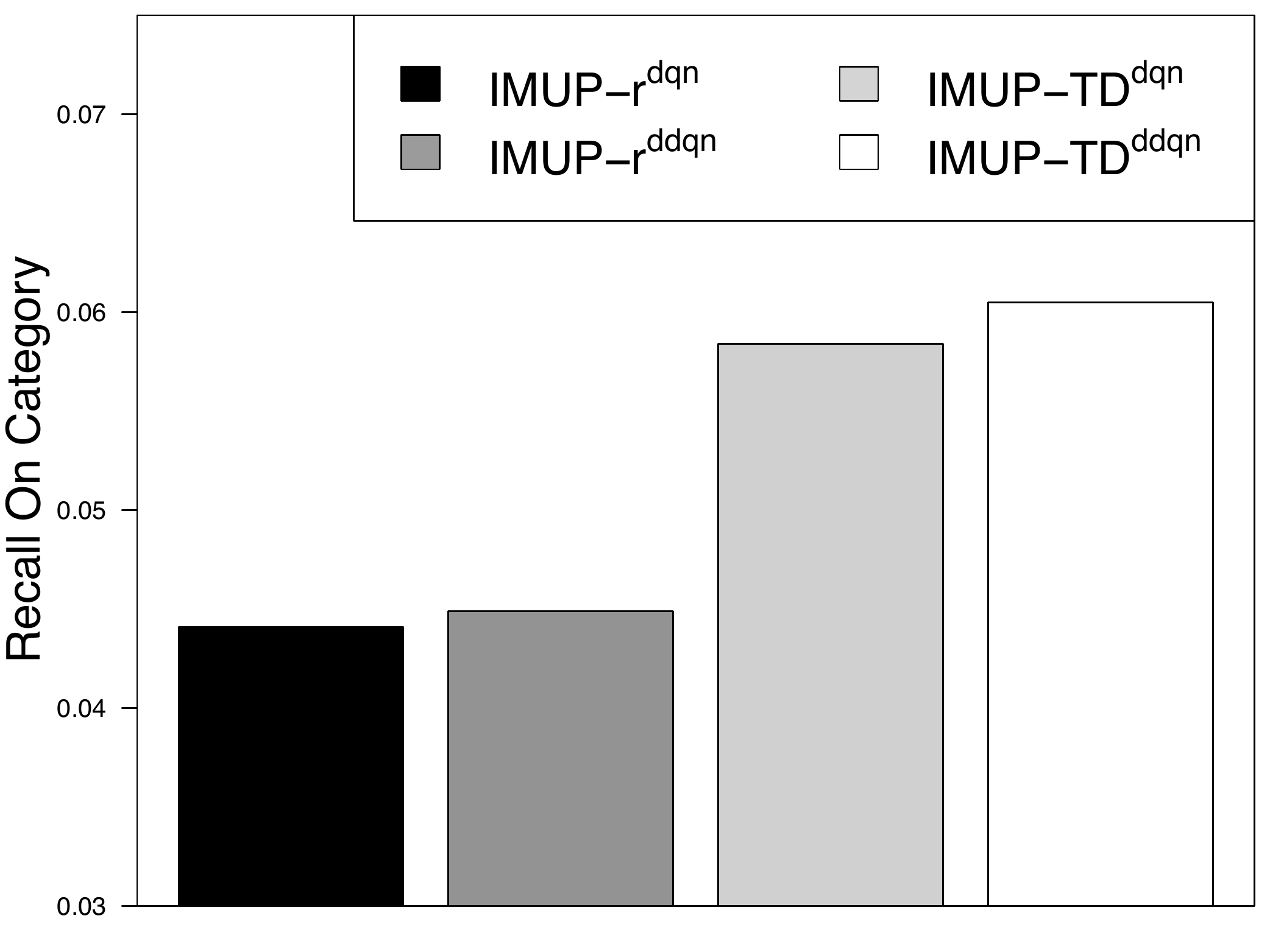}}
	\subfigure[Average Similarity]{\label{fig:nyc_diff_rein_sim}\includegraphics[width=4.35cm]{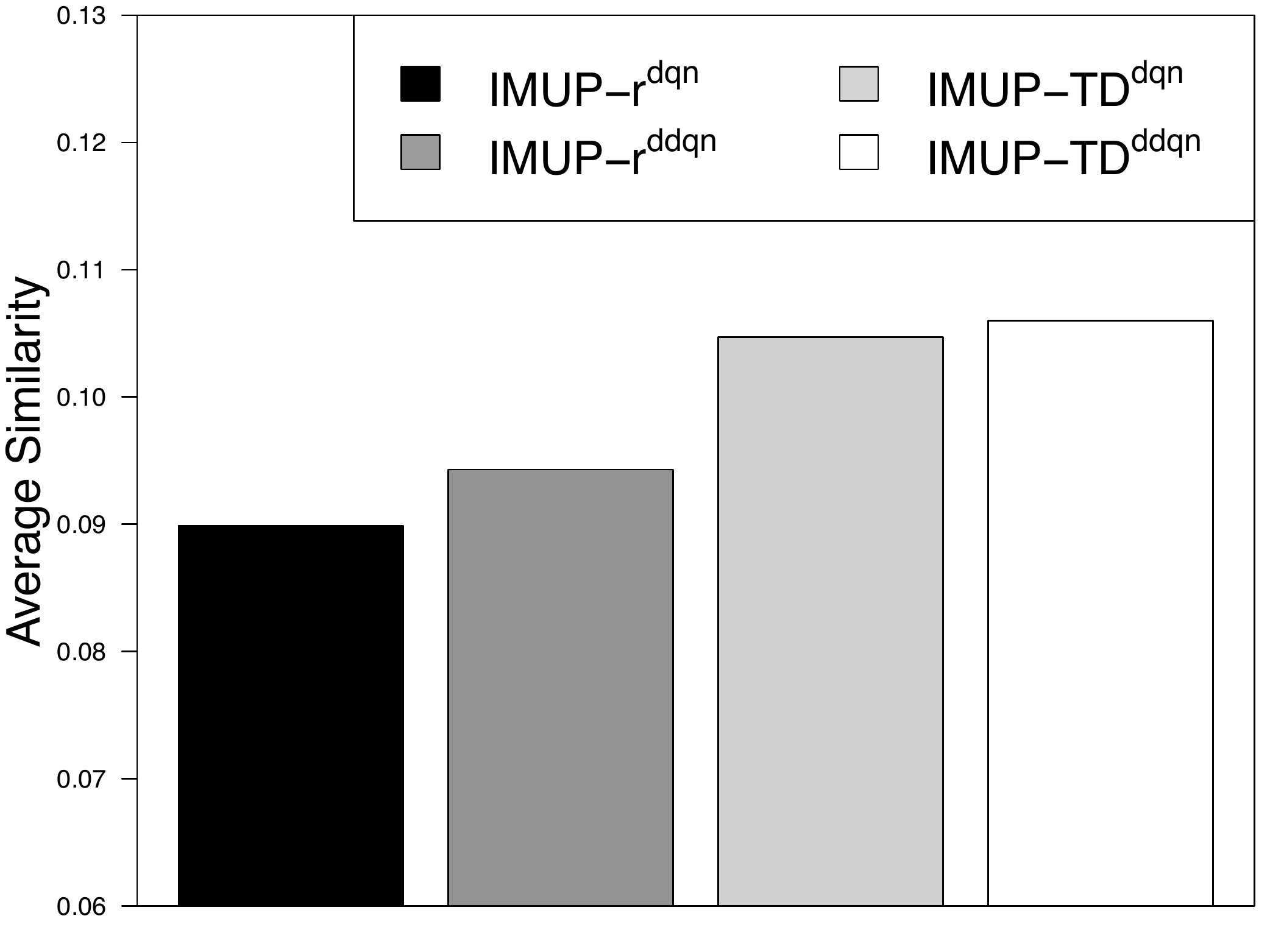}}
	\subfigure[Average Distance]{\label{fig:nyc_diff_rein_dis}\includegraphics[width=4.35cm]{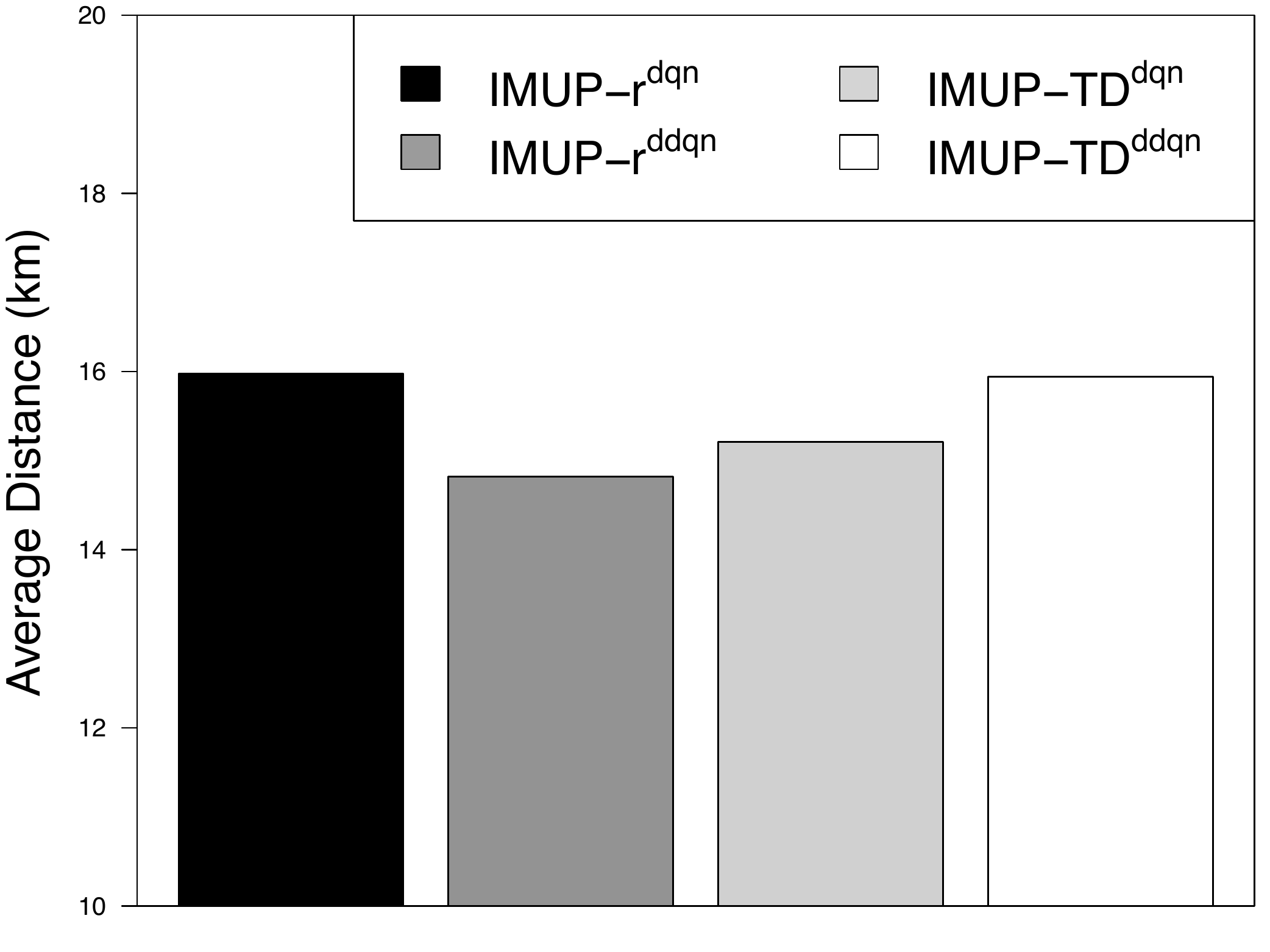}}
 	\vspace{-0.4cm}
	\captionsetup{justification=centering}
	\caption{The comparison between the original agent ($dqn$) and the enhanced agent ($ddqn$) {\it w.r.t.} New York dataset.}
 	\vspace{-0.4cm}
	\label{fig:diff_rein_nyc}
\end{figure*}

\begin{figure*}[!tb]
	\centering
	\subfigure[Precision on Category]{\label{fig:bj_diff_rein_precision}\includegraphics[width=4.35cm]{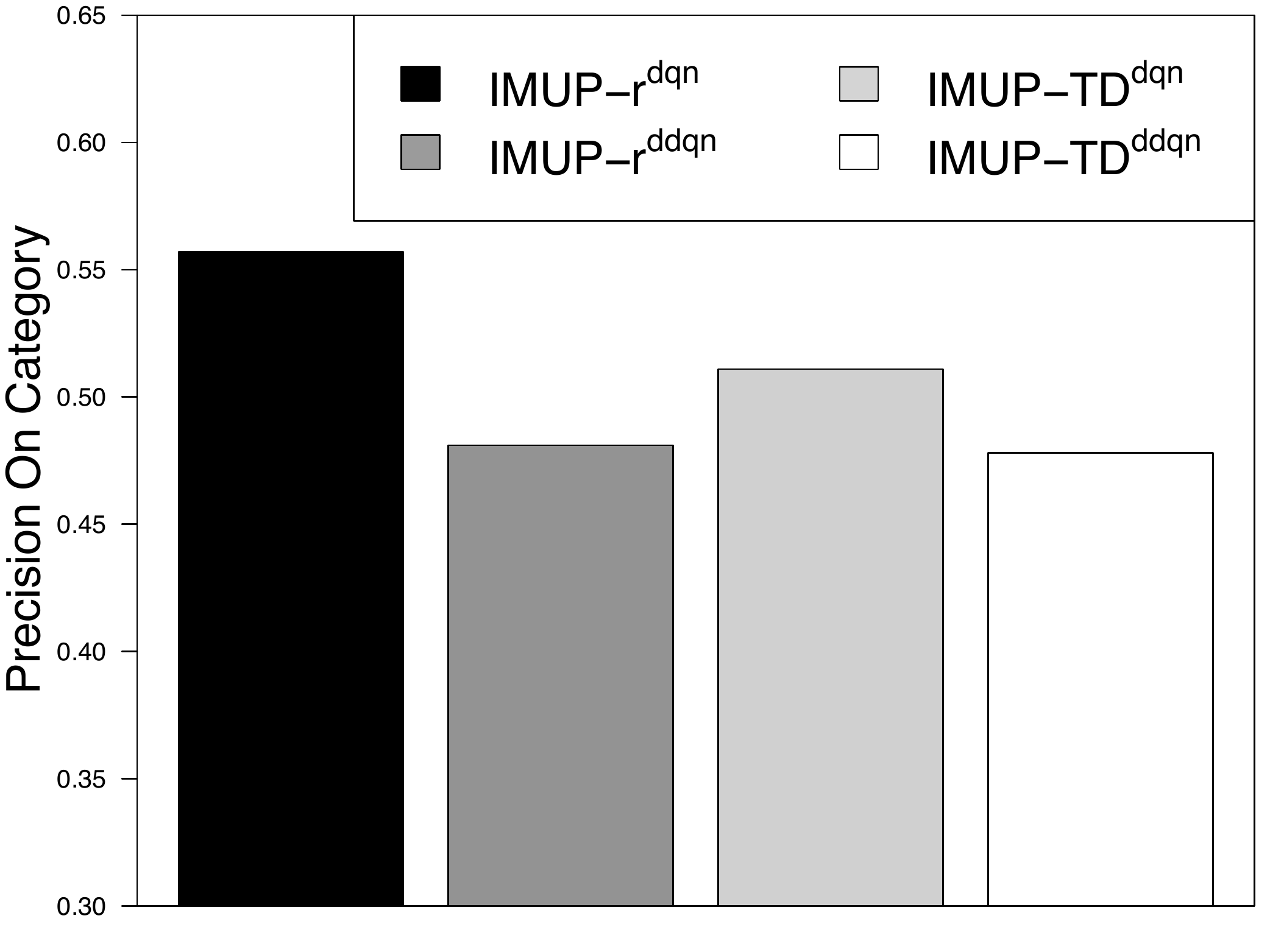}}
	\subfigure[Recall on Category]{\label{fig:bj_diff_rein_recall}\includegraphics[width=4.35cm]{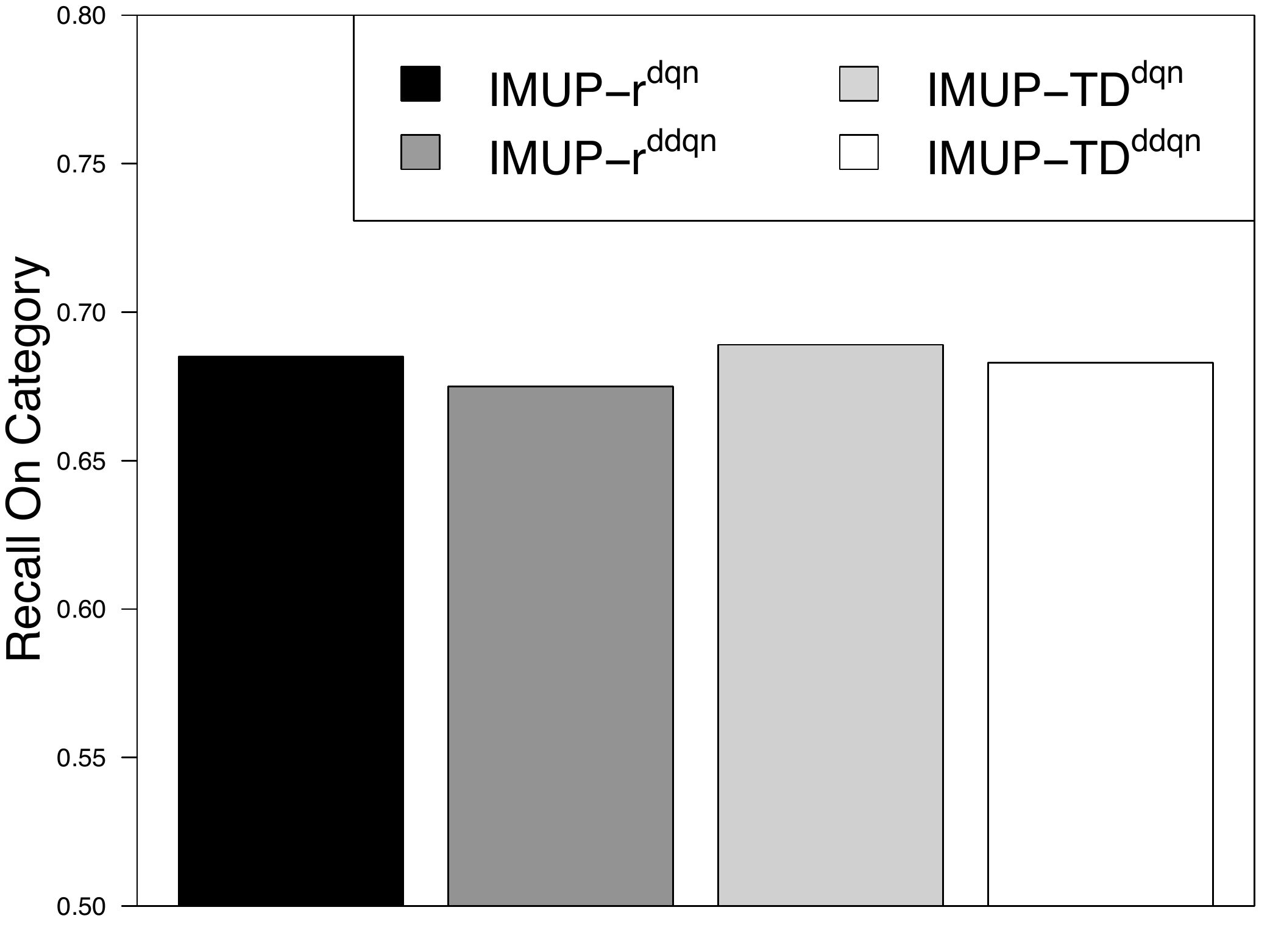}}
	\subfigure[Average Similarity]{\label{fig:bj_diff_rein_sim}\includegraphics[width=4.35cm]{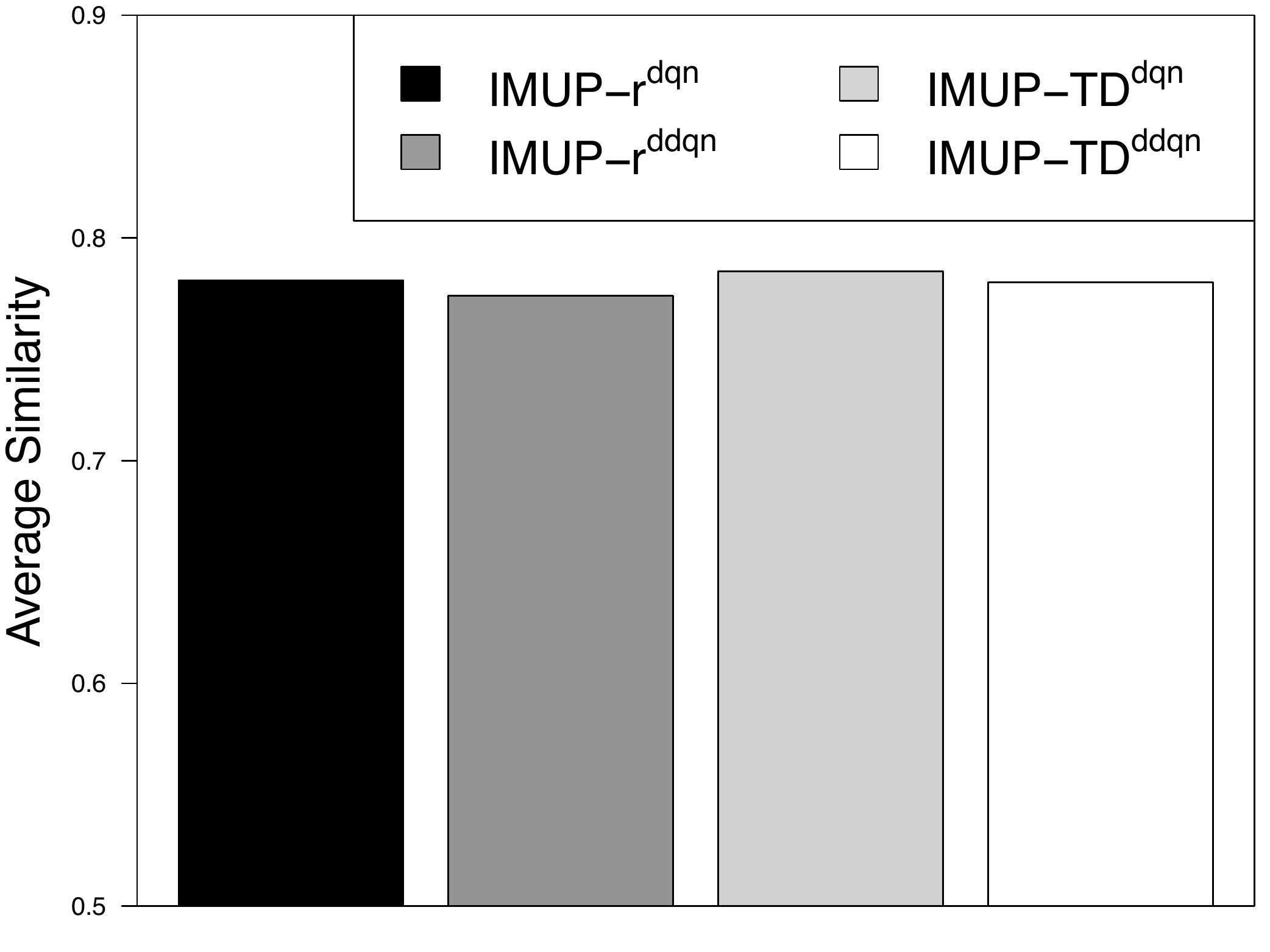}}
	\subfigure[Average Distance]{\label{fig:bj_diff_rein_dis}\includegraphics[width=4.35cm]{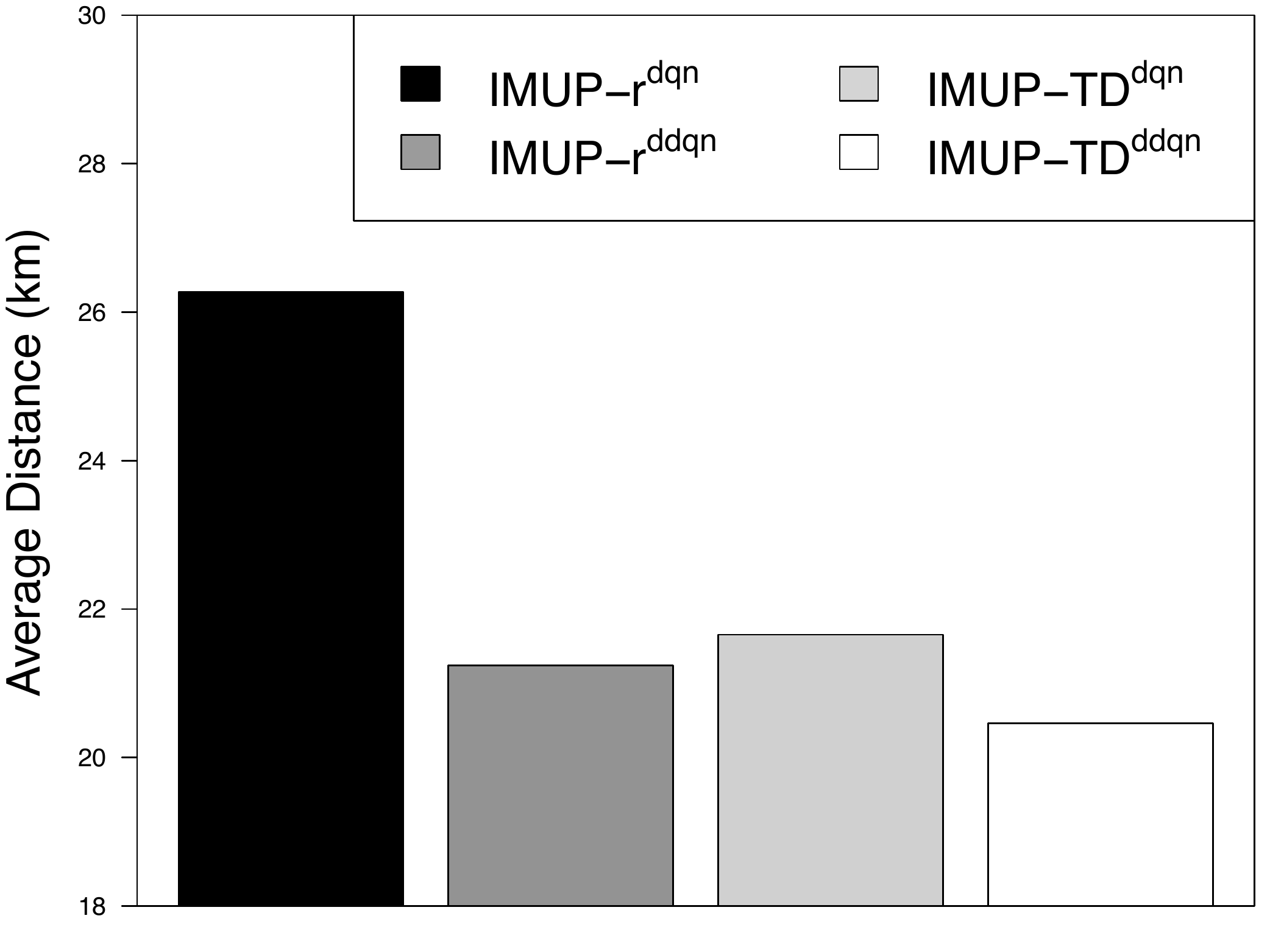}}
 	\vspace{-0.4cm}
	\captionsetup{justification=centering}
	\caption{The comparison between the original agent ($dqn$) and the enhanced agent ($ddqn$) {\it w.r.t.} Beijing dataset.}
 	\vspace{-0.4cm}
	\label{fig:diff_rein_bj}
\end{figure*}

\begin{figure*}[!tb]
	\centering
	\subfigure[Precision on Category]{\label{fig:nyc_diff_update_precision}\includegraphics[width=4.35cm]{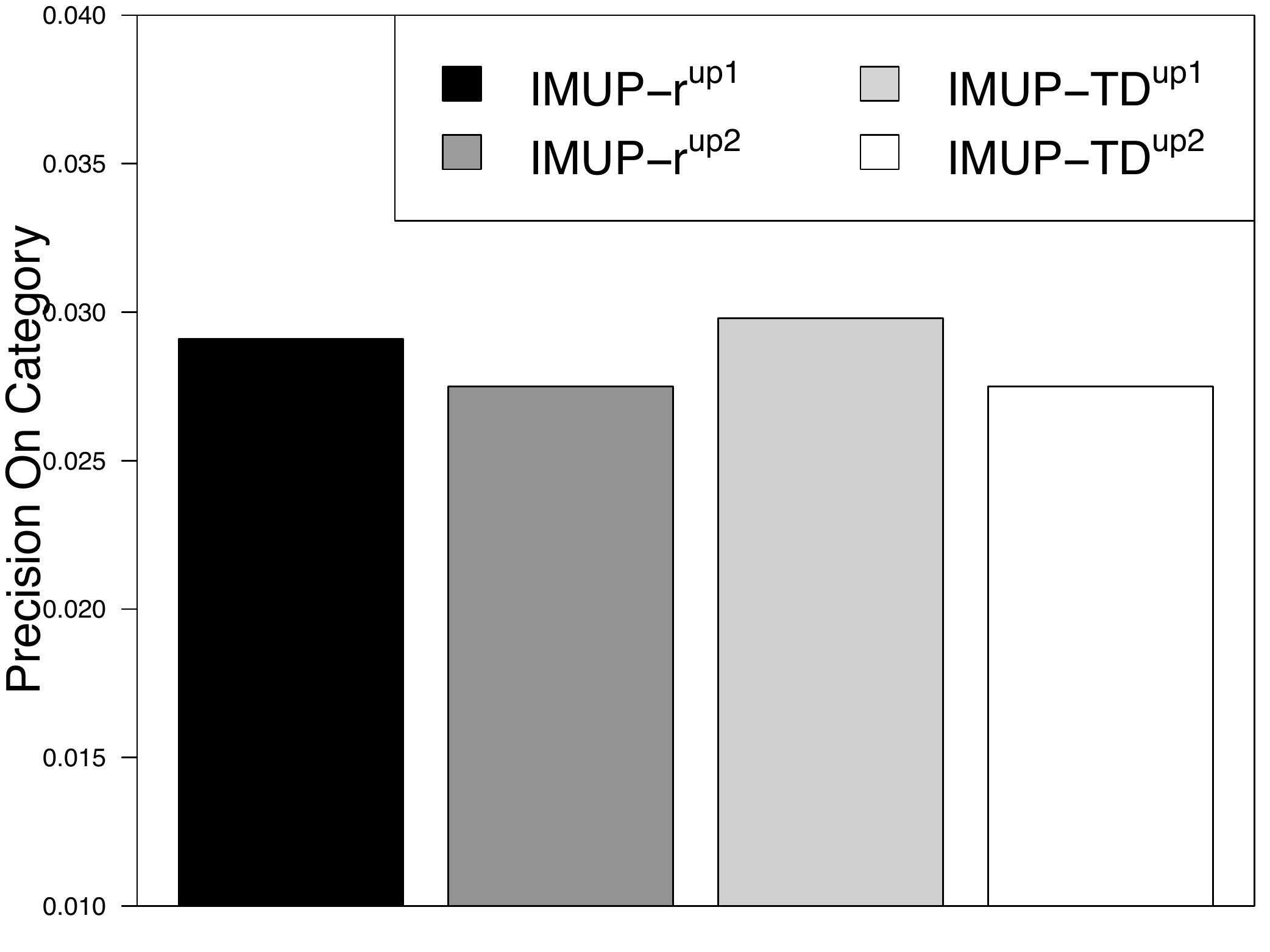}}
	\subfigure[Recall on Category]{\label{fig:nyc_diff_update_recall}\includegraphics[width=4.35cm]{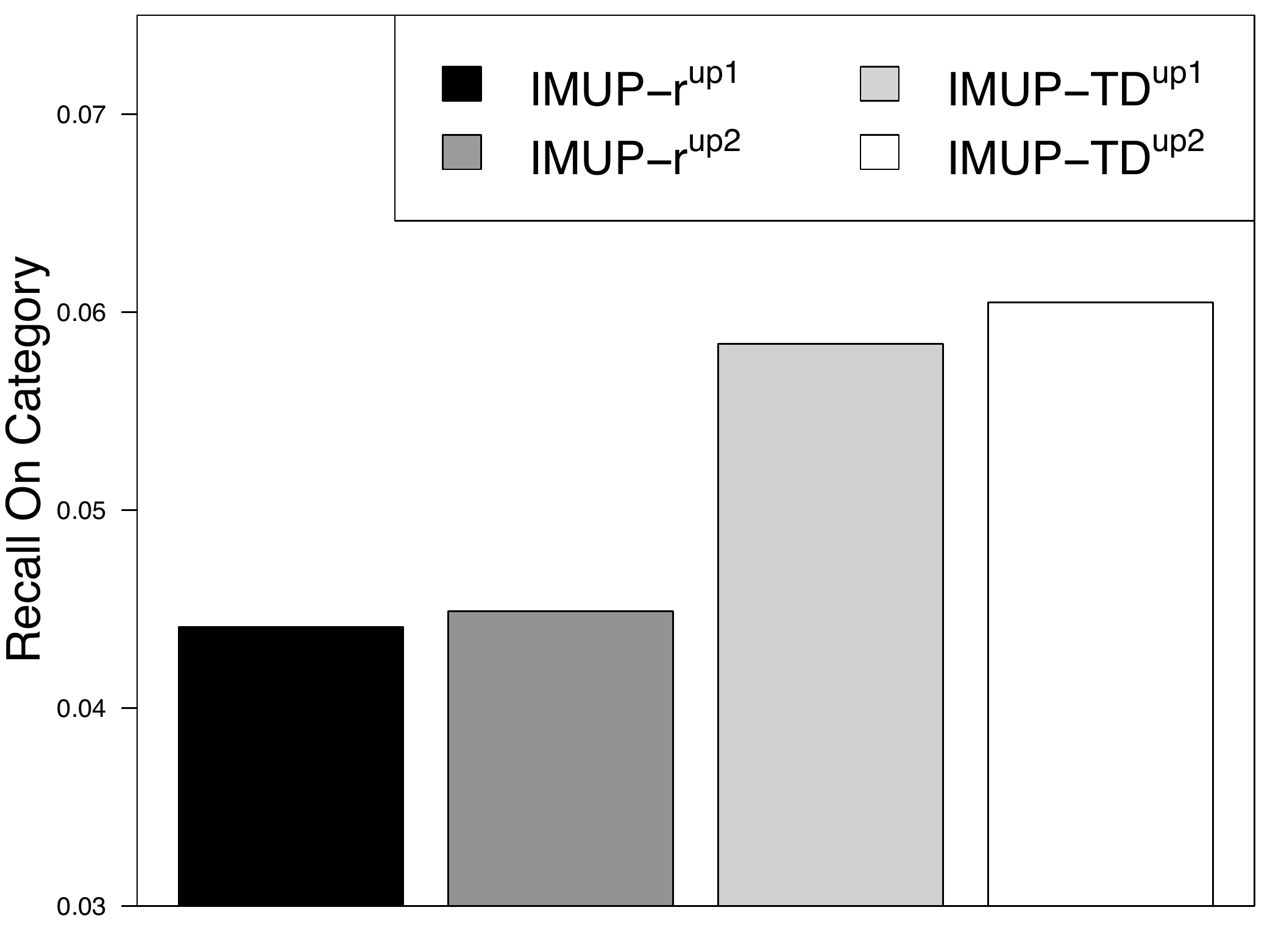}}
	\subfigure[Average Similarity]{\label{fig:nyc_diff_update_sim}\includegraphics[width=4.35cm]{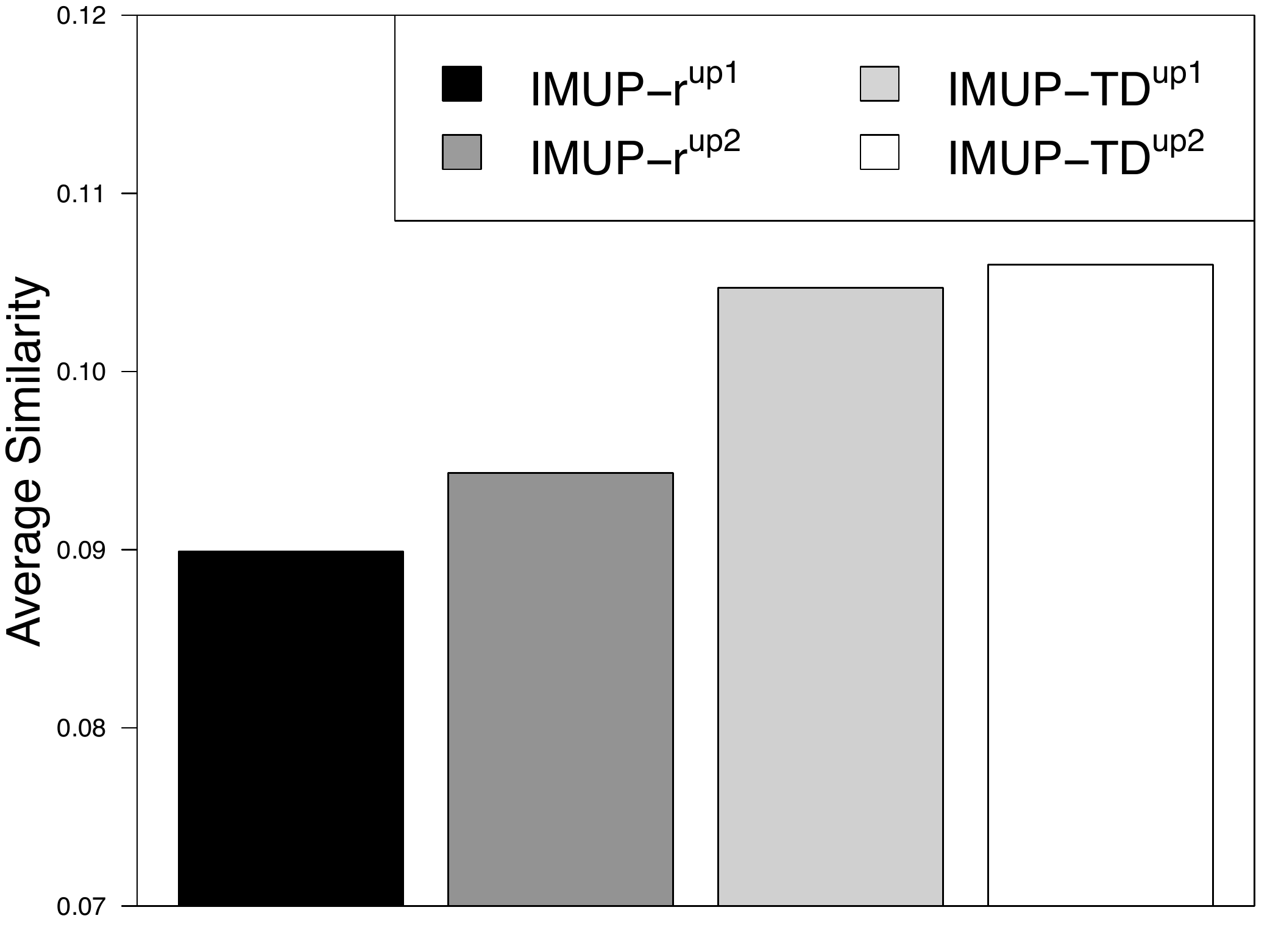}}
	\subfigure[Average Distance]{\label{fig:nyc_diff_update_dis}\includegraphics[width=4.35cm]{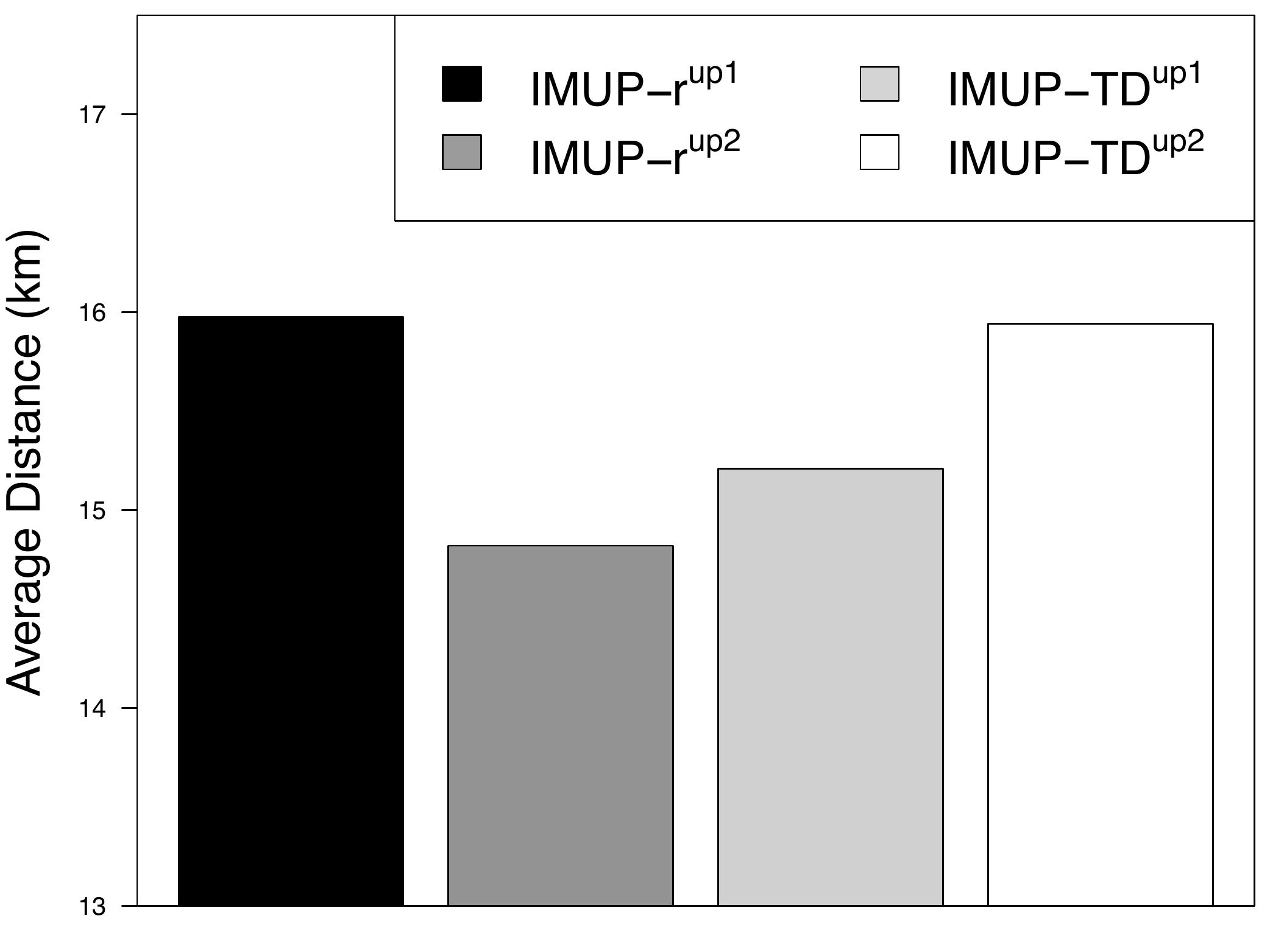}}
 	\vspace{-0.4cm}
	\captionsetup{justification=centering}
	\caption{The comparison between the original ($up1$) and enhanced ($up2$) state update strategies {\it w.r.t.} New York dataset.}
 	\vspace{-0.4cm}
	\label{fig:diff_update_nyc}
\end{figure*}

\begin{figure*}[!tb]
	\centering
	\subfigure[Precision on Category]{\label{fig:bj_diff_update_precision}\includegraphics[width=4.35cm]{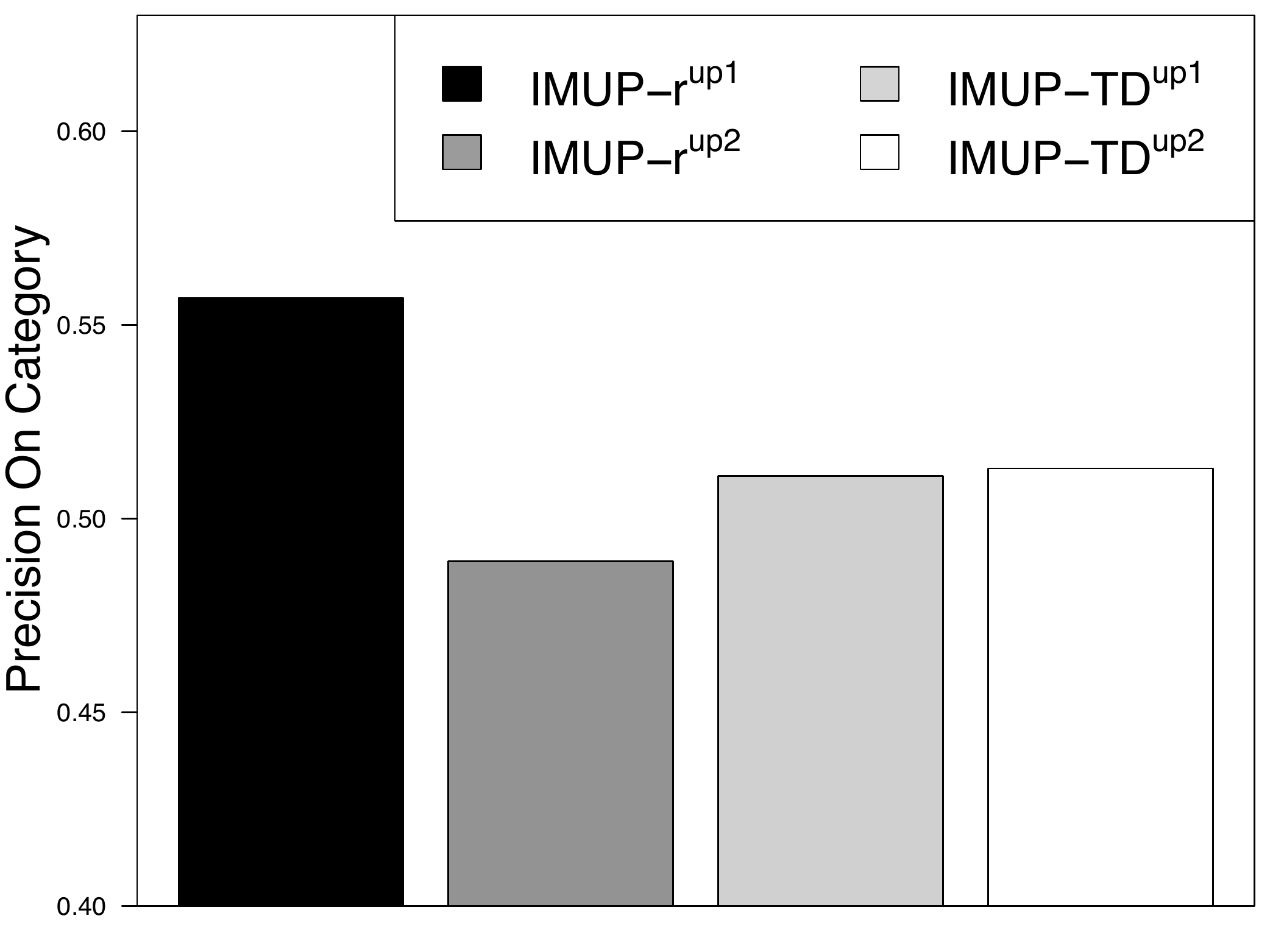}}
	\subfigure[Recall on Category]{\label{fig:bj_diff_update_recall}\includegraphics[width=4.35cm]{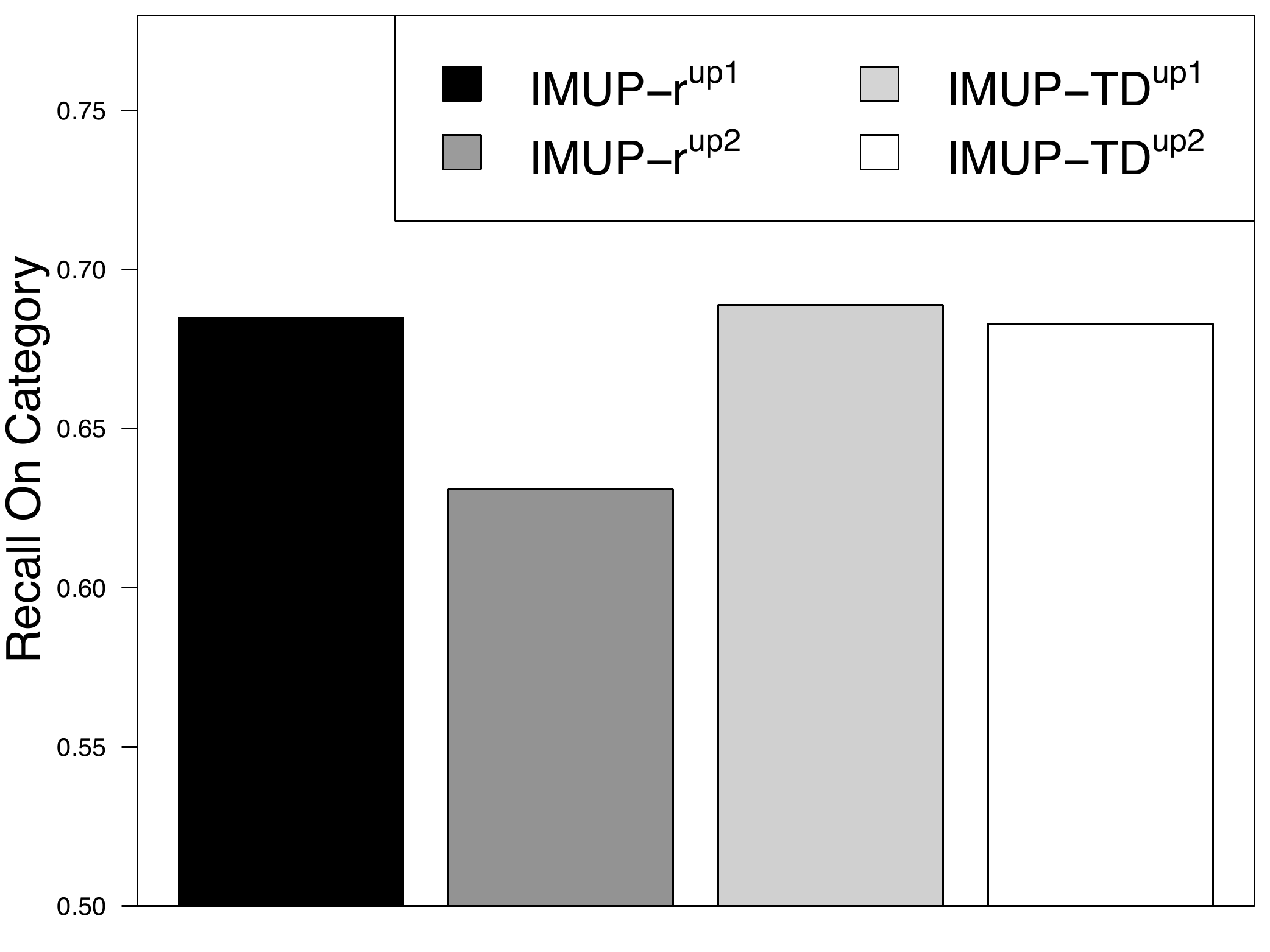}}
	\subfigure[Average Similarity]{\label{fig:bj_diff_update_sim}\includegraphics[width=4.35cm]{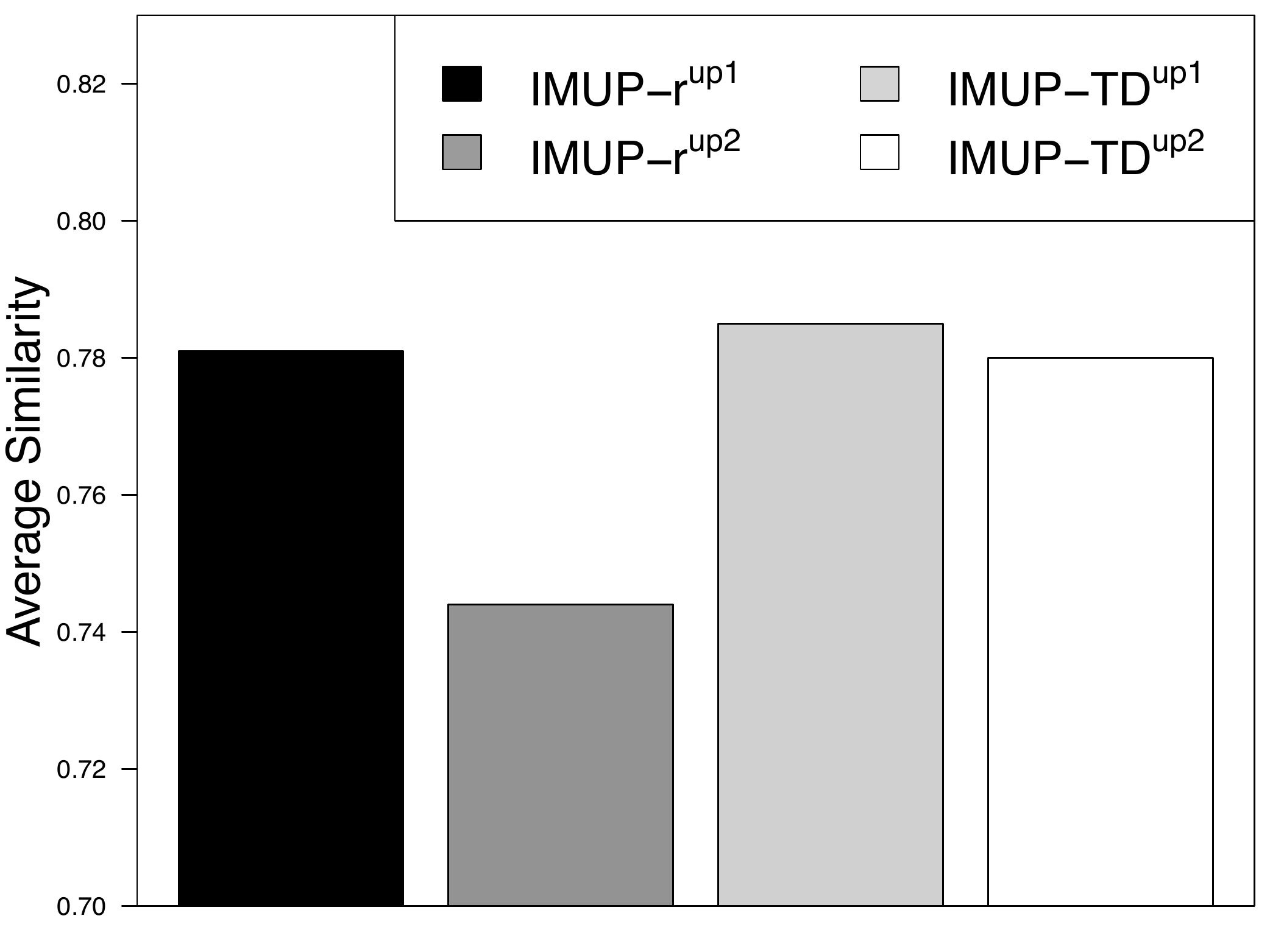}}
	\subfigure[Average Distance]{\label{fig:bj_diff_update_dis}\includegraphics[width=4.35cm]{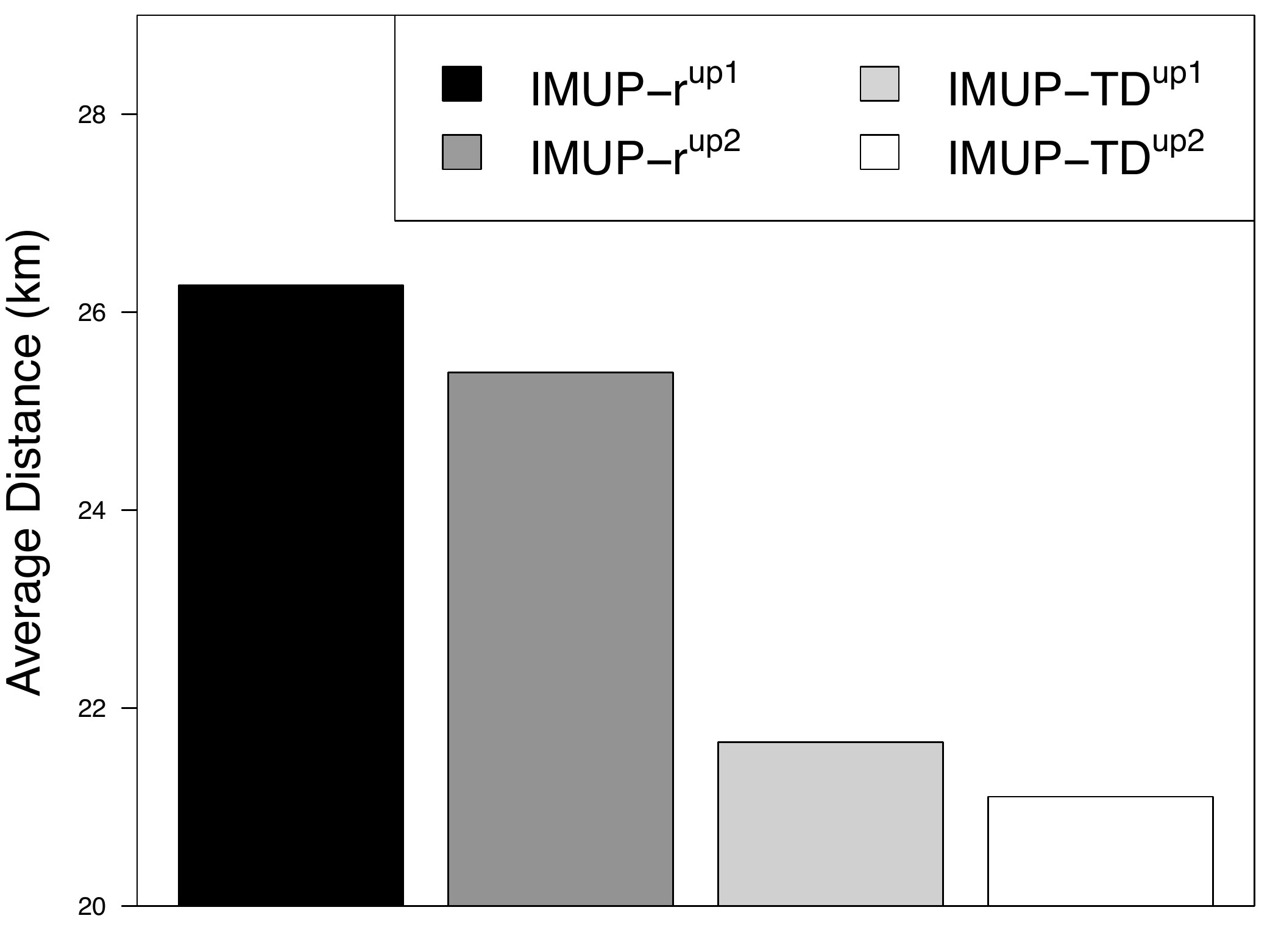}}
 	\vspace{-0.4cm}
	\captionsetup{justification=centering}
	\caption{The comparison between the original ($up1$) and enhanced ($up2$) state update strategies {\it w.r.t.} Beijing dataset.}
 	\vspace{-0.4cm}
	\label{fig:diff_update_bj}
\end{figure*}

\subsection{Comparison of different sampling strategies}
We propose two priority-based sampling strategies to improve the policy learning procedure.
Figure~\ref{fig:nyc_overall} and Figure~\ref{fig:bj_overall} show the comparison results in terms of all evaluation metrics.
We can find that the $TD$-based sampling strategy is slightly better than reward-based sampling strategy.
A potential reason is that the reinforced agent covers the obvious user visit interests but makes mistakes for several implicit user preferences.
The $TD$-based strategy can force the reinforced agent to learn such confused knowledge by sampling related  data samples in order to improve the user mobility imitation.


\subsection{Comparison of different rewards}
We replace the reward function ($r1$) of the conference version \cite{wang2020incremental} with the new reward function ($r2$) in this work.
We develop four variants based on different sampling strategies and the two rewards, namely "IMUP$-r^{r1}$","IMUP$-r^{r2}$",
"IMUP$-TD^{r1}$","IMUP$-TD^{r2}$".
Figure \ref{fig:diff_reward_nyc} and Figure \ref{fig:diff_reward_bj} show the comparison results.
We can find that $r2$  outperforms $r1$ significantly in terms of  ``Average Distance'', and keeps stable in terms of
``Precision on Category'', ``Recall on Category'', ``Average Similarity''.
The underlying driver is that $r2$ encourages the agent to search for possible visited POIs that are semantically and geographically close to the real visited POI.
If the predicted POI is further from the real one than our expectation on geographical distance, the agent will be punished via getting negative rewards.
Thus, in order to get more positive rewards, the agent makes more accurate next-visit predictions that
are more semantically and geographically close to real visited POIs.

\subsection{Comparison of different policy networks}
We replace the reinforced module of our framework from DQN with double DQN  to mitigate Q-value overestimation to improve model performance.
We use ``$dqn$'' and  ``$ddqn$''  to denote the DQN and double DQN respectively, and develop four variants of IMUP based on two sampling strategies:  "IMUP$-r^{dqn}$'',"IMUP$-r^{ddqn}$'',
"IMUP$-TD^{dqn}$'' and "IMUP$-TD^{ddqn}$''.

Figure \ref{fig:diff_rein_nyc} and Figure \ref{fig:diff_rein_bj} show the comparison results in terms of all evaluation metrics.
We can find that $ddqn$ fluctuates slightly compared with $dqn$ in terms of ``Precision on Category'', ``Recall on Category'', ``Average Similarity'' and outperforms $dqn$ in terms of ``Average Distance'' on both New York and Beijing datasets. 
A potential reason for this observation is that the double DQN alleviates the Q-value overestimation of DQN.
So the double DQN can provide an objective and fair Q-value for each action, resulting in learning more effective action selection and action evaluation policies.
Thus, the reinforced agent can imitate the user mobility behavior better than before.


\begin{figure*}[!thb]
	\centering
	\subfigure[Precision on Category]{\label{fig:substructure_precision_newyork}\includegraphics[width=4.35cm]{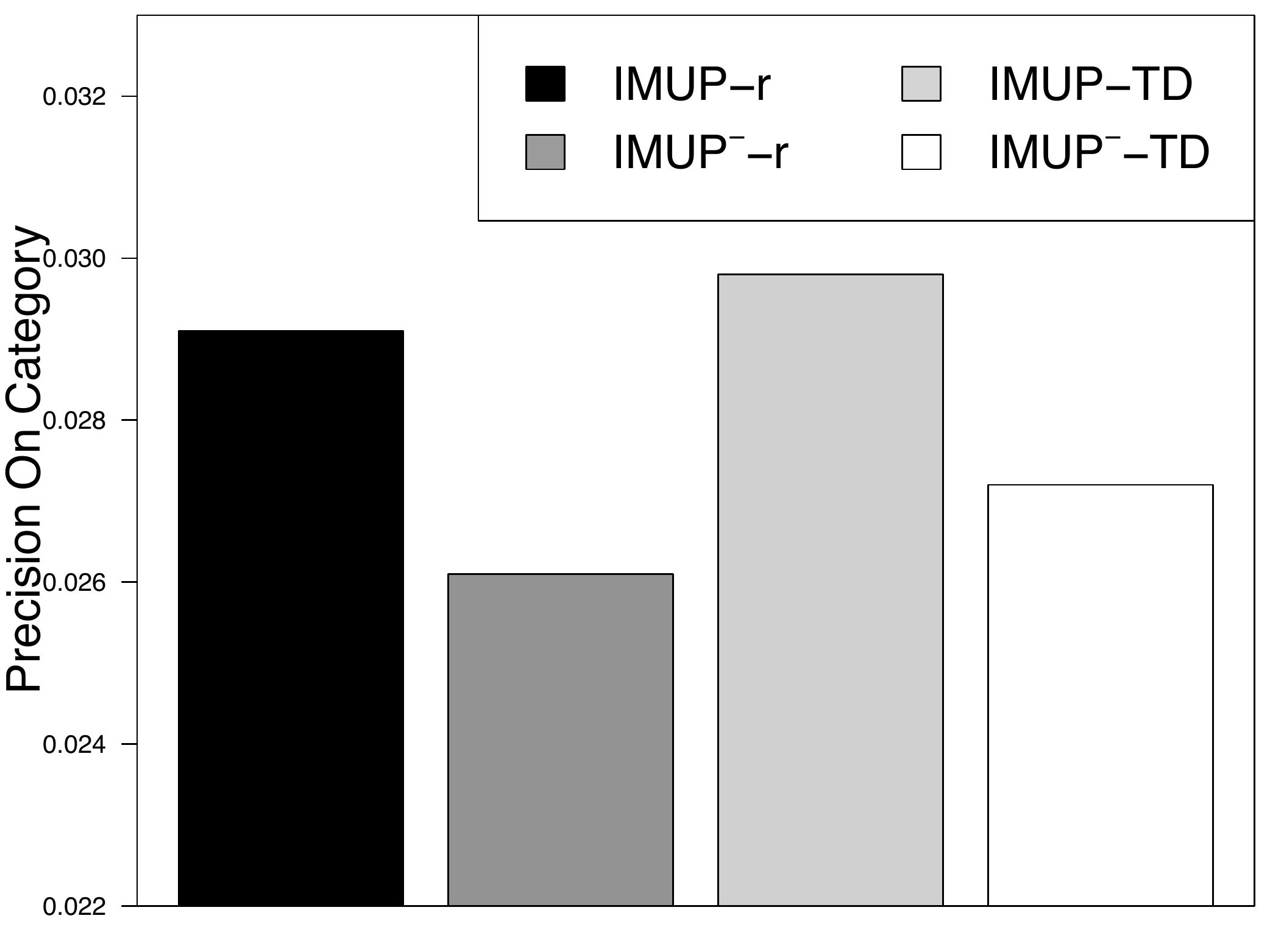}}
	\subfigure[Recall on Category]{\label{fig:substructure_newprecision_newyork}\includegraphics[width=4.35cm]{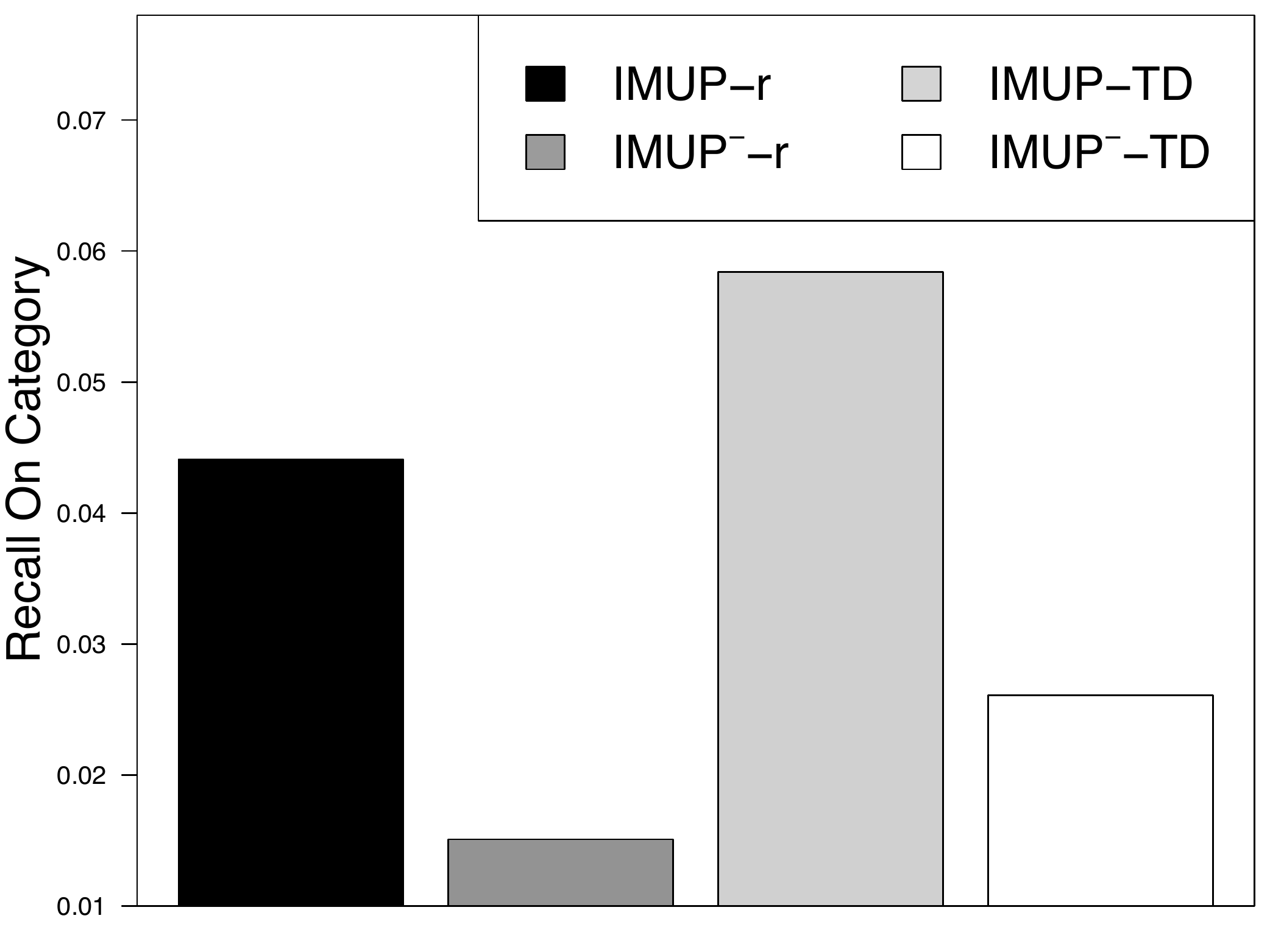}}
	\subfigure[Average Similarity]{\label{fig:substructure_precision_tokyo}\includegraphics[width=4.35cm]{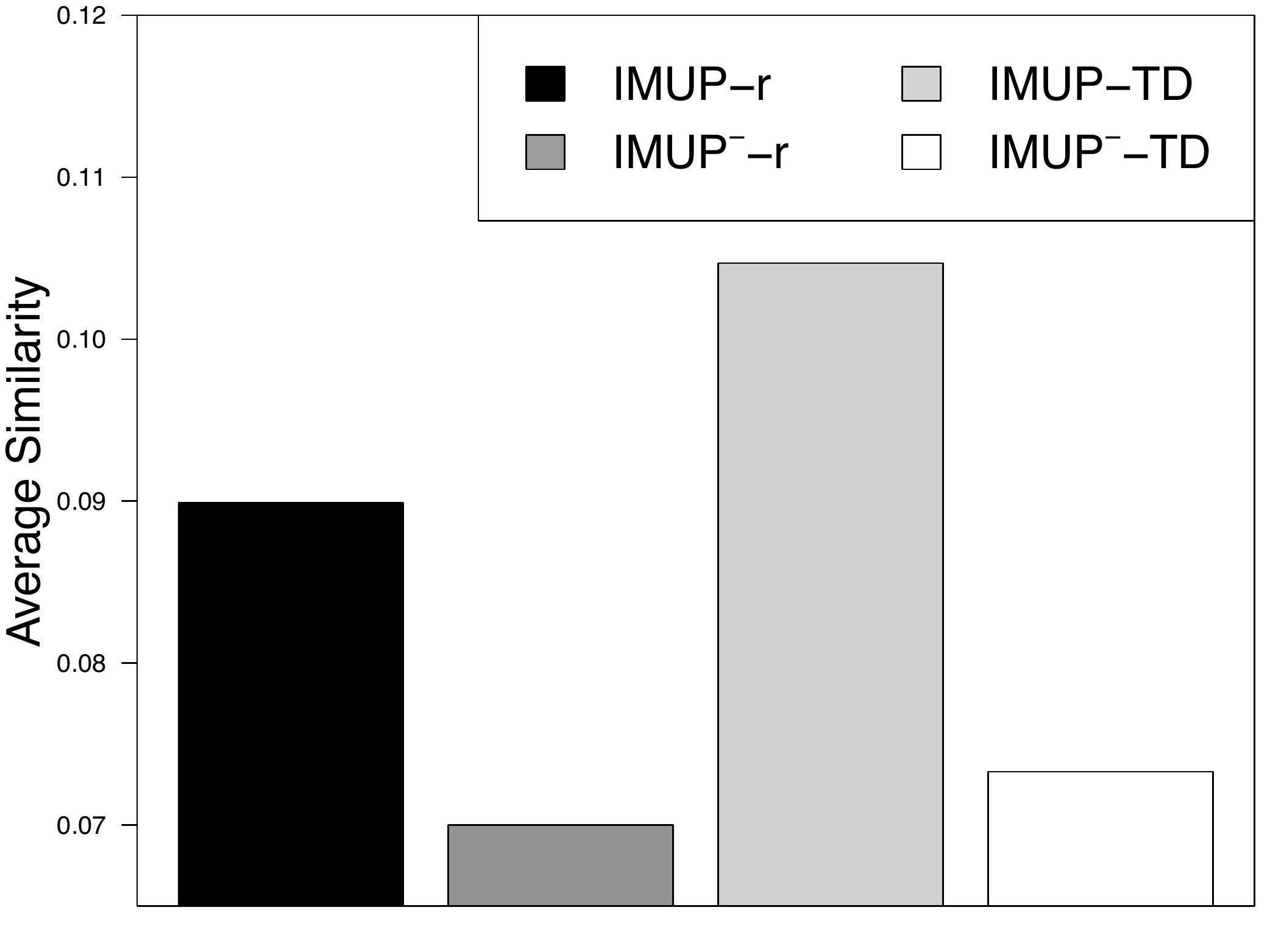}}
	\subfigure[Average Distance]{\label{fig:substructure_newprecision_tokyo}\includegraphics[width=4.35cm]{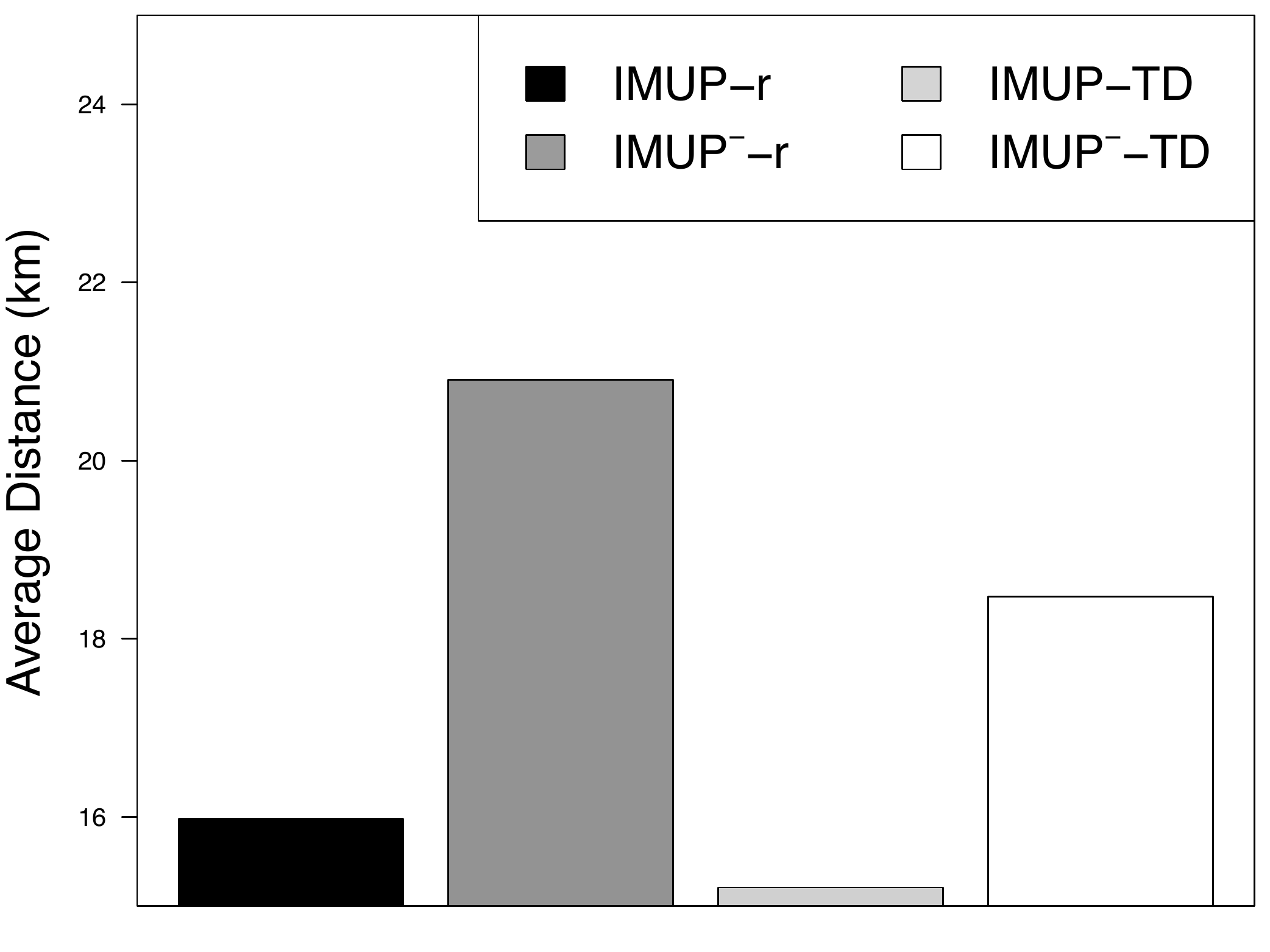}}
 		\vspace{-0.4cm}
	\captionsetup{justification=centering}
	\caption{Analysis of spatial {\it KG} {\it w.r.t.} New York dataset.}
 		\vspace{-0.4cm}
	\label{fig:KG nyc}
\end{figure*}

\begin{figure*}[!thb]
	\centering
	\subfigure[Precision on Category]{\label{fig:substructure_precision_newyork}\includegraphics[width=4.35cm]{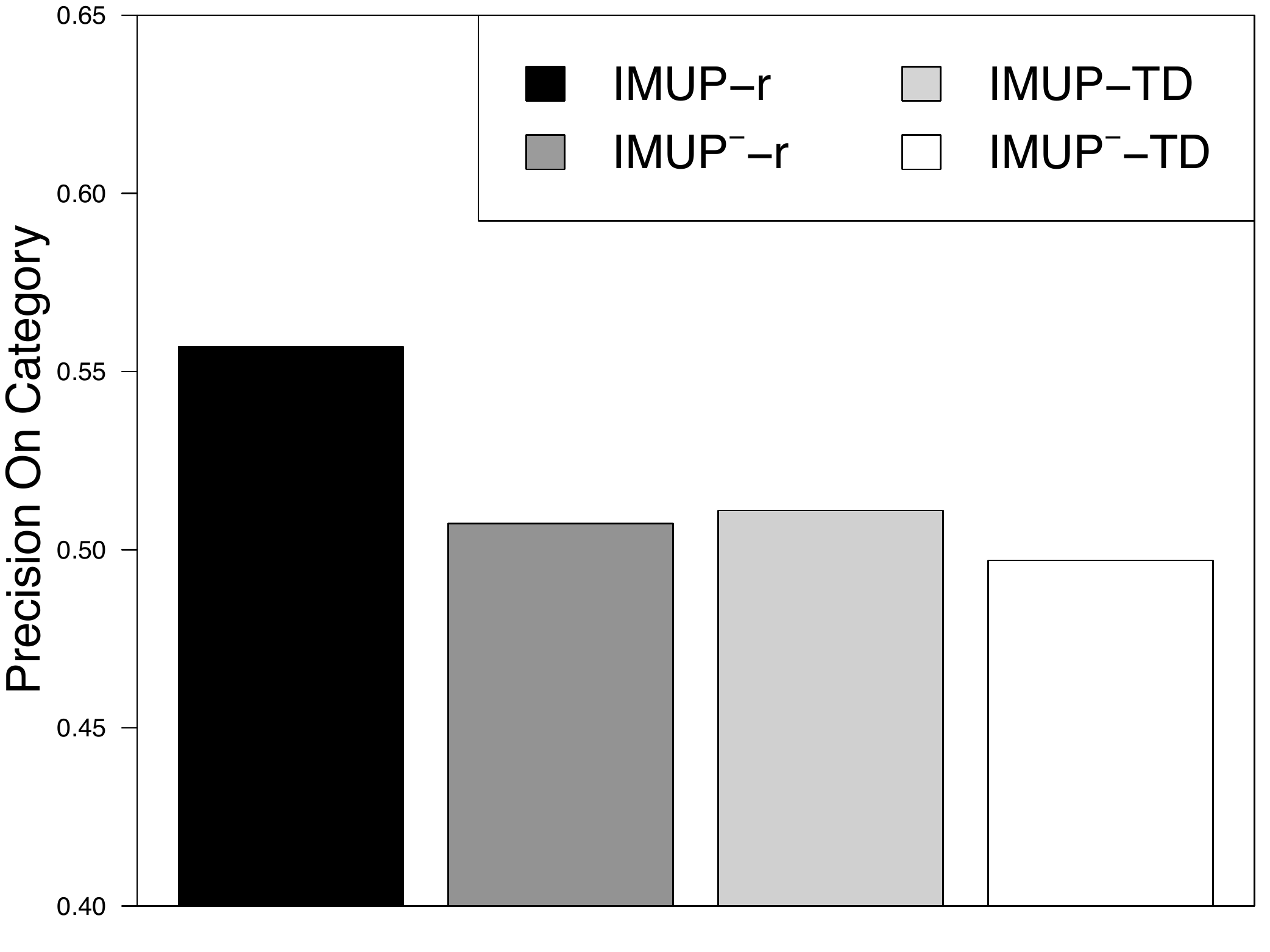}}
	\subfigure[Recall on Category]{\label{fig:substructure_newprecision_newyork}\includegraphics[width=4.35cm]{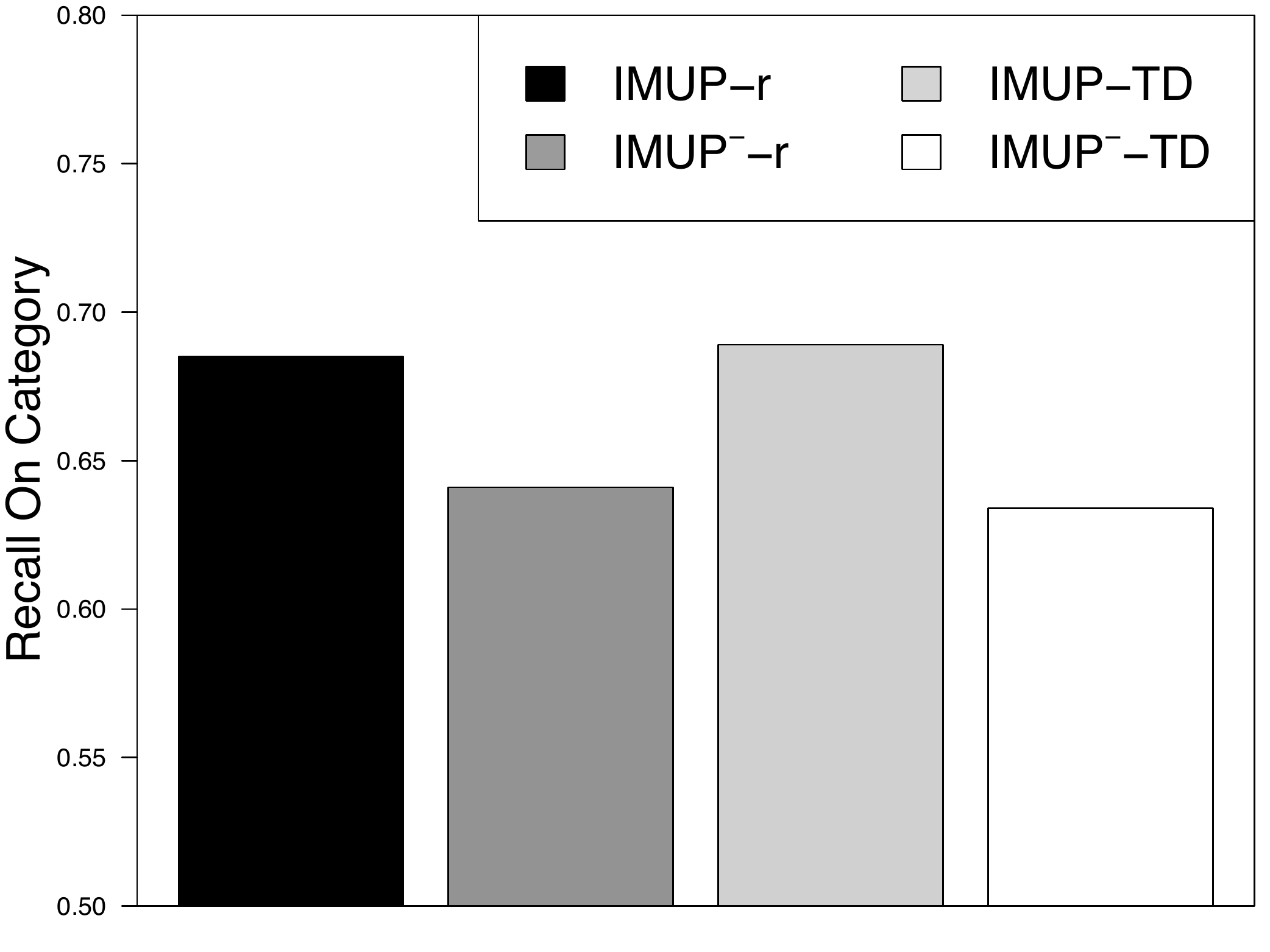}}
	\subfigure[Average Similarity]{\label{fig:substructure_precision_tokyo}\includegraphics[width=4.35cm]{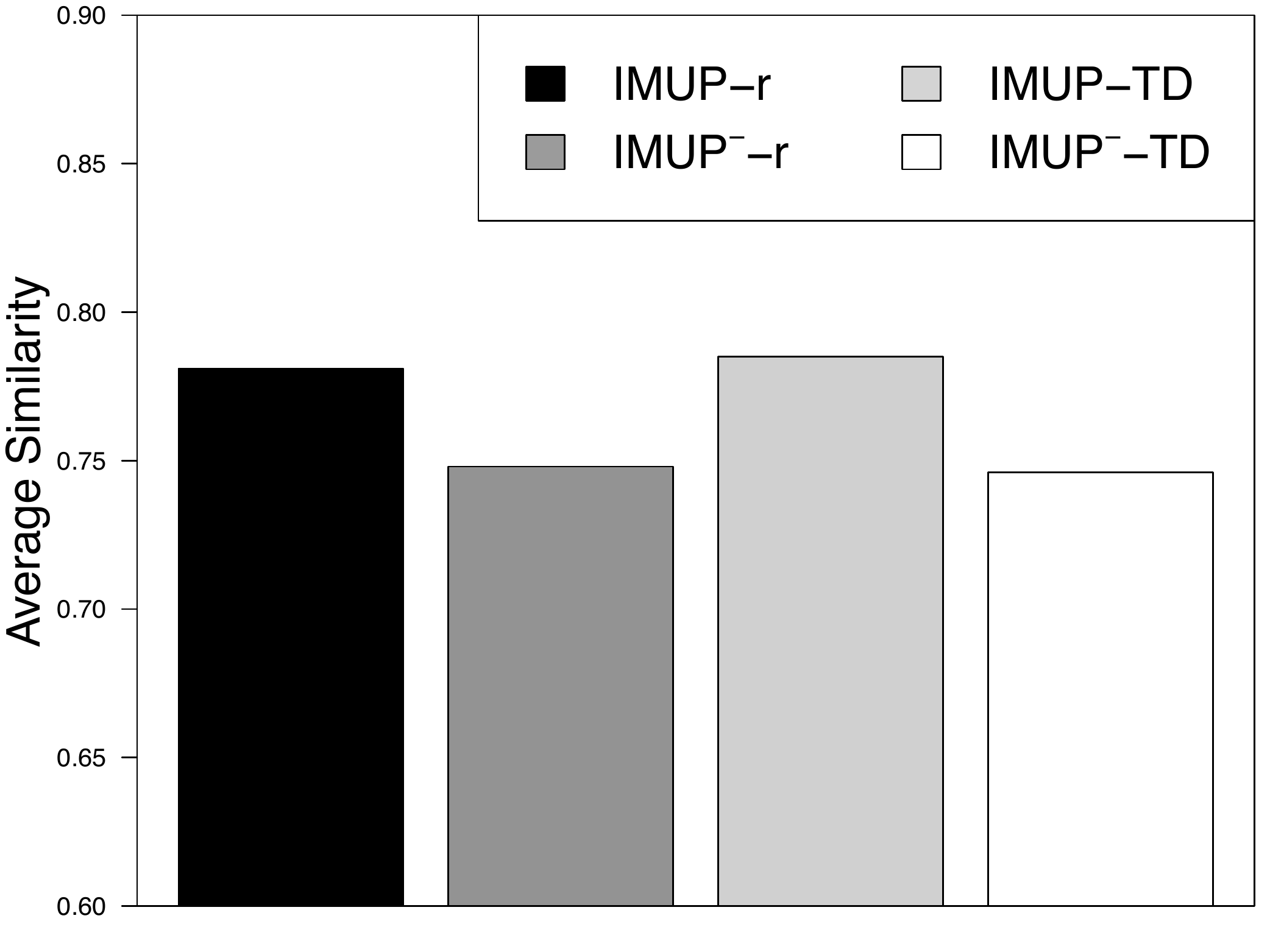}}
	\subfigure[Average Distance]{\label{fig:substructure_newprecision_tokyo}\includegraphics[width=4.35cm]{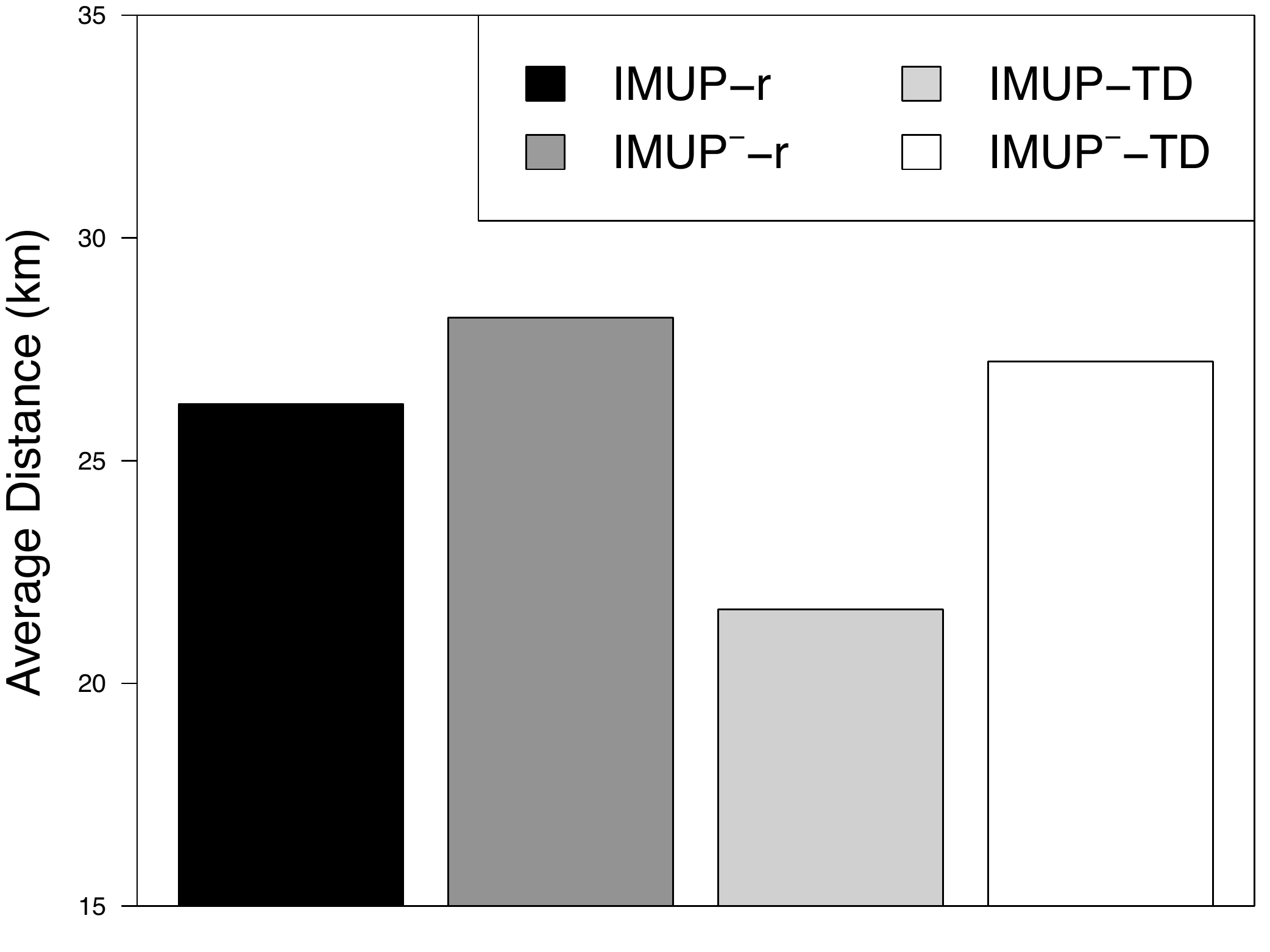}}
 		\vspace{-0.4cm}
	\captionsetup{justification=centering}
	\caption{Analysis of spatial {\it KG} {\it w.r.t.} Beijing dataset.}
 		\vspace{-0.35cm}
	\label{fig:KG bj}
\end{figure*}

\subsection{Comparison of different state update strategies}

We compare our newly proposed state update strategy ({\it i.e.}, long-short term influence of interactions) with the one from the conference version ({\it i.e.}, incremental update without differentiating the importance of interactions).
Specifically, we use $up1$ and $up2$ to denote the state update strategy of the conference version and the one newly proposed in this work respectively. 
Based on two sampling strategies, we further develop four variants of IMUP, namely "IMUP$-r^{up1}$'',"IMUP$-r^{up2}$'',
"IMUP$-TD^{up1}$'',"IMUP$-TD^{up2}$''.

Figure \ref{fig:diff_update_nyc}
and Figure \ref{fig:diff_update_bj} show the comparison results in terms of ``Precision on Category'', ``Recall on Category'', ``Average Similarity'', and ``Average Distance''.
We can observe that $up2$ is better than $up1$ in terms of majority evaluation metrics and fluctuates in a few ones on both New York and Beijing datasets.
A potential reason is that the long-short term influenced based state update strategy can discard old and ambiguous user visit preferences and capture the newest user interests in time. 
As a result, the state update strategy of this work can help the reinforced agent imitate user mobility behavior quickly and perfectly.

\begin{table}[!t]
  \caption{Comparison of different state initialization methods \textit{w.r.t.} New York dataset}
  \vspace{-0.4cm}
  \begin{center}
  \scalebox{0.98}{
    \begin{tabular}{c|c|c|c|c|c}
      \hline
       & StructRL & GAE & VGAE &  ARGA & ARGVA \\
      \hline
      Prec\_CAT & 0.02216 & 0.02859 & 0.02955 & 0.02628 & 0.02902   \\ \hline
      Rec\_CAT & 0.04942 & 0.04773 & 0.05465 &  0.05461 &  0.05364 \\ \hline
      Avg\_Sim & 0.09892 & 0.08328 & 0.1039 & 0.1039 & 0.09131  \\ \hline
     Avg\_Dist & 14.8151 & 13.2210 & 15.2624 & 16.7266 & 14.3054 \\ \hline
    \end{tabular}
    }
  \end{center}
  \label{tab:nyc_state_vary}
  \vspace{-0.35cm}
\end{table}

\begin{table}[!t]
  \caption{Comparison of different state initialization methods  \textit{w.r.t.} Beijing dataset}
  \vspace{-0.4cm}
  \begin{center}
  \scalebox{0.98}{
    \begin{tabular}{c|c|c|c|c|c}
      \hline
       & StructRL & GAE & VGAE  & ARGA & ARGVA \\
      \hline
      Prec\_CAT & 0.48019 & 0.47945 & 0.49185 & 0.49783 & 0.47597 \\ \hline
      Rec\_CAT & 0.69242 & 0.69242 & 0.67542 &  0.66461 &  0.66615 \\ \hline
      Avg\_Sim & 0.78636 & 0.78648 & 0.77479 & 0.76637 & 0.76746 \\ \hline
     Avg\_Dist & 20.7080 & 21.2318 & 29.8552 & 30.1306 & 22.9128 \\ \hline
    \end{tabular}
    }
  \end{center}
  \label{tab:bj_state_vary}
  \vspace{-0.5cm}
\end{table}

\subsection{Comparison of different state initialization }
We adopt StructRL to convert user mobility graphs to  representations, which are used to initialize the user state of our framework.
To study the influence of state initialization strategies on user profiling, we conduct the comparative experiment based on the journal version framework (IMUP$^*$) by replacing the StructRL with GAE~\cite{kipf2016variational}, VGAE~\cite{kipf2016variational}, ARGA~\cite{pan2018adversarially}, and ARGVA~\cite{pan2018adversarially} respectively.
Table ~\ref{tab:nyc_state_vary} and Table ~\ref{tab:bj_state_vary} show the comparison results in terms of ``Precision on Category'', ``Recall on Category'', ``Average Distance'', and ``Average Similarity''.
We can find that the next-visit prediction performance fluctuates slightly under different state initialization strategies.
This demonstrates the robustness of IMUP$^*$ against distinct state initialization methods.
A potential reason is that the state update strategy is able to discard the  old and useless information and capture the new and significant characteristics of user mobility behaviors in time.
As the progress of the learning process, the  information in the initial state will be replaced with new user mobility characteristics, resulting in the robustness of IMUP$^*$.

\subsection{Analysis of spatial {\it KG}}
We evaluate the spatial {\it KG}'s contribution on modeling user representations.
To set the control group, we develop a variant of the proposed ``IMUP'', namely ``IMUP$^{-}$''.
``IMUP$^{-}$'' takes only the user and POI representations as the environment state, while other components of remains the same.
The POI state (representations) is randomly initialized and state update is still based on the mutual interactions, but Equation~\ref{equ:kg} and Equation~\ref{equ:other poi} are omitted.

Figure~\ref{fig:KG nyc} and Figure~\ref{fig:KG bj} show the comparison results.
We can observe that the performance of ``IMUP'' outperforms ``IMUP$^{-}$'' in terms of the four metrics over both two datasets.
The results validate that the integration of semantics from spatial {\it KG} indeed enhances the modeling of user preferences on visit event. In the meantime, the incrementally updated next-visit planner (agent) is forced to bring in semantics from spatial {\it KG} to mimic the personalized user patterns, paced with the incrementally updated user representations (state).

\begin{figure*}[!tb]
	\centering
	\subfigure[Precision on Category]{\label{fig:substructure_precision_newyork}\includegraphics[width=4.35cm]{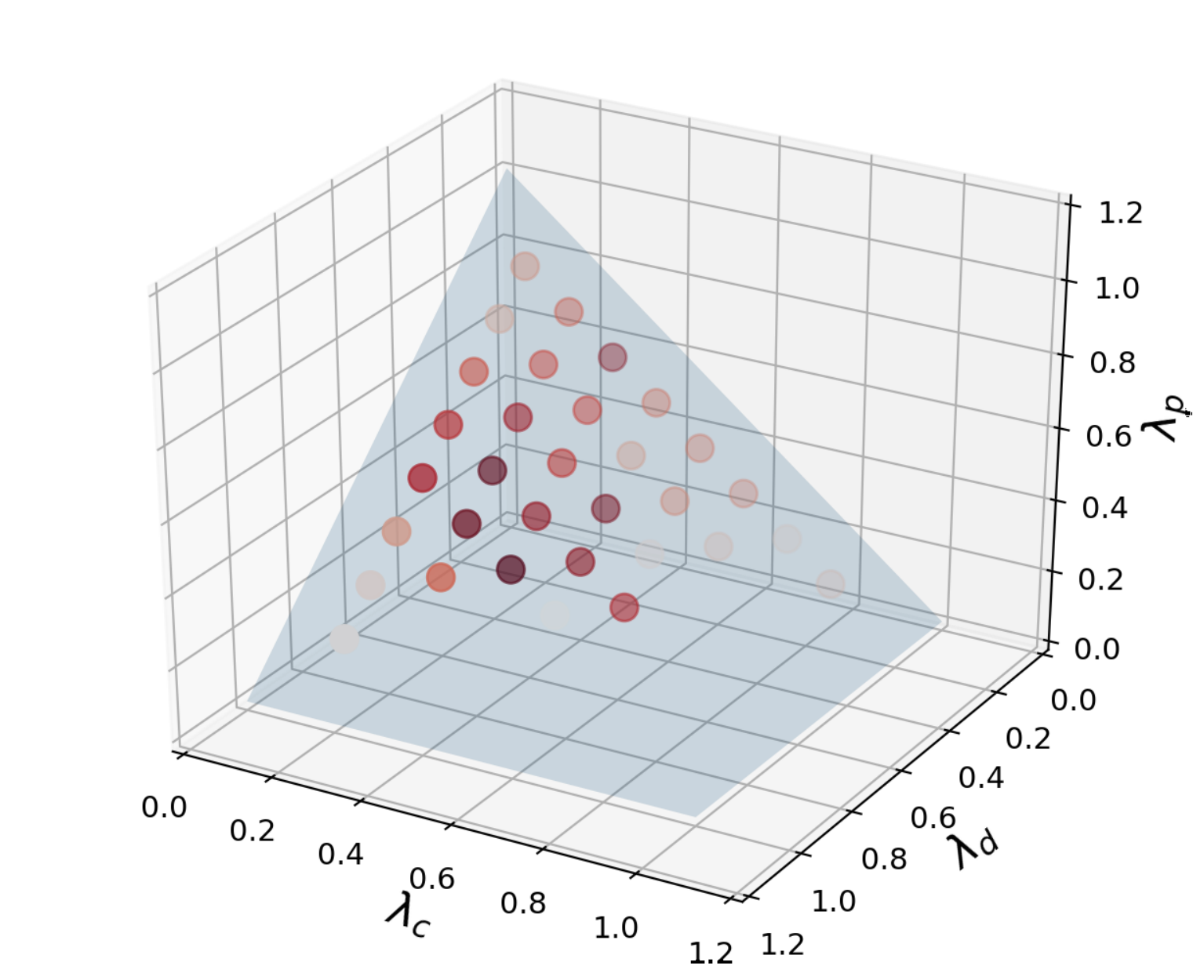}}
	\subfigure[Recall on Category]{\label{fig:substructure_newprecision_newyork}\includegraphics[width=4.35cm]{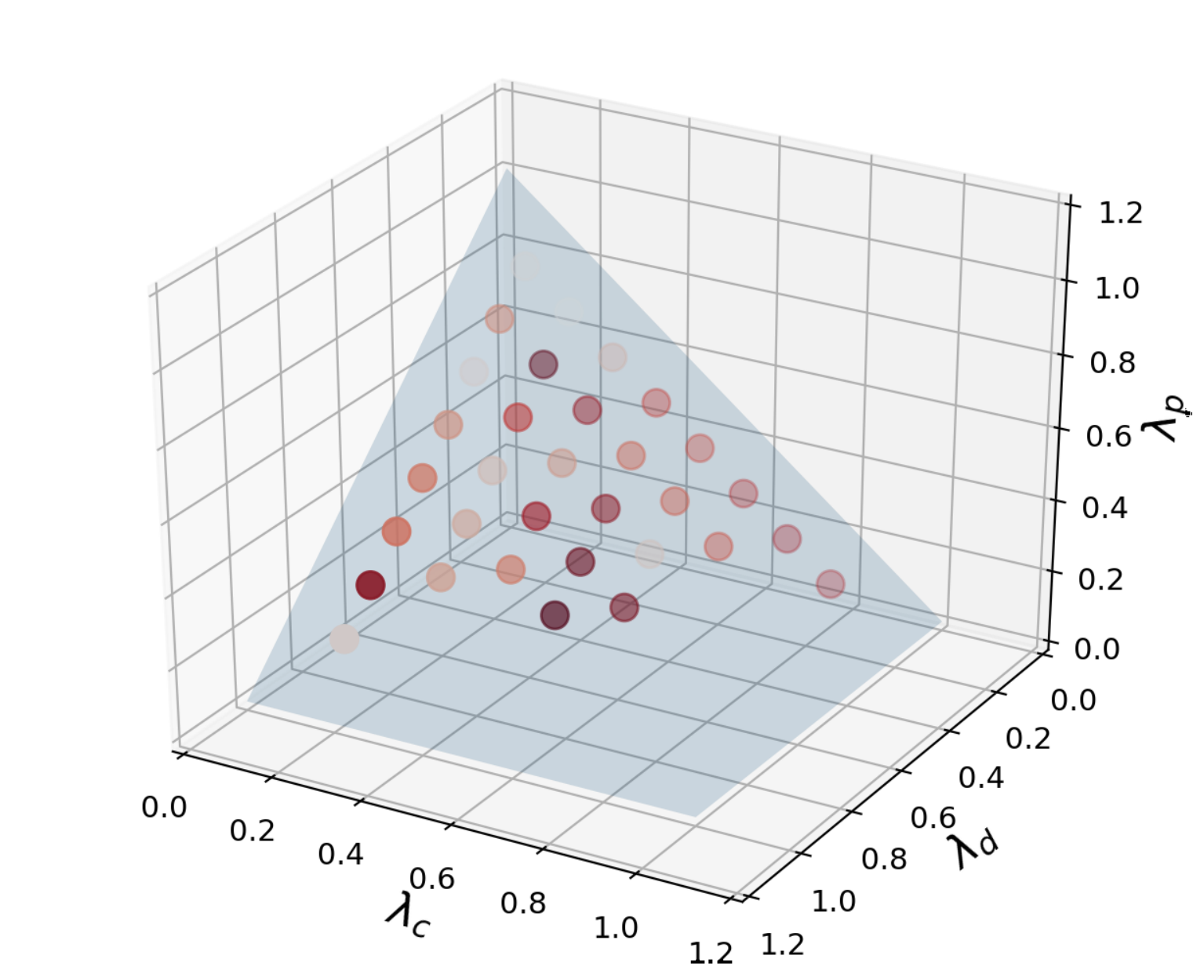}}
	\subfigure[Average Similarity]{\label{fig:substructure_precision_tokyo}\includegraphics[width=4.35cm]{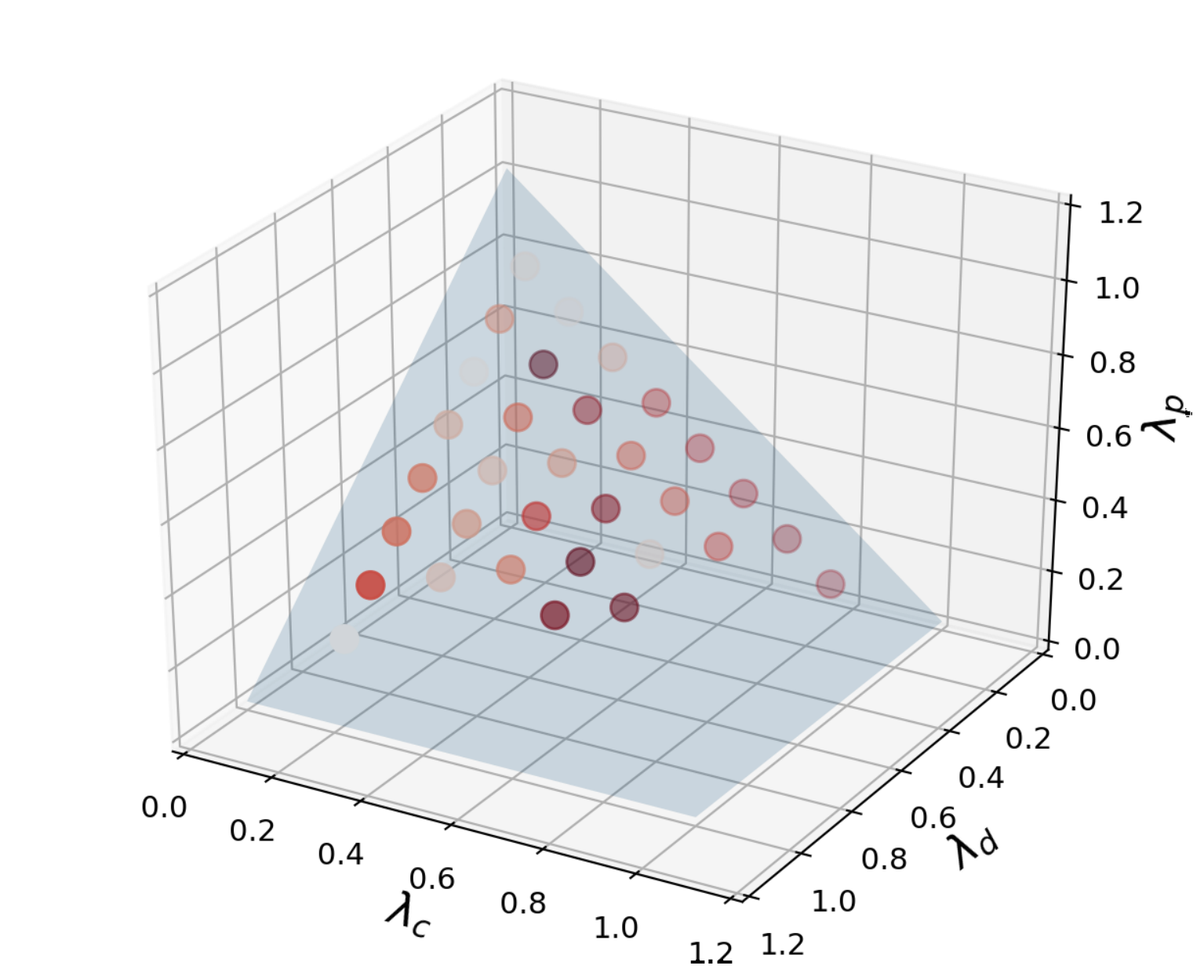}}
	\subfigure[Average Distance]{\label{fig:substructure_newprecision_tokyo}\includegraphics[width=4.35cm]{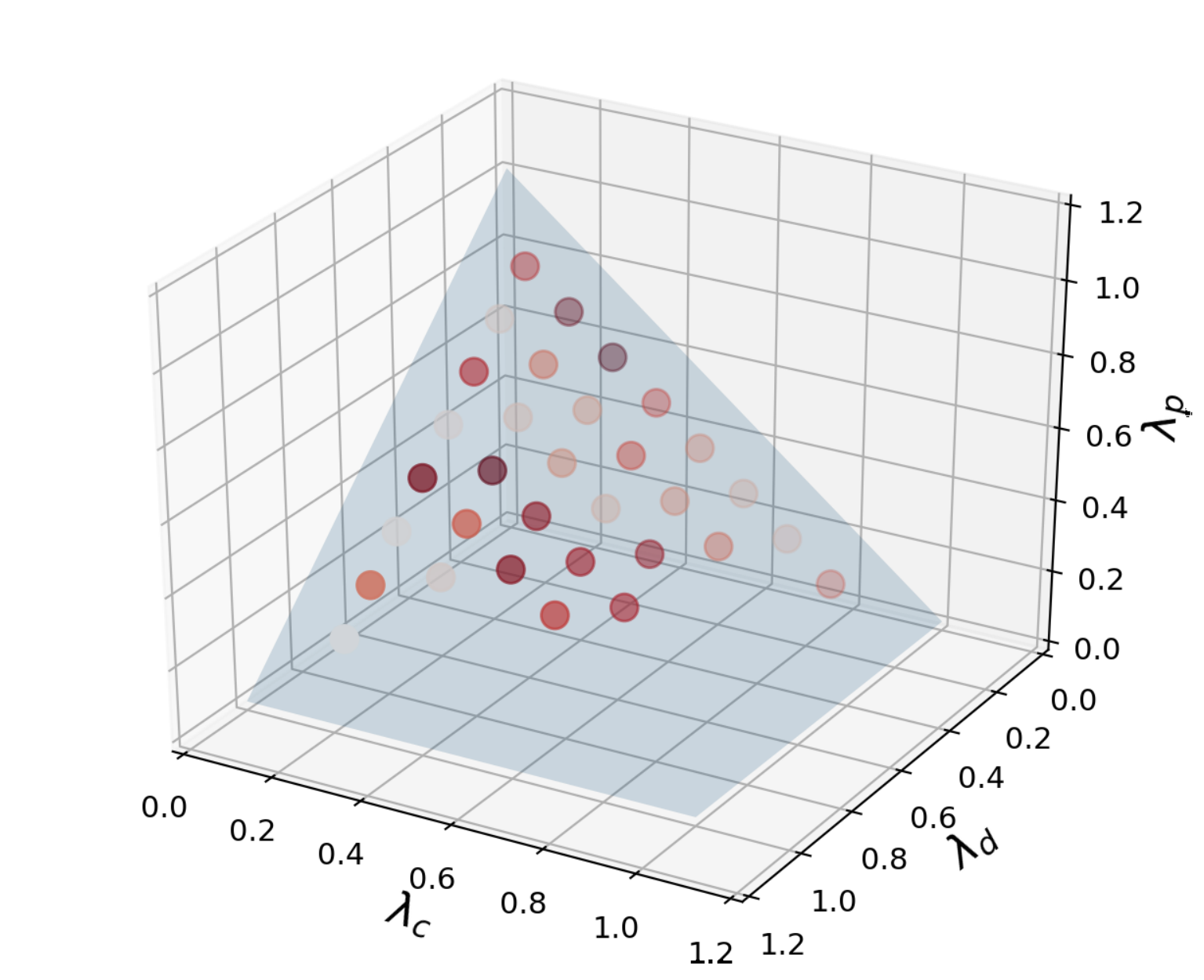}}
 	\vspace{-0.4cm}
	\captionsetup{justification=centering}
	\caption{Reward analysis of $r1$ {\it w.r.t.} New York dataset.}
 	\vspace{-0.4cm}
	\label{fig:reward nyc}
\end{figure*}

\begin{figure*}[!tb]
	\centering
	\subfigure[Precision on Category]{\label{fig:substructure_precision_newyork}\includegraphics[width=4.35cm]{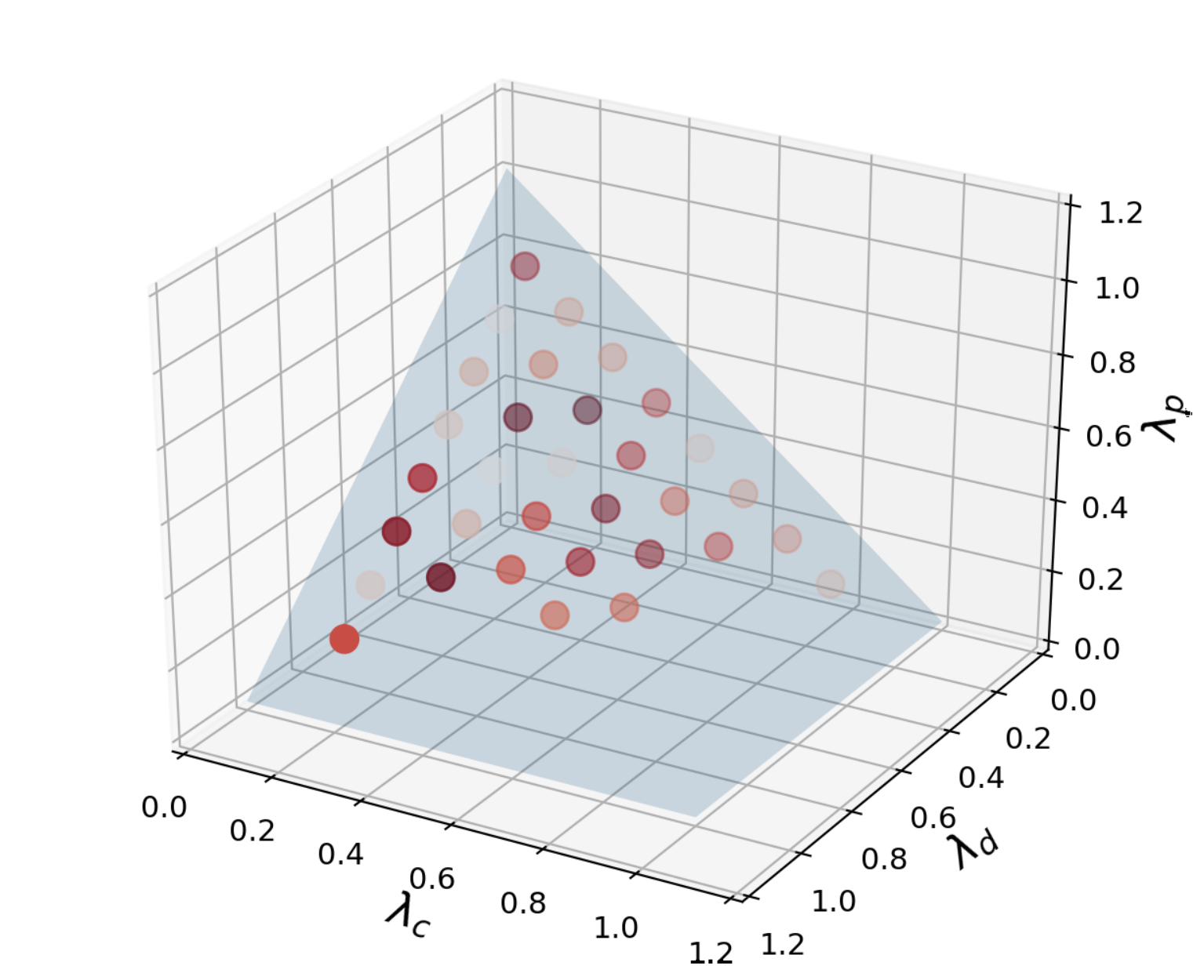}}
	\subfigure[Recall on Category]{\label{fig:substructure_newprecision_newyork}\includegraphics[width=4.35cm]{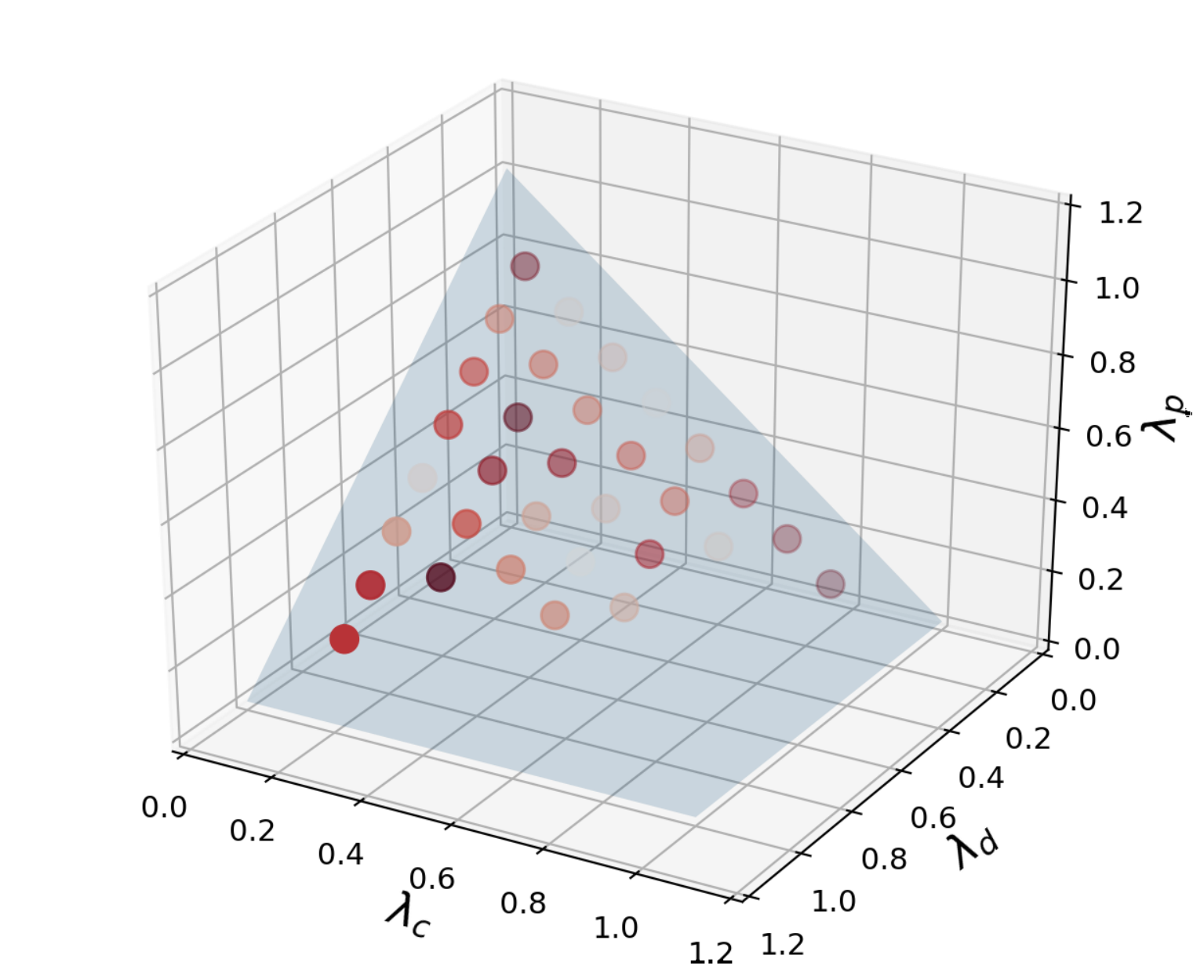}}
	\subfigure[Average Similarity]{\label{fig:substructure_precision_tokyo}\includegraphics[width=4.35cm]{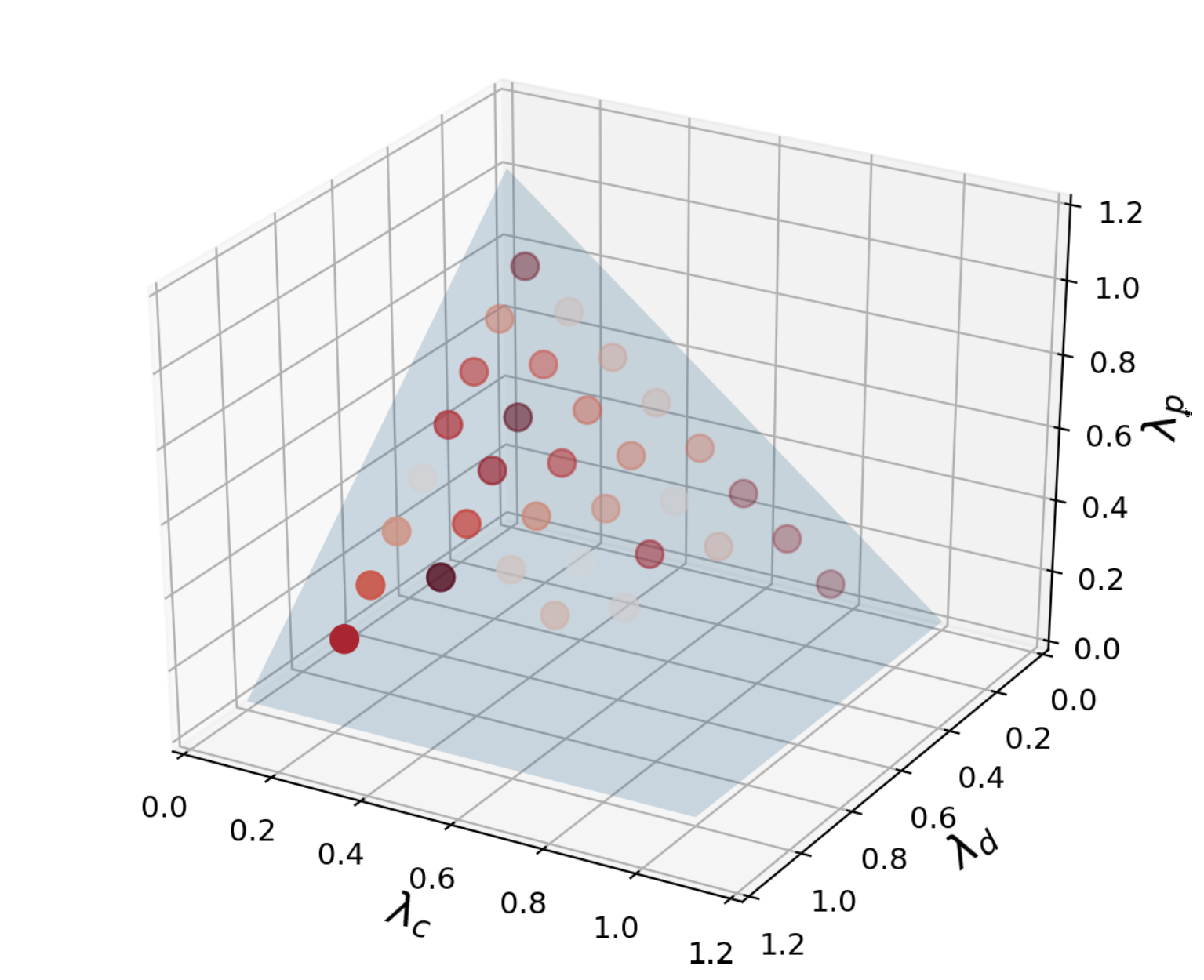}}
	\subfigure[Average Distance]{\label{fig:substructure_newprecision_tokyo}\includegraphics[width=4.35cm]{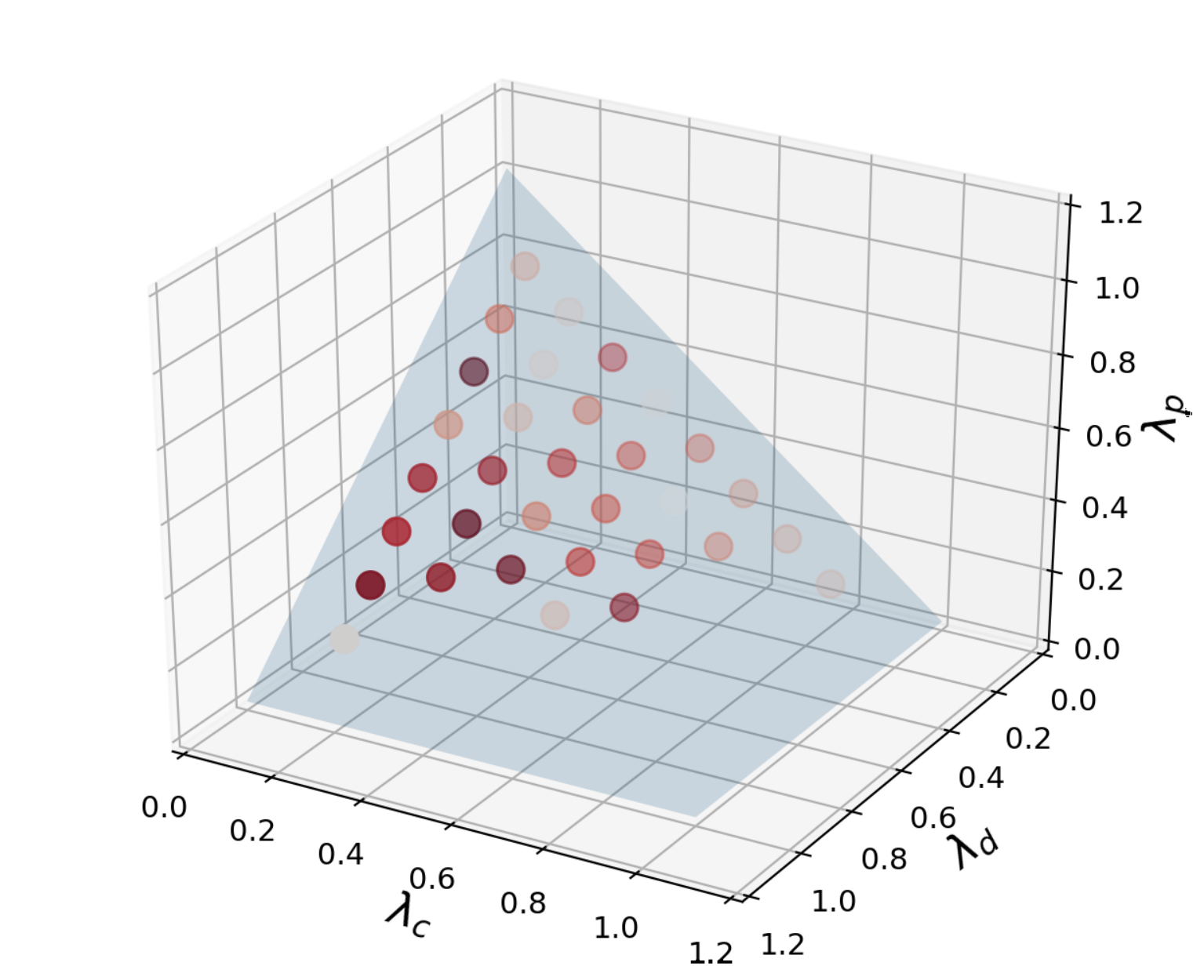}}
 	\vspace{-0.4cm}
	\captionsetup{justification=centering}
	\caption{Reward analysis of $r1$ {\it w.r.t.} Beijing dataset.}
 	\vspace{-0.4cm}
	\label{fig:reward bj}
\end{figure*}

\begin{figure*}[!tb]
	\centering
	\subfigure[Precision on Category]{\label{fig:substructure_precision_newyork}\includegraphics[width=4.35cm]{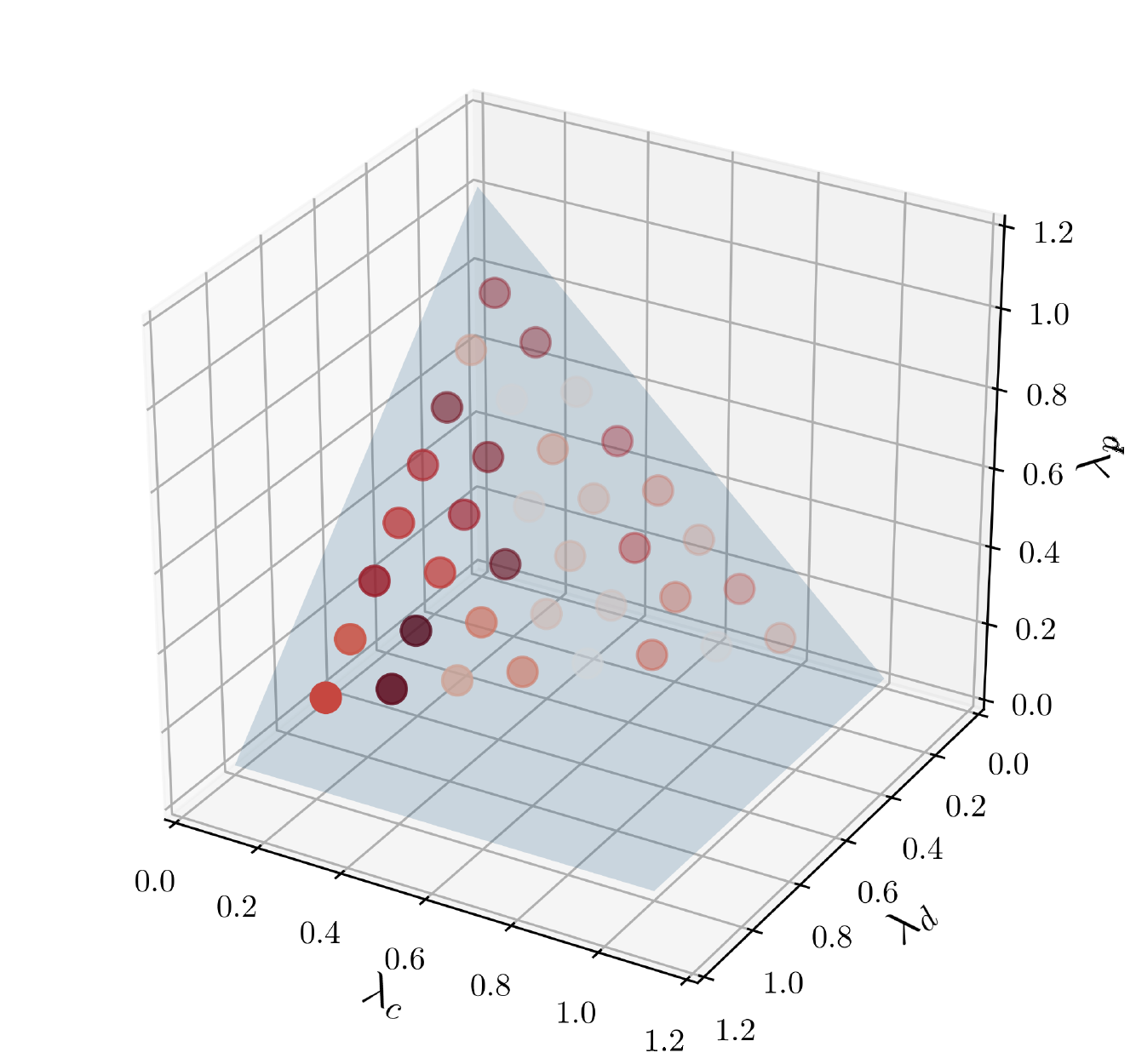}}
	\subfigure[Recall on Category]{\label{fig:substructure_newprecision_newyork}\includegraphics[width=4.35cm]{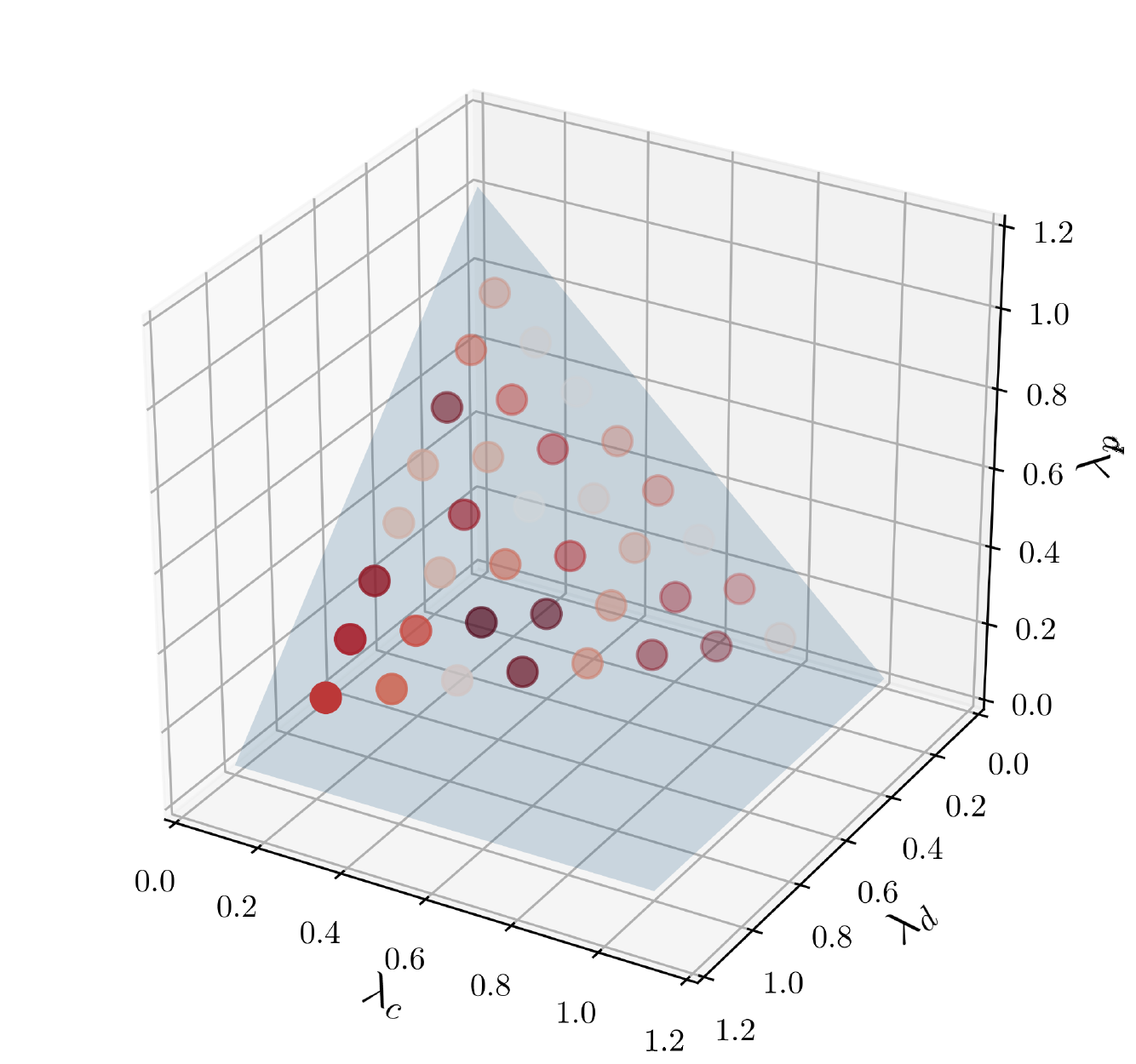}}
	\subfigure[Average Similarity]{\label{fig:substructure_precision_tokyo}\includegraphics[width=4.35cm]{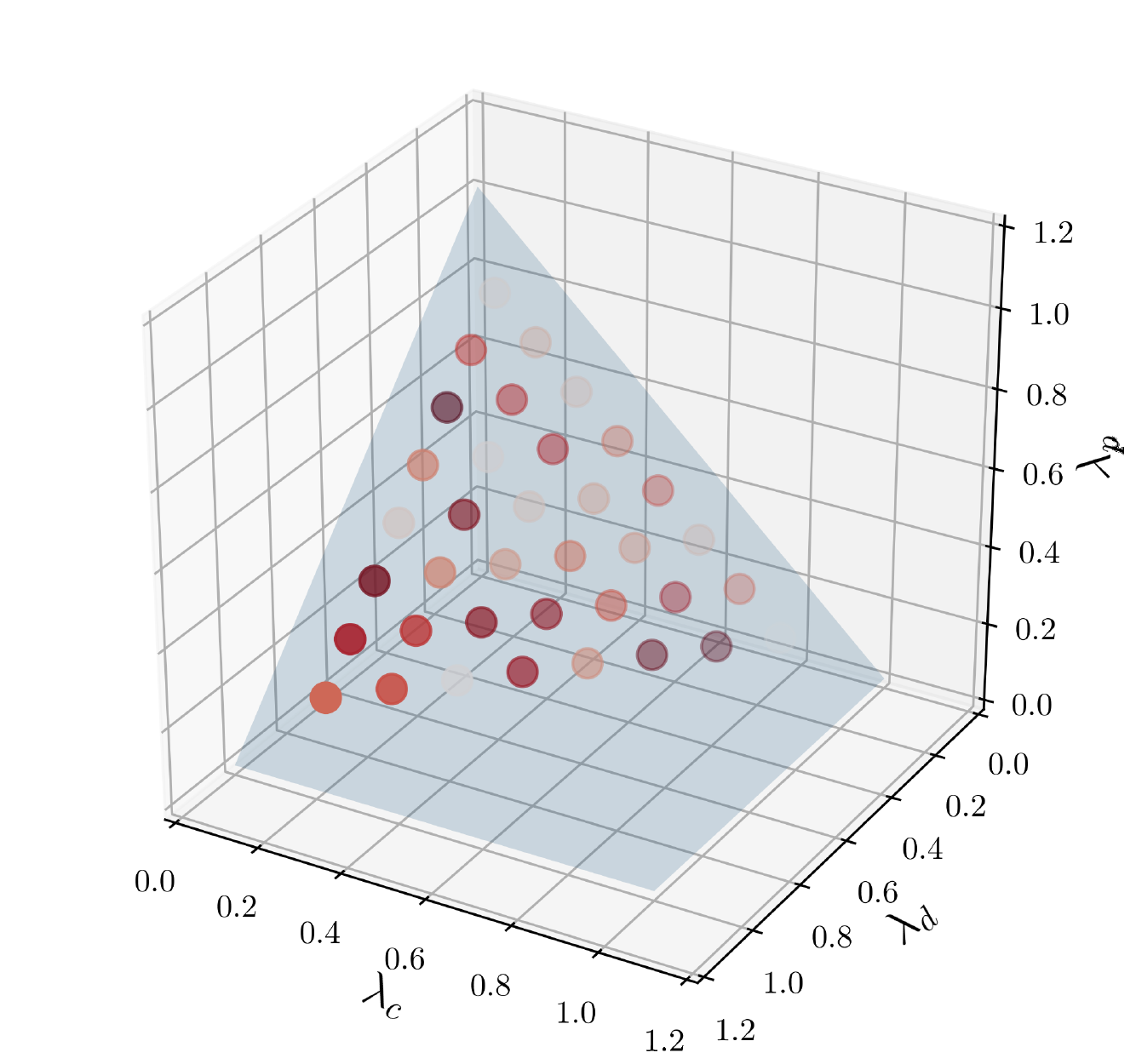}}
	\subfigure[Average Distance]{\label{fig:substructure_newprecision_tokyo}\includegraphics[width=4.35cm]{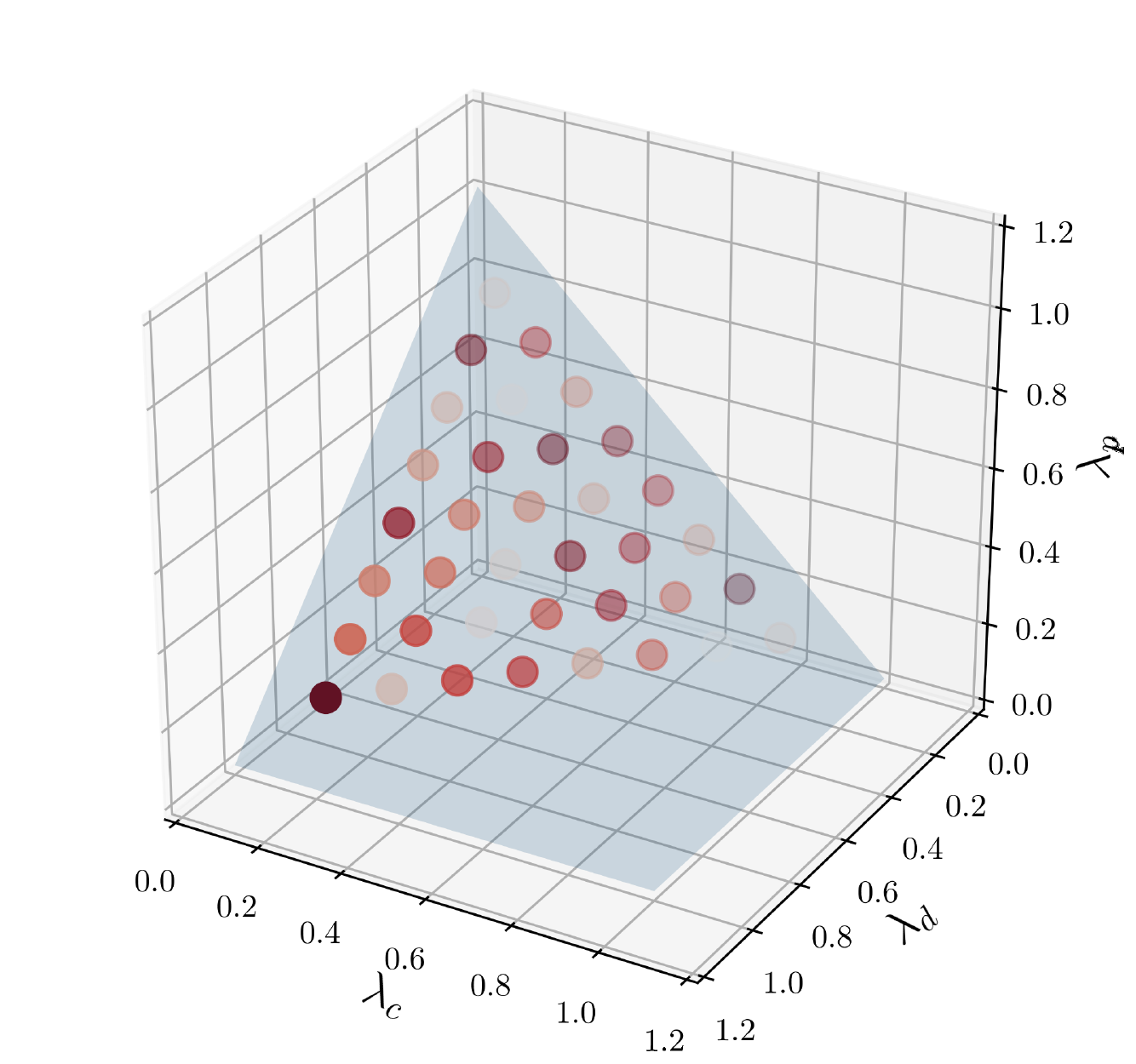}}
	\vspace{-0.4cm}
	\captionsetup{justification=centering}
	\caption{Reward analysis of $r2$ {\it w.r.t.} New York dataset.}
	\vspace{-0.4cm}
	\label{fig:r2 nyc}
\end{figure*}

\begin{figure*}[!tb]
	\centering
	\subfigure[Precision on Category]{\label{fig:substructure_precision_newyork}\includegraphics[width=4.35cm]{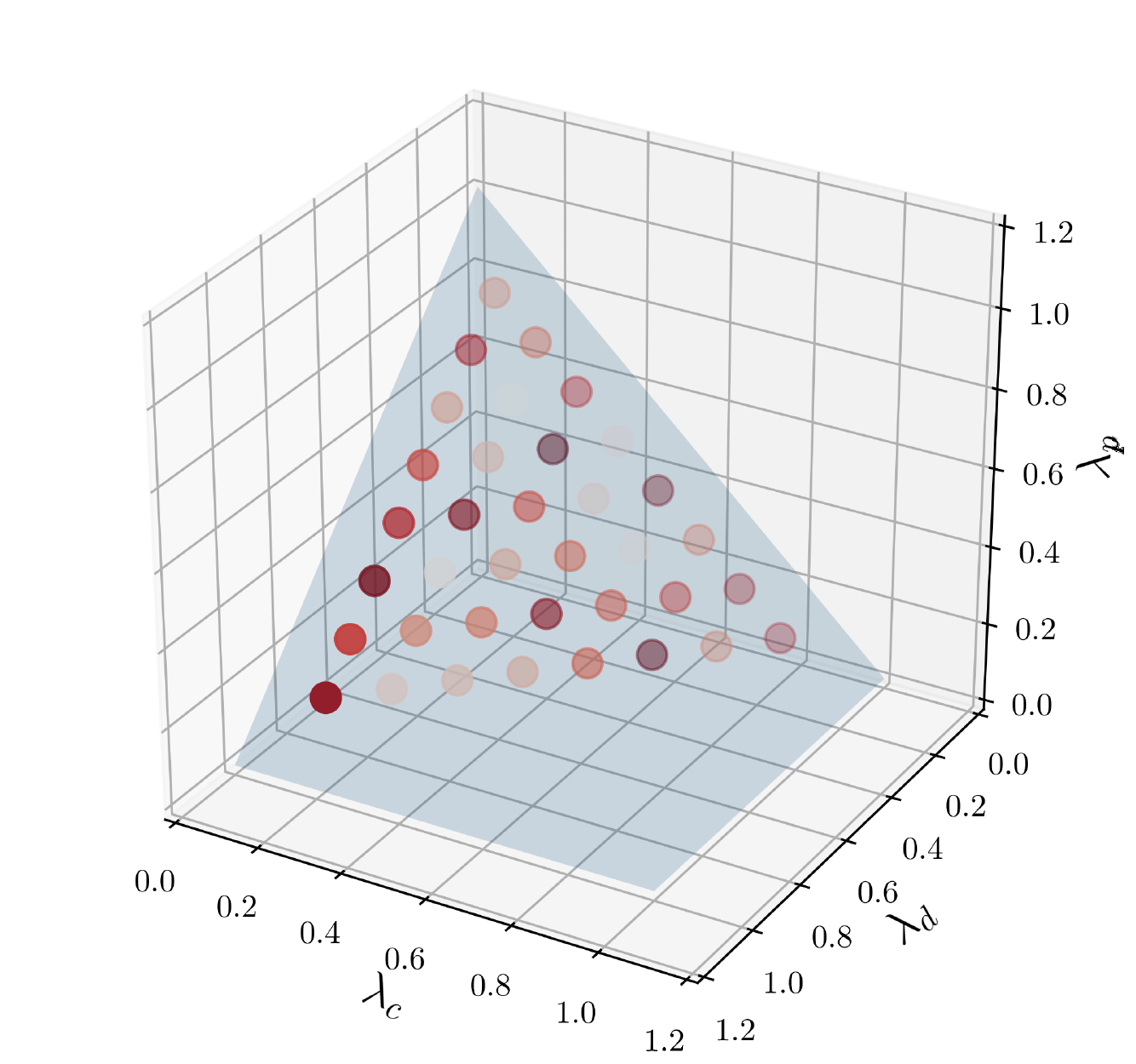}}
	\subfigure[Recall on Category]{\label{fig:substructure_newprecision_newyork}\includegraphics[width=4.35cm]{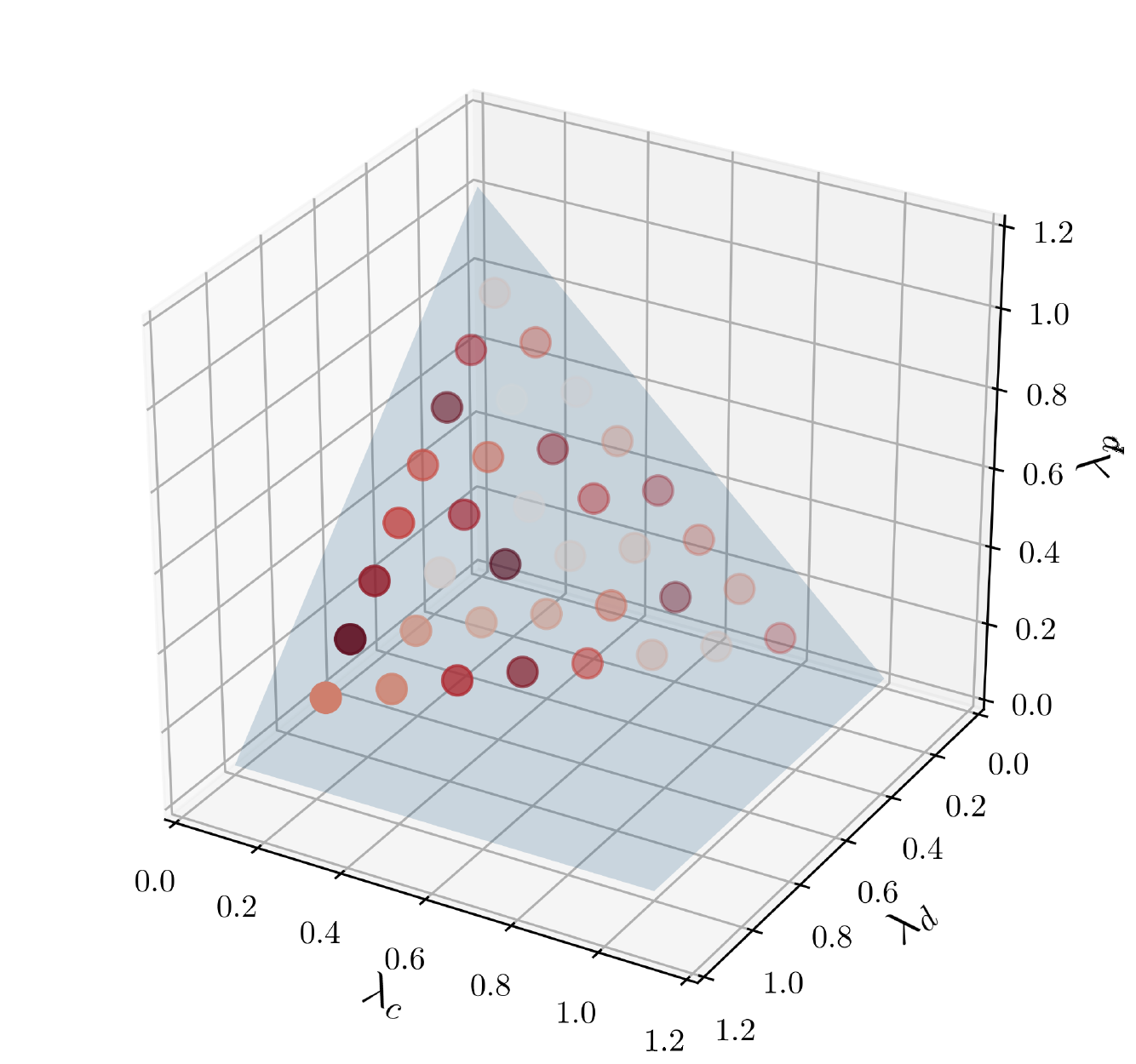}}
	\subfigure[Average Similarity]{\label{fig:substructure_precision_tokyo}\includegraphics[width=4.35cm]{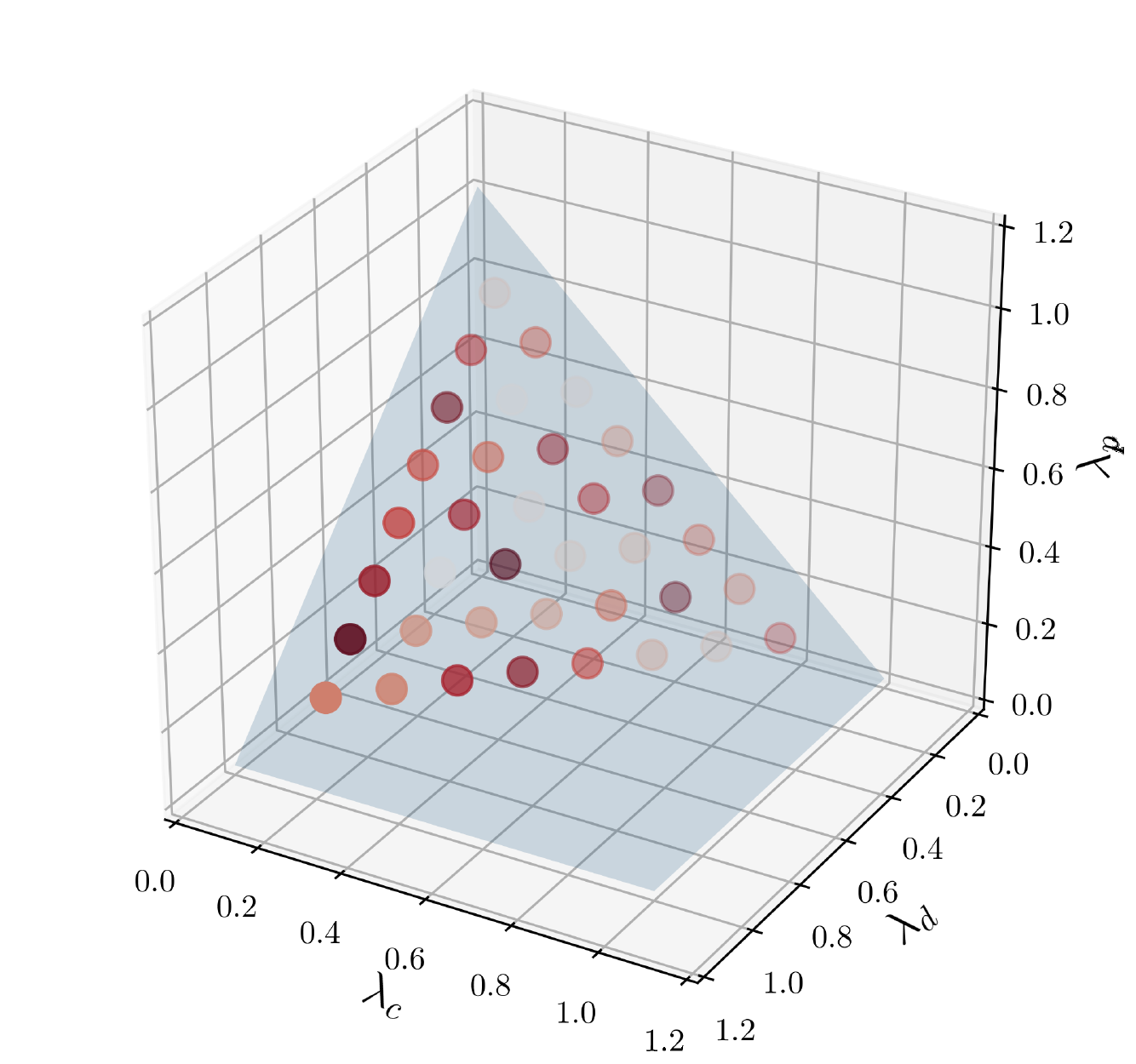}}
	\subfigure[Average Distance]{\label{fig:substructure_newprecision_tokyo}\includegraphics[width=4.35cm]{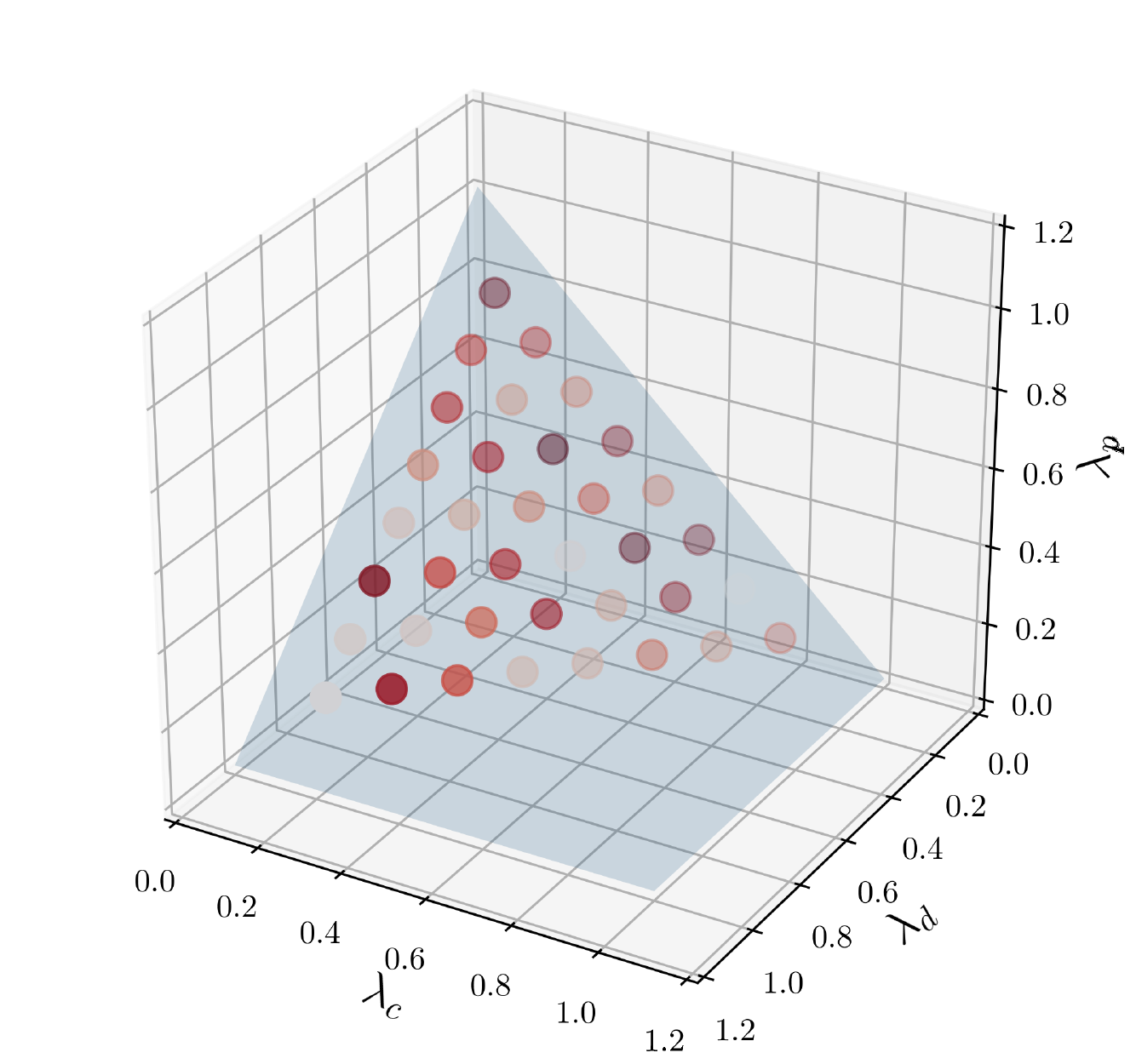}}
	\vspace{-0.4cm}
	\captionsetup{justification=centering}
	\caption{Reward analysis of $r2$ {\it w.r.t.} Beijing dataset.}
	\vspace{-0.5cm}
	\label{fig:r2 bj}
\end{figure*}

\subsection{Analysis of rewards}
There are two types of rewards in the experiment: $r1$ and $r2$.
Each reward function includes three parts: (1) distance $r_d$, (2) category similarity $r_c$, and (3) POI  difference between the predicted and real user visit events $r_p$, where contributions are controlled by three weights $\lambda_d$, $\lambda_c$, and $\lambda_p$ respectively.
To analyze the contribution of these three factors, given the learning rate=$1e-5$, we project the results $(\lambda_d, \lambda_c, \lambda_p, \text{metric})$ into the 3D space, where the $x, y, z$ axis corresponds to each factor respectively.
We assign the gradient color to each point such that the better the performance over the metric, the darker the color is assigned.

Figure~\ref{fig:reward nyc} and Figure~\ref{fig:reward bj}
show the reward analysis of $r1$. An interesting observation is that in the case of ``Precision on Category'', ``Recall on Category'' and ``Average Similarity'', the contribution of category similarity $r_c$ is higher than the other two factors, but in the case of ``Average Distance'', the contributions of distance $r_d$ and POI $r_p$ surpass the category similarity $r_c$.
The reason is quite intuitive that the category similarity $r_c$ directly determines the direction  of policy training towards exploring more similar POI categories, while distance $r_d$ and POI $r_p$ guides the policy to find POIs as close as possible to the users' intention.

A careful inspection of Figure~\ref{fig:reward nyc} and Figure~\ref{fig:reward bj} suggests that although
the performance over certain metric is highly related to specific factors ({\it e.g.}, the category similarity $r_c$ is highly related to category-related performance), the best results are not achieved at the extreme case such that some factors are pushed to zeros.
On the contrary, the best results are achieved at the balance of these three factors, which may reveal some dependencies among POI locations and categories introduced by the spatial {\it KG}.

Figure~\ref{fig:r2 nyc} and Figure ~\ref{fig:r2 bj} show the result of reward analysis for the reward function $r2$.
An interesting observation is that compared with $r1$, the dark points all shift to the left side for all evaluation metrics.
The observation means that in order to obtain better predictions, we need to increase the contribution of $r_d$ by increasing the value of $\lambda_d$.
A potential reason is after subtracting baseline expectations, the value of $r_d$ is lower than $r_c$ and $r_p$.
As a result, the agent considers fewer geographical factors of user mobility habits, resulting in 
poor next-visit prediction performance.
To accurately simulate user mobility, we must consider the geographical and semantic characteristics of user mobility completely.
Thus, we can improve the contribution of $r_d$ by increasing the value of $\lambda_d$.



\section{Related Work}
\noindent{\bf Mobile User Profiling.}
Our work is connected to Mobile user profiling.
User profiling refers to quantifying users' characteristics~\cite{wang2019adversarial}.
User profiling methods can be categorized into two groups: (1) explicit extraction, in which user profiles are explicitly predefined on documents, and (2) learning-based approach, which focus on learning user representations from users' historical behavior data~\cite{wang2019adversarial,wang2021reinforced}.
Our work is especially related to the learning-based approach for mobile users.
For example, factorization-based approaches are exploited to model the integration of geographical and temporal influences of human mobility behaviors~\cite{griesner2015poi,lian2014geomf};
deep learning-based approaches are then proposed to learn latent representations of users by leveraging the power of deep neural networks~\cite{yang2017bridging,yin2017spatial};
more advanced techniques ({\it e.g.}, adversarial learning) are further introduced with emphasis on substructures of user mobility patterns~\cite{wang2019adversarial,wang2020exploiting}.

\noindent{\bf Reinforcement Learning for Online User Modeling.}
Our work is related to reinforcement learning for online user modeling.
Reinforcement learning-based algorithms model online user behavior ({\it e.g.}, clicks, reviews, purchases) by regarding the online behavior as a sequential decision-making process~\cite{zhang2019deep,wang2022online,wang2022multi}.
For example, 
Zhao {\it et al.} propose the model ``DEERS'' to model both the positive and negative feedback through learning the sequential interactions via the reinforcement learning, with recommending trial-and-error items and receiving reinforcements of these items from users' feedback~\cite{zhao2018recommendations}.
Zheng {\it et al.} propose a reinforcement learning framework to provide personalized new recommendation by considering user return pattern to capture implicit user feedback and provide new news by effective exploring strategies~\cite{zheng2018drn}.
Chen {\it et al.} propose to utilize Markov Decision Process to model the sequential interactions between users and online recommender systems, and employing reinforcement learning to force an optimal policy for generating recommendation~\cite{choi2018reinforcement}. 
Isshu {\it et al.} exploits the multi-armed bandit approach to solve the cold-start problem based on the rewards from testing each options~\cite{munemasa2018deep}.
\section{Conclusion Remarks}
In this paper, we propose a new learning criteria, namely reinforced imitative graph learning (RIGL), for mobile user profiling.
Different from the minimization of traditional reconstruction loss or prediction loss, the proposed RIGL leverages an agent to mimic users' patterns and preferences.
The optimal user profiles will be obtained as the agent can perfectly imitate users' decisions.
In such setting, we integrated spatial {\it KG} to reinforcement learning to incrementally learn user representations and generate the next-visit prediction.
Specifically, we formulated the state as the combination of users and spatial {\it KG}, where the mutual interactions were modeled to incrementally update the representations of users and spatial {\it KG} based on the temporal context, by considering long-short term influences of interactions.
The newly developed policy network exploits Double DQN to avoid the problem of overestimating user decisions.
The policy aimed to mimic users' patterns to generate accurate next-visit prediction with the newly designed hierarchical {\it DQN} and improved sampling strategy.
From the experiment, we can observe that the semantics introduced by the spatial {\it KG} improve the modeling of user representations for better understanding user patterns and preferences.
This demonstrates that the proposed priority-based sampling strategy enhances the learning procedure, where {\it TD}-based is better than reward-based priority approach.

\ifCLASSOPTIONcompsoc
\section*{Acknowledgments}
This research was supported by the National Science Foundation(NSF) via the grant number: 1755946.

\bibliographystyle{plain}
\bibliography{tkde}

\end{document}